%% file: ms.tex
\documentclass[10pt]{article}

\usepackage[utf8]{inputenc}
\usepackage{csquotes}
\usepackage{graphicx}
\usepackage{mathtools}
\usepackage{bm}
\usepackage{bbm}
\usepackage{pifont}
\usepackage{dsfont}
\usepackage{graphicx}
\usepackage{color,colortbl}
\usepackage{array}
\usepackage{rotating}
\usepackage{pdflscape}
\usepackage{subcaption}
\usepackage{sidecap}
\usepackage[inline,shortlabels]{enumitem}
\usepackage{adjustbox}
\usepackage[linesnumbered,ruled]{algorithm2e} 

\usepackage{caption}

\input{latex-math/basic-math.tex}
\input{latex-math/basic-ml.tex}

\input{latex-math/ml-automl.tex}

\input{latex-math/ml-mbo.tex}

\newcommand{\rpkg}[1]{\texttt{\href{https://cran.r-project.org/package=#1}{#1}}}
\newcommand{\cmark}{\ding{51}}%
\newcommand{\xmark}{\ding{55}}%

\DeclareMathOperator{\Eop}{\mathds{E}}

\usepackage[natbib=true,backend=biber,style=apa,sortcites=false]{biblatex}
\AtEveryBibitem{%
  \ifboolexpr{%
    test{\ifentrytype{book}} or
    test{\ifentrytype{inbook}} or
    test{\ifentrytype{inproceedings}} or
    test{\ifentrytype{incollection}} or
    test{\ifentrytype{article}} or
    test{\ifentrytype{report}}
  }{%
    \clearfield{urlday}
    \clearfield{urlmonth}
    \clearfield{urlyear}
    \clearfield{primaryClass}
    \clearfield{url}
    \clearfield{issn}
    \clearfield{doi}
    \clearfield{isbn}
  }{}%
  \clearfield{language}
  \clearfield{note}
  \clearfield{extra}
}

\usepackage{fullpage}
\usepackage{setspace}
\usepackage{parskip}
\usepackage{titlesec}
\usepackage[title]{appendix}
\usepackage[section]{placeins}
\usepackage{xcolor}
\usepackage{breakcites}
\usepackage{lineno}
\usepackage{hyphenat}

\PassOptionsToPackage{hyphens}{url}
\usepackage[colorlinks = true,
            linkcolor = blue,
            urlcolor  = blue,
            citecolor = blue,
            anchorcolor = blue]{hyperref}
\usepackage{etoolbox}

\renewenvironment{abstract}
  {{\bfseries\noindent{\abstractname}\par\nobreak}\footnotesize}
  {\bigskip}

\titlespacing{\section}{0pt}{*3}{*1}
\titlespacing{\subsection}{0pt}{*2}{*0.5}
\titlespacing{\subsubsection}{0pt}{*1.5}{0pt}

\usepackage{authblk}
\usepackage{graphicx}
\usepackage[space]{grffile}
\usepackage{latexsym}
\usepackage{textcomp}
\usepackage{longtable}
\usepackage{tabulary}
\usepackage{booktabs,array,multirow}
\usepackage{amsfonts,amsmath,amssymb}
\providecommand\citet{\cite}
\providecommand\citep{\cite}

\newif\iflatexml\latexmlfalse

\AtBeginDocument{\DeclareGraphicsExtensions{.pdf,.PDF,.eps,.EPS,.png,.PNG,.tif,.TIF,.jpg,.JPG,.jpeg,.JPEG}}

\usepackage[english]{babel}
\usepackage{float}
\usepackage[margin=1.5in]{geometry}
\newif\ifarxiv
\arxivtrue

\begin{document}

\title{Hyperparameter Optimization: Foundations, Algorithms, Best Practices and Open Challenges}

% List each person's full name, ORCID iD, affiliation, email address, and
% any conflicts of interest. Please use an asterisk (*) to indicate the
% corresponding author.

% The preferred (but optional) format for author names is First Name,
% Middle Initial, Last Name.~~

% The submitting author is required to provide
% an~\href{https://authorservices.wiley.com/author-resources/Journal-Authors/Submission/orcid.html}{ORCID
% iD}, and all other authors are encouraged to do so.~

\author[1,*]{Bernd Bischl}
\author[1]{Martin Binder}
\author[1,2]{Michel Lang}
\author[1]{Tobias Pielok}
\author[1,2]{Jakob Richter}
\author[1]{Stefan Coors}
\author[1]{Janek Thomas}
\author[3]{Theresa Ullmann}
\author[1]{Marc Becker}
\author[4]{Anne-Laure Boulesteix}
\author[5]{Difan Deng}
\author[5]{Marius Lindauer}

%\author[1]{Alberto Pepe}%
\affil[1]{Department of Statistics, Ludwig-Maximilians-Universität München - \textit{first}.\textit{last}@stat.uni-muenchen.de}%
\affil[2]{Department of Statistics, TU Dortmund University - \textit{last}@statistik.tu-dortmund.de}
\affil[3]{Institute for Medical Information Processing, Biometry and Epidemiology, Ludwig-Maximilians-Universität München - tullmann@ibe.med.uni-muenchen.de}%
\affil[4]{Institute for Medical Information Processing, Biometry and Epidemiology, Ludwig-Maximilians-Universität München - \textit{last}@ibe.med.uni-muenchen.de}%
\affil[5]{Institute for Information Processing, Leibniz University Hannover - \textit{last}@tnt.uni-hannover.de}%
\affil[*]{Corresponding Author}%
\vspace{-1em}

  \date{}

\begingroup
\let\center\flushleft
\let\endcenter\endflushleft
\maketitle
\endgroup

\selectlanguage{english}
\begin{abstract}
\input{sec_01_abstract}
\end{abstract}%

\par\null

\section{Introduction}\label{sec:introduction}
\input{sec_02_introduction}

\section{Related Work}
\label{sec:related_work}

\input{sec_02_02_related}

\section{Supervised Machine Learning}\label{sec:supervised_ml}

\subsection{Terminology and Notations} \label{ssec:terminology} 
\input{sec_03_01_terminology_and_notations}

\subsection{Evaluation of ML Algorithms} \label{ssec:evaluation_of_ml}
\input{sec_03_03_evaluation_of_ml_algorithms}

\section{Hyperparameter Optimization}\label{sec:foundations_of_hpo}
\input{sec_04_hyperparameter_optimization}

\section{Pipelining, Preprocessing, and AutoML}\label{sec:preproc} 
\input{sec_05_pipelining_preprocessing_and_automl}

\section{Practical Aspects of HPO}\label{sec:practical}
\input{sec_06_practical_aspects_of_hpo}

\section{Related Problems}\label{sec:related_problems}
\input{sec_07_related_fields}

\section{Conclusion and Open Challenges}\label{sec:conclusion}
\input{sec_08_conclusion_challenges_and_open_issues}

\input{sec_09_end}

% \section*{References}\sloppy
% \phantomsection

\printbibliography
\newpage

\appendix
\begin{appendices}
\begin{refsection}
\input{appendix_A_learners_and_search_spaces}

\input{appendix_B_preprocessing}

\input{appendix_C_evaluation_metrics}

\section{Software}\label{app:software}
This section lists relevant software packages in both R and Python. Libraries providing popular machine learning algorithms are listed first, followed by additional information about useful packages and relevant considerations for HPO in both languages.

\input{appendix_D_01_software_r.tex}

\input{appendix_D_02_software_python}

\printbibliography[heading=subbibliography]
\end{refsection}
\end{appendices}
\newpage

\end{document}

%% file: latex-math/basic-math.tex
\ifdefined\N                                                                % N, naturals
\renewcommand{\N}{\mathds{N}}                                                % N defined by "siunitx" (which we use), for "NEWTON"
\else
  \newcommand{\N}{\mathds{N}}
\fi
\newcommand{\R}{\mathds{R}}                                                 % R, reals
\ifdefined\C
  \renewcommand{\C}{\mathds{C}}                                             % C, complex
\else
  \newcommand{\C}{\mathds{C}}
\fi
\DeclareMathOperator*{\argmax}{arg\,max}
\DeclareMathOperator*{\argmin}{arg\,min}

\newcommand{\I}{\mathbb{I}}                                                 % I, indicator
\newcommand{\sumin}{\sum\limits_{i=1}^n}											% summation from i=1 to n
\newcommand{\sumim}{\sum\limits_{i=1}^m}											% summation from i=1 to m
\newcommand{\xv}{\mathbf{x}}													% vector x (bold)
\newcommand{\xb}{\mathbf{x}}													% WE SHOULD NOT USE THIS 
\newcommand{\yv}{\mathbf{y}}													% vector y (bold)
\renewcommand{\P}{\mathds{P}}                                               % P, probability
\newcommand{\E}{\mathds{E}}                                                 % E, expectation
\newcommand{\iid}{\overset{i.i.d}{\sim}}                                    % dist with i.i.d superscript

%% file: latex-math/basic-ml.tex
\newcommand{\Xspace}{\mathcal{X}}                                           % X, input space
\newcommand{\Yspace}{\mathcal{Y}}                                           % Y, output space
\newcommand{\Pxy}{\P_{xy}}                                                  % P_xy
\newcommand{\allDatasets}{\mathds{D}}                                       % The set of all datasets
\newcommand{\D}{\mathcal{D}}                                                      % D, data
\newcommand{\obs}[1][i]{\left(\xv^{(#1)},y^{(#1)}\right)}                                                                               % observation (x^(i), y^(i))
\newcommand{\Dset}{\left( \obs[1], \ldots, \obs[n]\right)}    % {(x1,y1)), ..., (xn,yn)}, data
\newcommand{\defAllDatasets}{\bigcup_{n \in \N}(\Xspace \times \Yspace)^n}  % Def. of the set of all datasets 
\renewcommand{\xi}[1][i]{\mathbf{x}^{(#1)}}                                          % x^i, i-th observed value of x
\newcommand{\yi}[1][i]{y^{(#1)}}                                            % y^i, i-th observed value of y 
\newcommand{\Dtrain}{\mathcal{D}_{\text{train}}}                            % D_train, training set
\newcommand{\Dtest}{\mathcal{D}_{\text{test}}}                              % D_test, test set
\newcommand{\preimageInducerShort}{\allDatasets\times\bm{\Lambda}}     % Set of all datasets times the hyperparameter space
\newcommand{\inducer}{\mathcal{I}}                                                % Inducer, inducing algorithm, learning algorithm 

\newcommand{\fx}{f(\mathbf{x})}                                                      % f(x), continuous prediction function
\newcommand{\Hspace}{\mathcal{H}}														% hypothesis space where f is from
\newcommand{\fh}{\hat{f}}                                                   % f hat, estimated prediction function
\newcommand{\fhDlambda}{\fh_{\D, \lambdav}}                                       %model learned on D with hp lambda
\newcommand{\yh}{\hat{y}}                                                   % yhat for prediction of target
\newcommand{\yih}{\hat{y}^{(i)}}                                            % yhat^(i) for prediction of ith targiet

\newcommand{\thetah}{\bm{\hat{\theta}}}  
\newcommand{\thetab}{\bm{\theta}}											% theta vector
\newcommand{\pikxh}{\hat \pi_k(\mathbf{x})}                                          % pi_k(x) hat, P(y = k | x) hat
\newcommand{\pikxih}{\hat \pi_k(\xi)}                                   % pi_k(x^(i)) with hat

\newcommand{\Lxy}{L\left(y, \fx\right)}                                               % L(y, f(x)), loss function
\newcommand{\risk}{\mathcal{R}}                                             % R, risk
\newcommand{\riskf}{\risk(f)}                                               % R(f), risk
\newcommand{\riske}{\mathcal{R}_{\text{emp}}}                               % R_emp, empirical risk (without factor 1 / n
\newcommand{\riskef}{\riske(f)}                                             % R_emp(f)
\newcommand{\risket}{\mathcal{R}_{\text{emp}}(\bm{\theta})}                      % R_emp(theta)
\newcommand{\ntest}{n_{\mathrm{test}}}                              % size of the test set
\newcommand{\ntrain}{n_{\mathrm{train}}}                            % size of the train set
\newcommand{\ntraini}[1][i]{n_{\mathrm{train},#1}}                  % size of the i-th train set
\newcommand{\Jtrain}{J_\mathrm{train}}                              % index vector associated to the train data
\newcommand{\Jtest}{J_\mathrm{test}}                                % index vector associated to the test data
\newcommand{\Jtraini}[1][i]{J_{\mathrm{train},#1}}                  % index vector associated to the i-th train dataset
\newcommand{\Jtesti}[1][i]{J_{\mathrm{test},#1}}                    % index vector associated to the i-th test dataset
\newcommand{\Dtraini}[1][i]{\mathcal{D}_{\text{train},#1}}          % D_train,i, i-th training set
\newcommand{\JtrainSpace}{\{1,\dots,n\}^{\ntrain}}                  % space of train indices of size m_train
\newcommand{\JtestSpace}{\{1,\dots,n\}^{\ntest}}                    % space of train indices of size m_test
\newcommand{\yJ}[1][J]{\yv_{#1}}                                    % output vector associated to index J
\newcommand{\JJ}{\mathcal{J}}                                       % cali-J, set of all splits
\newcommand{\JJset}{\left((\Jtraini[1], \Jtesti[1]),\dots,(\Jtraini[B], \Jtesti[B])\right)}
\newcommand{\GEh}{\widehat{\mathrm{GE}}}                                             % GE-hat 
\newcommand{\GEfull}[1][\ntrain]{\mathrm{GE}(\inducer, \lambdav, #1, \rho)}          % GE(I, lam, ?, rho)
\newcommand{\GEhholdout}{\GEh_{\Jtrain, \Jtest}(\inducer, \lambdav, \ntrain, \rho)}    % GE-hat_{Jtrain,Jtest} (I, lam, |J|, rho)
\newcommand{\GEhholdouti}[1][i]{\GEh_{\Jtraini[#1], \Jtesti[#1]}(\inducer, \lambdav, |\Jtraini[#1]|, \rho)}           
\newcommand{\GEhresa}{\GEh(\inducer, \JJ, \rho, \lambdav)}                   % GE-hat_I,J,rho(lam) 

\newcommand{\agr}{\mathrm{agr}}                       % aggregate function
\newcommand{\rhoL}{\rho_L}                 % perf. measure derived from pointwise loss function L
\newcommand{\F}{\boldsymbol{F}}             % matrix of prediction scores
\newcommand{\Fi}[1][i]{\F^{(#1)}}             % i'th row vector of the prediction scores matrix
\newcommand{\FJ}[1][J]{\F_{#1}}             % prediction scores matrix regarding index vector J 
\newcommand{\FJf}{\FJ[J,f]}             % prediction scores matrix regarding index vector J and model f
\newcommand{\FJtestftrain}{\F_{\Jtest,\inducer(\Dtrain, \lambdav)}}             % prediction scores matrix regarding index vector Jtest and model f
\newcommand{\FJtestftraini}[1][i]{\F_{\Jtesti[#1],\inducer(\Dtraini[#1], \lambdav)}}             % prediction scores matrix regarding i-th index vector Jtest and model f
\newcommand{\preimageRho}{\bigcup_{m\in\N}\left(\Yspace^m\times\R^{m\times g}\right)}     % Set of all datasets times the hyperparameter space

%% file: latex-math/ml-automl.tex
\newcommand{\lambdav}{\bm{\lambda}}											% lambda vector

%% file: latex-math/ml-mbo.tex
\newcommand{\Ilam}{\inducer_{\lambdav}}						% I_lambda
\newcommand{\lami}{\lambdav^{(i)}}							% i-th candidate
\newcommand{\clam}{c(\lambdav)}                             % c(lambda)
\newcommand{\clamh}{c(\lamh)}                               % c(lambda-hat)
\newcommand{\lams}{\lambdav^{*}}		                    % Theoretical min of c
\newcommand{\lamh}{\hat{\lambdav}}		                    % returned lambda of HPO
\newcommand{\LamS}{\tilde\Lambda}                           % search space
\newcommand{\lamp}{\lambdav^+}                              % proposed lambda
\newcommand{\clamp}{c(\lamp)}                               % c of proposed lambda
\newcommand{\archive}{\mathcal{A}}                          % archive at time step t
\newcommand{\archivet}[1][t]{\mathcal{A}^{[#1]}}            % archive at time step t

\newcommand{\tuner}{\mathcal{T}}                            % tuner
\newcommand{\tunerfull}{\tuner_{\inducer,\LamS, \rho,\JJ}}
\newcommand{\chlam}{\hat{c}(\lambdav)}                       % post mean of SM 
\newcommand{\shlam}{\hat{\sigma}(\lambdav)}                  % post sd of SM
\newcommand{\ulam}{u(\lambdav)}                              % acquisition function

\newcommand{\lamfid}{\lambda_{\text{fid}}}              % single lambda_budget komponent HP
\newcommand{\lamfidl}{\lamfid^{\textrm{low}}}
\newcommand{\lamfidu}{\lamfid^{\textrm{upp}}}
\newcommand{\etahb}{\eta_{\text{HB}}}                        % HB multiplier eta

%% file: sec_01_abstract.tex
Most machine learning algorithms are configured by a set of hyperparameters whose values must be carefully chosen and which often considerably impact performance. To avoid a time-consuming and irreproducible manual process of trial-and-error to find well-performing hyperparameter configurations, various automatic hyperparameter optimization (HPO) methods -- e.g., based on resampling error estimation for supervised machine learning -- can be employed.
After introducing HPO from a general perspective, this paper reviews important HPO methods, from simple techniques such as grid or random search to more advanced methods like evolution strategies, Bayesian optimization, Hyperband and racing.
This work gives practical recommendations regarding important choices to be made when conducting HPO, including the HPO algorithms themselves, performance evaluation, how to combine HPO with machine learning pipelines, runtime improvements, and parallelization.
\ifarxiv
This work is accompanied by an appendix that contains information on specific software packages in R and Python, as well as information and recommended hyperparameter search spaces for specific learning algorithms. We also provide notebooks that demonstrate concepts from this work as supplementary files.
\fi

%% file: sec_02_introduction.tex
Machine learning (ML) algorithms are highly configurable by their hyperparameters (HPs).
These parameters often substantially influence the complexity, behavior, speed as well as other aspects of the learner, and their values must be selected with care in order to achieve optimal performance.  
Human trial-and-error to select these values is time-consuming, often somewhat biased, error-prone and computationally irreproducible.

As the mathematical formalization of hyperparameter optimization (HPO) is essentially black-box optimization, often in a higher-dimensional space, this is better delegated to appropriate algorithms and machines to increase efficiency and ensure reproducibility.
Many HPO methods have been developed to assist in and automate the search for well-performing hyperparameter configuration (HPCs) over the last 20 to 30 years.
However, more sophisticated HPO approaches in particular are not as widely used as they could (or should) be in practice. 
We postulate that the reason for this may be a combination of the following factors:
\begin{itemize}
    \item poor understanding of HPO methods by potential users, who may perceive them as (too) complex \enquote{black-boxes}; 
    \item poor confidence of potential users in the superiority of HPO methods over trivial approaches and resulting skepticism of the expected return on (time) investment;
    \item missing guidance on the choice and configuration of HPO methods for the problem at hand;
    \item difficulty to define the search space of an HPO process appropriately.
\end{itemize}

With these obstacles in mind, this paper formally and algorithmically introduces HPO, with many hints for practical application. 
Our target audience are scientists and users with a basic knowledge of ML and evaluation. 
%However, to avoid misunderstandings and ambiguities, the here used terminology and notation for ML are introduced formally and concisely before diving into HPO. We propose to skip the first sections if terminology and evaluation concepts of machine learning are already known.

In this article, we mainly discuss HP for supervised ML, which is arguably the default scenario for HPO. 
We mainly do this to keep notation simple and to not overwhelm less experienced readers, especially for less experienced readers.
Nevertheless, all covered techniques can be applied to practically any algorithm in ML in which the algorithm is trained on a collection of instances and performance is quantitatively measurable -- e.g., in semi-supervised learning, reinforcement learning, and potentially even unsupervised learning\footnote{where the measurement of performance is arguably much less straightforward, especially via a single metric.}.

Subsequent sections of this paper are organized as follows:
Section~\ref{sec:related_work} discusses related work.
Section~\ref{sec:supervised_ml} introduces the concept of supervised ML and discusses the evaluation of ML algorithms. 
The principle of HPO is introduced in Section~\ref{sec:foundations_of_hpo}. Major classes of HPO methods are described, including their strengths and limitations. The problem of over-tuning, the handling of noise in the context of HPO, and the topic of threshold tuning are also addressed. 
Section~\ref{sec:preproc} introduces the most common preprocessing steps and the concept of ML pipelines, which enables us to include preprocessing and model selection within HPO.
Section~\ref{sec:practical} offers practical recommendations on how to choose resampling strategies as well as define tuning search spaces, provides guidance on which HPO algorithm to use, and describes how HPO can be parallelized.
In Section~\ref{sec:related_problems}, we also briefly discuss how HPO directly connects to a much broader field of algorithm selection and configuration beyond ML and other related fields. 
Section~\ref{sec:conclusion} concludes with a discussion of relevant open issues in HPO.

\ifarxiv
The appendices contain additional material of particularly practical relevance for HPO: Appendix~\ref{app:important_ml_algorithms} lists the most popular ML algorithms, describes some of their properties, and proposes sensible HP search spaces; Appendix~\ref{app:preprocessing} does the same for preprocessing methods; Appendix~\ref{app:evaluation} contains a table of common evaluation metrics; Appendix~\ref{app:software} lists relevant considerations and software packages for ML and HPO in the two popular ML scripting languages R (\ref{ssec:software_r}) and Python (\ref{ssec:software_python}). Furthermore, we provide several R markdown notebooks as ancillary files which demonstrate many practical HPO concepts and implement them in \rpkg{mlr3} \citep{lang_mlr3_2019}.
\fi

%% file: sec_02_02_related.tex
As one of the most studied sub-fields of automated ML (AutoML), there exist several previous surveys on HPO. \citet{feurer-automlbook19a} offered a thorough overview about existing HPO approaches, open challenges, and future research directions. In contrast to our paper, however, that work does not focus on specific advice for issues that arise in practice. 
\citet{yang-neuro20a} provide a very high-level overview of search spaces, HPO techniques, and tools. Although we expect that the paper by \citet{yang-neuro20a} will be a more accessible paper for first-time users of HPO compared to the survey by \citet{feurer-automlbook19a}, it does not explain HPO's mathematical and algorithmic details or practical tips on how to apply HPO efficiently. Last but not least, \citet{Andonie19} provides an overview about HPO methods, but with a focus on computational complexity aspects. We see this work described here as filling the gap between these papers by providing all necessary details both for first-time users of HPO as well as experts in ML and data science who seek to understand the concepts of HPO in sufficient depth.

Our focus is on providing a general overview of HPO without a special focus on concrete ML model classes. However, since the ML field has many large sub-communities by now, there are also several specialized HPO and AutoML surveys. For example, \citet{he-kbs21a}
\ifarxiv
and \citet{talbi2020optimization}
\fi
focus on AutoML for deep learning models, \citet{khalid2020survey} on HPO for forecasting models in smart grids, and \citet{zhang-ijcai21a} on AutoML on graph models.
\ifarxiv
\citet{bartz2021experimental} investigate model-based HPO and also give search spaces and examples for many specific ML algorithms. Other more general reviews of AutoML are \citet{yao2019taking}, \citet{elshawi2019automated}, and \citet{yu2020hyperparameter}.
\fi

%  further automl survey: zoeller-jair21a

% Further papers without peer-reviewing: 
% * https://arxiv.org/abs/2107.08761
% * https://arxiv.org/abs/2003.05689
% * https://hal.inria.fr/hal-02570804v2/document
% * https://arxiv.org/abs/1810.13306
% * https://arxiv.org/abs/1906.02287

%% file: sec_03_01_terminology_and_notations.tex
Supervised ML addresses the problem of inferring a model from labeled training data that is then used to predict data from the same underlying distribution with minimal error.
Let $\D$ be a labeled data set with $n$ observations, where each observation $(\xi, \yi)$ consists of a $p$-dimensional feature vector\footnote{More generally, with a slight abuse of notation, the feature vector $\xi$ could be taken as a tensor (for example when the feature is an image).} 
$\xi \in \Xspace$ and its label $\yi \in \Yspace$. Hence, we define the data set\footnote{More precisely, $\D$ is an indexed tuple, but we will continue to use common terminology and call it a data \textit{set}.} $\D = \Dset$.
We assume that $\D$ has been sampled i.i.d.\ from an underlying, unknown distribution, so $\D \sim (\Pxy)^n$. 
Each dimension of a $p$-dimensional $\xb$ will usually be of numerical, integer, or categorical type.
While some ML algorithms can handle all of these data types natively (e.g., tree-based methods), others can only work on numeric features and require encoding techniques for categorical types. The most common supervised ML tasks are \textit{regression} and \textit{classification}, where
$\Yspace=\R$ for regression and $\Yspace$ is finite and categorical for classification with $\vert \Yspace\vert=g$ classes.
Although we mainly discuss HPO in the context of regression and classification, all covered concepts easily generalize to 
other supervised ML tasks, such as Poisson regression, survival analysis, cost-sensitive classification, multi-output tasks, and many more.
An ML model is a function $f:\Xspace \rightarrow \R^g$ that assigns a prediction in $\R^g$ to a feature vector from $\Xspace$. 
For regression, $g$ is $1$, while in classification the output represents the $g$ \textit{decision scores} or posterior probabilities of the $g$ candidate classes.
%, in which case the component functions $f_j$, $j=1,\dots,g$, are called \textit{discriminant functions}. 
Binary classification is usually simplified to $g=1$, with a single decision score in $\R$ or only the posterior probability for the positive class.
%and negative values to the negative class) or in $[0,1]$ (representing the probability of the positive class). 
The function space -- usually parameterized -- to which a model belongs is called the \textit{hypothesis space} and denoted as $\Hspace$.

The goal of supervised ML is to fit a model given $n$ observations sampled from $\Pxy$, so that it generalizes well to new observations from the same data generating process.
%$\fh$ or its associated parameter vector $\thetah$ from a data set $\D$, so that $\fh$ predicts data sampled from $\Pxy$ best. 
Formally, an ML learner or \textit{inducer} $\inducer$ configured by HPs $\lambdav \in \Lambda$ maps a data set $\D$ to a model $\fh$ 
or equivalently to its associated parameter vector $\thetah$, i.e.,
\begin{equation}\label{eq:ml_algorithm}
\begin{aligned}
\inducer : \preimageInducerShort &\to \Hspace, \quad
(\D, \lambdav) &\mapsto \fh,
\end{aligned}
\end{equation}
where $\allDatasets := \defAllDatasets$ is the set of all data sets. 
%The inducer can also be seen as a mapping to the associated parameter vector $\thetah$.
While model parameters $\thetah$ are an output of the learner $\inducer$, HPs $\lambdav$ are an input.
We also write $\Ilam$ for $\inducer$ or $\fhDlambda$ for $\fh$ if we want to stress that the inducer was configured with $\lambdav$ or that the model was learned on $\D$ by an inducer configured by $\lambdav$.
A loss function $L : \Yspace \times \R^g \rightarrow \R^+_0$ measures the discrepancy between the prediction and the true label.
Many ML learners use the concept of \emph{empirical risk minimization} (ERM) in their training routine to produce their fitted model $\fh$, i.e., they optimize $\riskef$ or $\risket$ over all candidate models $f \in \Hspace$
\begin{equation}
   \riskef := \sumin L\left(\yi,f(\xi)\right)\mathrm{,}\qquad \fh = \argmin_{f} \riskef
        \label{eq:emprisk}
\end{equation}
on the training data $\D$ (c.f. Figure~\ref{fig:riskmin2}). 
%Naturally, we can also express the empirical risk w.r.t. $\theta$ by writing $\risket: \Theta \rightarrow \R$ instead.
%
This empirical risk is only a stochastic proxy for what we are actually interested in, namely the theoretical risk or true generalization error
%
%\begin{equation}
    $\riskf := \E_{(\xv, y) \sim \Pxy} [\Lxy]$.
%    = \int \Lxy \text{d}\Pxy.
%\end{equation}
%
%\begin{figure}[b!]
%    \centering
%    \includegraphics[width = 0.8\textwidth]{figures/riskmin_bilevel1.pdf}
%    \caption{The learning algorithm (or inducer) $\inducer$ takes the input data, internally performs empirical risk minimization, and returns the model $\fh$, represented by $\thetah$, that minimizes the empirical risk.}
%    \label{fig:riskmin1}
%\end{figure}
%
For many complex hypothesis spaces, $\riskef$ can become considerably smaller than its true risk $\riskf$.
This phenomenon is known as overfitting, which in ML is usually addressed by either constraining the hypothesis space or regularized risk minimization, i.e., adding a complexity penalty $J(\thetab)$ to~(\ref{eq:emprisk}).

%The overfitting problem has major consequences which will be addressed in the next sections: 
%\begin{enumerate*}[(i)]
    %\item a good inducer tries to avoid overfitting as far as possible when training the model, via \textit{regularization}, and
%    \item the generalization error is better estimated using an independent test data set or through \textit{resampling}.
%\end{enumerate*}

%% file: sec_03_03_evaluation_of_ml_algorithms.tex
After training an ML model $\fh$, a natural follow-up step is to evaluate its future performance given new, unseen data. We seek to use an unbiased, high-quality statistical estimator, which numerically quantifies the performance of our model when it is used to predict the target variable for new observations drawn from the same data-generating process $\Pxy$.
%Several problems arise which make ML model evaluation more difficult than it may initially seem. 

%\begin{itemize}
%\item Performance metrics in ML are sometimes defined as set functions instead of point-wise losses, e.g.\ the area under the ROC curve (AUC).
%\item Evaluating the model on the data that it was trained on would give an optimistically biased performance estimate, leading to the \enquote{evaluate only on an untouched test set} principle.
%\item A critical problem in ML working against this principle is that usually only a single data set is available, which must be simultaneously used for building and evaluating the model. It follows that this data set has to be split. 
%\item If the size of the data set is limited, fitting and evaluating the learner on a single training and test split usually results in a high variance estimator of the performance, hence the need for repeating this splitting procedure in order to smooth out variance by averaging. Such procedures are referred to as \lq\lq resampling''.
%\item HPO optimizes resampled performance, but needs to be evaluated \textit{itself} by the same principle from outside. 
%\end{itemize}

%eduardogarrido90: We can also say here, as a problem, that our data sample (training data) may belong to a subspace of the probability distribution generating data. Hence, we will not learn all the features of the prob dist.
%eduardogarrido90: Also, speak about the tradeoff of obtaining a better estimator (lower bias and variance) vs computational time (e.g. train-test split vs one leave out cross validation)

\subsubsection{Performance Metrics} 
A general performance measure $\rho$ for an arbitrary data set of size $m$ is defined as a two-argument function that maps the $m$-size vector of true labels $\yv$ and the $m \times g$ matrix of prediction scores $\F$ to a scalar performance value:
\begin{equation}
\rho: \preimageRho \rightarrow \R, \quad (\yv, \F) \mapsto \rho(\yv, \F).
\end{equation}
This more general set-based definition is needed for performance measures -- such as area under the ROC curve (AUC) -- or for most measures from survival time analysis, where loss values cannot be computed with respect to only a single observation.
For usual point-wise losses $\Lxy$,
%Many measures from regression or classification consider point-wise loss $\Lxy$ which measures the difference between $\fx$ and $y$ for a single point $\xv$.
we can simply extend $L$ to $\rho$ by averaging over the size-$m$ set used for testing:
\begin{equation}
\rhoL (\yv, \F) = \frac{1}{m}\sumim L(\yi, \Fi),
\end{equation}
where $\Fi$ is the $i$-th row of $\F$; this corresponds to estimating the theoretical risk $\riskf$ corresponding to the given loss.
\ifarxiv
Popular performance metrics corresponding to different loss functions can be found in Table~\ref{tab:measures} in Appendix~\ref{app:evaluation}.
\fi

Furthermore, the introduction of $\rho$ allows the evaluation of a learner with respect to a different performance metric than the loss used for risk minimization. Because of this, we call the loss used in \eqref{eq:emprisk} \textit{inner loss}, and $\rho$ the outer performance measure or \textit{outer loss}\footnote{\textit{Surrogate loss} for the inner loss and \textit{target loss} for the outer loss are also commonly used terminologies.}.
Both can coincide,
%,e.g., we could optimize a binary classifier with respect to log-loss and also evaluate with respect to it.,
but quite often we select an outer performance measure based on the prediction task we would like to solve, and opt to approximate this metric with a computationally cheaper and possibly differentiable version during inner risk minimization. 
%This requires a careful choice of the inner metric s.t. there is sufficient correlation between the outer and the inner metric. 

\subsubsection{Generalization Error} 
\begin{figure}[b!]
    \centering
    \includegraphics[width = 0.6\textwidth]{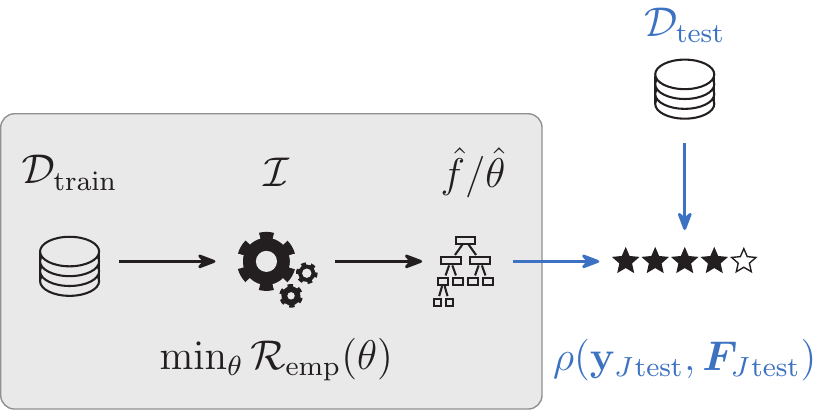}
    \caption{Learner $\inducer$ takes input data, performs ERM, and returns model $\fh$ and its parameters $\thetah$. The GE of $\fh$ is evaluated on the fresh test set $\Dtest$.}
    \label{fig:riskmin2}
\end{figure}

Due to potential overfitting, every predictive model should be evaluated on unseen test data to ensure unbiased performance estimation. 
%An independent test set is not always available in practice. 
%
%Hence, we usually split the available data set $\D$ with $n$ observations into two subsets, $\Dtrain$ and $\Dtest$ of sizes $\ntrain$ and $\ntest$, respectively, which are usually non-overlapping and with $\ntrain + \ntest = n$. 
%
Assuming (for now) dedicated train and test data sets $\Dtrain$ and $\Dtest$ of sizes $\ntrain$ and $\ntest$, respectively, 
%we can also represent both sets as index vectors $\Jtrain \in \JtrainSpace$ and $\Jtest \in \JtestSpace$.
%For an index vector $J$ of length $m$ one can define the corresponding vector of labels $\yJ \in \Yspace^m$,
%and the corresponding matrix of prediction scores $\FJf \in \R^{m\times g}$ for a model $f$.
%
%, i.e.,$\FJf = \FJfDef.$ Here, $\xv^{(J^{(i)})}$ is the i-th feature vector associated to $J$.
%
%
we define the \emph{generalization error} of a learner $\inducer$ with HPs $\lambdav$ trained on $\ntrain$ observations, w.r.t.\ measure $\rho$ as
\begin{equation}
\label{eq:gef}
\GEfull := \lim_{\ntest\rightarrow\infty} \E_{\Dtrain,\Dtest \sim \Pxy} \left[ \rho\left(\yv_{\mathrm{test}}, \F_{\Dtest,\inducer(\Dtrain, \lambdav)}\right)\right]\textrm{,}
\end{equation}
where we take the expectation over the data sets $\Dtrain$ and $\Dtest$, both i.i.d.\ from $\Pxy$, and $\F_{\Dtest,\inducer(\Dtrain, \lambdav)}$ is the matrix of predictions when the model is trained on $\Dtrain$ and predicts on $\Dtest$.
Note that in the simpler and common case of a point-wise loss $\Lxy$, the above trivially reduces to the more common form 
\begin{equation}
\label{eq:ge}
\mathrm{GE}(\inducer, \lambdav, \ntrain, \rho_L) =  \E_{\Dtrain,(\xv,y) \sim \Pxy} \left[L(y, \Ilam(\Dtrain)(\xv)) \right]
\end{equation}
with expectation over data set $\Dtrain$ and test sample $(\xv, y)$, both independently sampled from $\Pxy$.
This corresponds to the expectation of $\riskf$ -- which references a given, fixed model -- over all possible models fitted to different realizations of $\Dtrain$ of size $\ntrain$ (see Figure~\ref{fig:riskmin2}).

\subsubsection{Data splitting and Resampling} 

The generalization error must usually be estimated from a single given data set $\D$. For a simple estimator based on a single random split, $\Dtrain$ and $\Dtest$ can be represented as index vectors $\Jtrain \in \JtrainSpace$ and $\Jtest \in \JtestSpace$, which usually partition the data set. 
For an index vector $J$ of length $m$, one can define the corresponding vector of labels $\yJ \in \Yspace^m$, and the corresponding matrix of prediction scores $\FJf \in \R^{m\times g}$ for a model $f$. 
%, i.e.,$\FJf = \FJfDef.$ Here, $\xv^{(J^{(i)})}$ is the i-th feature vector associated to $J$.
The \emph{holdout estimator} is then:
%If the generalization error is estimated based on a single split as
\begin{equation}
\GEhholdout = \rho(\yJ[\Jtest], \FJtestftrain).
\label{eq:holdout}
\end{equation}
%we call this \textit{hold-out},
%because one part of the data is not used (i.e., held out) for training.
%and the index usually forms a random partitioning of $\D$. 
The holdout approach has the following trade-off:
%statistical properties:
(i) Because $\ntrain$ must be smaller than $n$, the estimator is a pessimistically biased estimator of $\GEfull[n]$, as we do not use all available data for training. In a certain sense, we are estimating with respect to the wrong training set size.
(ii) If $\Dtrain$ is large, $\Dtest$ will be small, and the estimator (\ref{eq:holdout}) has a large variance.
%When choosing the splitting ratio, we must make a trade-off decision between (i) and (ii).
%between the variance and bias of the estimator (\ref{eq:testset}), 
This trade-off not only depends on relative sizes of $\ntrain$ and $\ntest$, but also the absolute number of observations, as learning with respect to $\ntrain$ sample size and test error estimation based on $\ntest$ samples both show a saturating effect for larger sample sizes. 
However, a typical rule of thumb is to choose $\ntrain = \frac{2}{3} n$ \citep{kohavi1995cross,dobbin2011optimally}.

Resampling methods offer a partial solution to this dilemma.
These methods repeatedly split the available data into training and test sets, then apply an estimator (\ref{eq:holdout}) for each of these, and finally aggregate over all obtained $\rho$ performance values. 
Formally, we can identify a resampling strategy with a vector of corresponding splits, i.e.,
$\JJ = \JJset,$
where $\Jtraini, \Jtesti$ are index vectors and $B$ is the number of splits. 
Hence, the estimator for Eq.~\eqref{eq:gef} is:
\begin{equation}
\label{eq:ges}
\begin{aligned}
%\begin{alignedat}{2}
& \GEhresa = \\
&= \agr\Big(\GEhholdouti[1], \dots, \GEhholdouti[B]\Big)\\
&= \agr\Big(\rho\left(\yv_{\Jtesti[1]},\FJtestftraini[1]\right), \dots,\rho\left(\yv_{\Jtesti[B]},\FJtestftraini[B]\right)\Big),
%\end{alignedat}
\end{aligned}
\end{equation}
where the aggregator $\agr$ is often chosen to be the mean. 
%Depending on the performance measure $\rho$, however, other aggregating functions might be more appropriate. 
For Eq.~\eqref{eq:ges} to be a valid estimator of Eq.~\eqref{eq:ge}, we must specify to what $\ntrain$ training set size an estimator refers in $\GEfull$. 
As the training set sizes can be different during resampling (they usually do not vary much), it should at least hold that $\ntrain \approx \ntraini[1] \approx \dots \approx \ntraini[B]$, and we could take the average for such a required reference size with $\ntrain = \frac{1}{B}\sum_{j=1}^B \ntraini[j]$.

Resampling uses the data more efficiently than a single holdout split, as the repeated estimation and averaging over multiple splits results in an estimate of generalization error with lower variance \citep{kohavi1995cross, simon2007}. 
Additionally, the pessimistic bias of simple holdout is also kept to a minimum and can be reduced to nearly $0$ by choosing training sets of size close to $n$.
%
%The discussed properties of resampling only hold for i.i.d.\ data.
%For more data with a dependence structure, such as time series, different resampling strategies, e.g., an increasing window CV for time series data should be used \citep{BERGMEIR201870}.
%
%
The most widely-used resampling technique is arguably
\textbf{$k$-fold-cross-validation} (CV), which partitions the available data in $k$ subsets of approximately equal size, and uses each partition to evaluate a model fitted on its complement.
% \ifarxiv
% as depicted in the appendix in Figure~\ref{fig:3cv}. 
% \else
% .
% \fi
For small data sets, it makes sense to repeat CV with multiple random partitions and to average the resulting estimates in order to average out the variability, which results in \textbf{repeated $k$-fold-cross-validation}.
Furthermore, note that performance values generated from resampling splits and especially CV splits are not statistically independent because of their overlapping traing sets, so the variance of $\GEhresa$ is not proportional to $1/B$. Somewhat paradoxically, a leave-one-out strategy is not the optimal choice, and repeated cross-validation with many (but fewer than $n$) folds and many repetitions is often a better choice \citep{bengio2004no}.
An overview of existing resampling techniques can be found in \citet{bischl2012} or \citet{boulesteix2008evaluating}.
%Other basic approaches include \textbf{subsampling}, i.e., independently repeated holdout splitting, or bootstrapping, where training data sets are drawn with repetition until multiple in-bag sets of size $n$ is created and where the corresponding test sets are all out-of-bag observations for this fold. 
%For large data sets on the other hand, simple holdout splitting (sample large number of observations for training, predict and evaluate on left-out observations) is a popular alternative.
%
%Usually, if one is interested only in the point estimator of $\GEhresa$ without further inference, CV results in a better estimator than subsampling or bootstrapping \citep{kohavi1995study}.

%One should be aware that the individual CV splits are not independent samples, and that error estimates from individual splits are therefore not i.i.d.\ -- one should not use cross-validated statistics for tests or estimates of confidence intervals naively \citep{bengio2004no, vanwinckelen2012estimating}.

%Please note that all outlined schemes in this section discuss the evaluation of running a learner in its default (or an a-priorily fixed) configuration. If HPO is involved, it is required to use nested cross-validation, which we discuss in Section~\ref{ssec:nested}.

%% file: sec_04_hyperparameter_optimization.tex
\subsection{HPO Problem Definition} \label{ssec:hpo_definiton}

Most learners are highly configurable by HPs, and their generalization performance usually depends on this configuration in a non-trivial and subtle way. 
HPO algorithms automatically identify a well-performing HPC $\lambdav \in \LamS$ for an ML algorithm $\Ilam$.
%Since some HPCs might be meaningless in the first place and can be excluded based on expert knowledge, the actually 
The search space $\LamS \subset \Lambda$ contains all considered HPs for optimization and their respective ranges:
\begin{equation}
    \LamS = \LamS_1 \times \LamS_2 \times \dots \times \LamS_l,
\end{equation}
where $\LamS_i$ is a bounded subset of the domain of the $i$-th HP $\Lambda_i$, and can be either continuous, discrete, or categorical. This already mixed search space can also contain \textit{dependent HPs}, leading to a hierarchical search space:
An HP $\lambda_i$ is said to be \textit{conditional} on $\lambda_j$ if $\lambda_i$ is only active when $\lambda_j$ is an element of a given subset of $\Lambda_j$ and inactive otherwise, i.e., not affecting the resulting learner \citep{autoweka}. 
Common examples are kernel HPs of a kernelized machine such as the SVM, when we tune over the kernel type and its respective hyperparameters as well. %The RBF width HP $\gamma$ might only be active if we fit a SVM with a polynomial or radial basis function kernel.
Such conditional HPs usually introduce tree-like dependencies in the search space, and may in general lead to dependencies that may be represented by directed acyclic graphs.  

\begin{figure}[h]
    \centering
    \includegraphics[width = 1.0\textwidth]{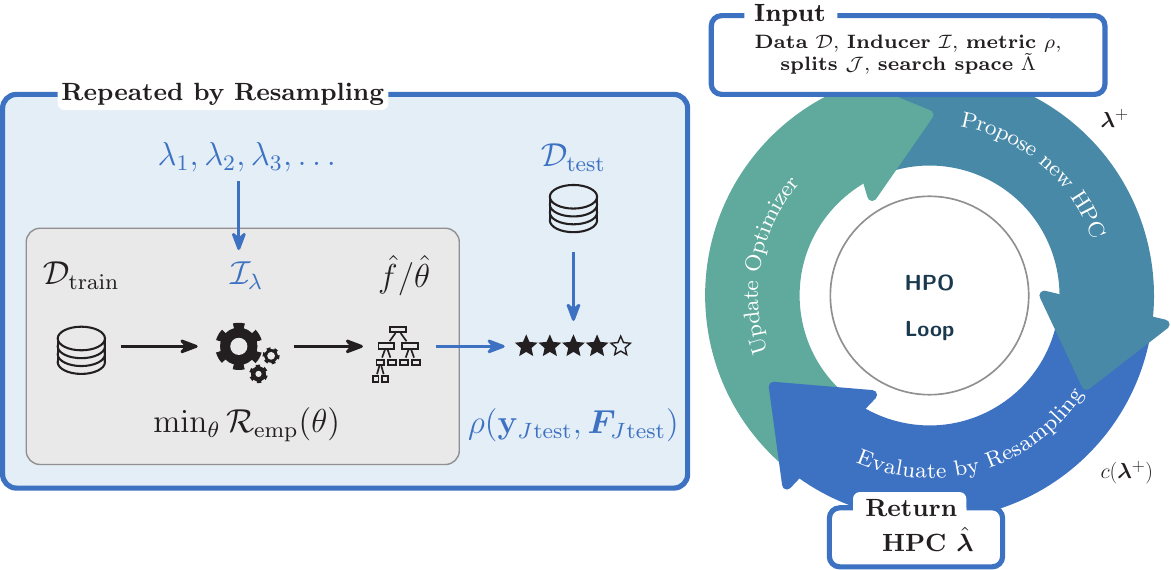}
    %{figures/riskmin_bilevel3.png}
\caption{General HPO loop with inner risk minimization.}
    \label{fig:hpo_loop_1}
\end{figure}

The general HPO problem as visualized in Figure~\ref{fig:hpo_loop_1} is defined as:
\begin{eqnarray}
    \lams \in \argmin_{\lambdav \in \LamS} \clam = \argmin_{\lambdav \in \LamS} 
    %\GEhlamsubIJrho
    \GEhresa
    \label{eq:hpo_objective}
\end{eqnarray}
where $\lams$ denotes the theoretical optimum, and $\clam$ is a shorthand for the estimated generalization error when $\inducer, \JJ, \rho$ are fixed. 
We therefore estimate and optimize the generalization error $\GEhresa$ of a learner $\Ilam$, w.r.t.\ an HPC $\lambdav$, based on a resampling split $\JJ = \JJset$.\footnote{Note again that optimizing the resampling error will result in biased estimates, which are problematic when reporting the generalization error; use nested CV for this, see Section~\ref{ssec:nested}.}
Note that $\clam$ is a black-box, as it usually has no closed-form mathematical representation, and hence no analytic gradient information is available. 
Furthermore, the evaluation of $\clam$ can take a significant amount of time.
Therefore, the minimization of $\clam$ forms an \emph{expensive black-box} optimization problem. 

Taken together, these properties define an optimization problem of considerable difficulty. Furthermore, they rule out many popular optimization methods that require gradients or entirely numerical search spaces or that must perform a large number of evaluations to converge to a well-performing solution, like many meta-heuristics.
Furthermore, as $\clam = \GEhresa$, which is defined via resampling and evaluates $\lambdav$ on randomly chosen validation sets, $c$ should be considered a stochastic objective -- although many HPO algorithms may ignore this fact or simply handle it by assuming that we average out the randomness through enough resampling replications.

We can thus define the HP tuner $\tau: (\D,\inducer,\LamS,\rho)\mapsto \lamh$ that proposes its estimate $\lamh$ of the true optimal configuration $\lams$ given a dataset $\D$, an inducer $\inducer$ with corresponding search space $\LamS$ to optimize, and a target measure $\rho$. 
The specific resampling splits $\JJ$ used can either be passed into $\tau$ as well or are internally handled to facilitate adaptive splitting or multi-fidelity optimization (e.g., as done in \citealp{klein2017fast}).

\subsection{Well-Established HPO Algorithms} 
\label{ssec:important_hpo_algorithms}

All HPO algorithms presented here work by the same principle: they iteratively \emph{propose} HPCs $\lamp$ and then \emph{evaluate} the performance on these configurations. We store these HPs and their respective evaluations successively in the so-called \textit{archive} $\archive = ((\lambdav^{(1)}, c(\lambdav^{(1)})), (\lambdav^{(2)}, c(\lambdav^{(2)})), \dots)$, with $\archivet[t+1] = \archivet[t] \cup (\lamp, \clamp)$ if a single point is proposed by the tuner. % The best solution found by the tuner (i.e. HPO algorithm) will be denoted as $\lamh$.

Many algorithms can be characterized by how they handle two different trade-offs: 
a) The exploration vs. exploitation trade-off refers to how much budget an optimizer spends on either attempting to directly exploit the currently available knowledge base by evaluating very close to the currently best candidates (e.g., local search) or exploring the search space to gather new knowledge (e.g., random search).
b) The inference vs. search trade-off refers to how much time and overhead is spent to induce a model from the currently available archive data in order to exploit past evaluations as much as possible.
Other relevant aspects that HPO algorithms differ in are: \textit{Parallelizability}, i.e., how many configurations a tuner can (reasonably) propose at the same time; \textit{global vs. local} behavior of the optimizer, i.e., if updates are always quite close to already evaluated configurations; \textit{noise handling}, i.e., if the optimizer takes into account that the estimated generalization error is noisy; \textit{multifidelity}, i.e., if the tuner uses cheaper evaluations, for example on smaller subsets of the data, to infer performance on the full data; \textit{search space complexity}, i.e., if and how hierarchical search spaces as introduced in Section~\ref{sec:preproc} can be handled.

\subsubsection{Grid Search and Random Search}  
\label{sssec:grid_random_search}
\begin{figure}[htb]
    \centering
    \includegraphics[width = 0.9\textwidth]{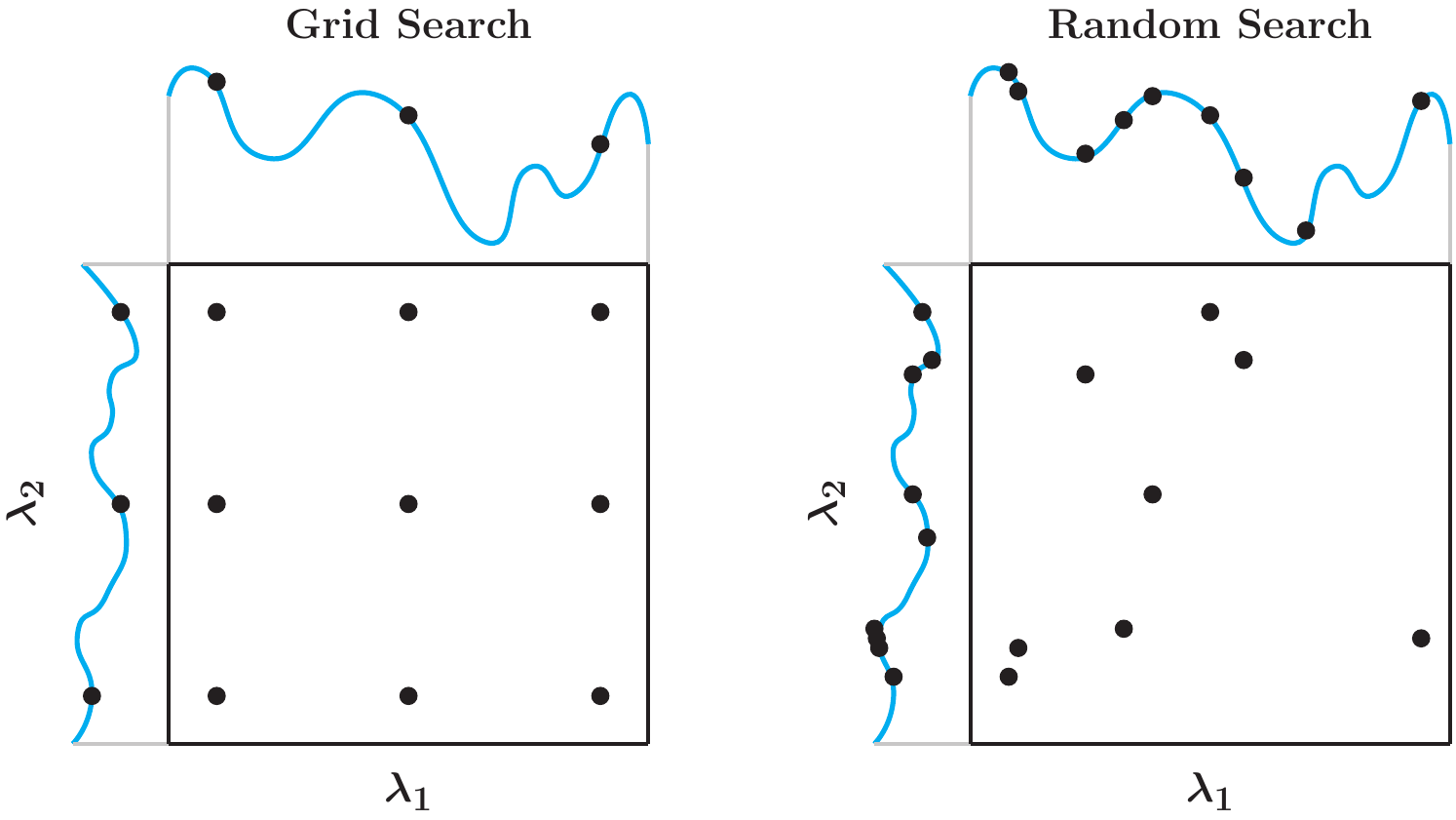}
    \caption{RS and GS where only HP $\lambda_1$ has a strong influence on $c$ (figure based on \citet{bergstra_jmlr12a}).}
    \label{fig:rd-grid-search}
\end{figure}

Grid search (GS) is the process of discretizing the range of each HP and exhaustively evaluating every combination of values. Numeric and integer HP values are usually equidistantly spaced in their box constraints.
The number of distinct values per HP is called the \textit{resolution} of the grid.
For categorical HPs, either a subset or all possible values are considered. 
A second simple HPO algorithm is \emph{random search} (RS).
In its simplest form, values for each HP are drawn independently of each other and from a pre-specified (often uniform) distribution, which works for (box-constrained) numeric, integer, or categorical parameters (c.f.\ Figure~\ref{fig:rd-grid-search}). Due to their simplicity, both GS and RS can handle hierarchical search spaces.

RS often has much better performance than GS in higher-dimensional HPO settings \citep{bergstra12}.
GS suffers directly from the \textit{curse of dimensionality} \citep{bellman2015}, as the required number of evaluations increases exponentially with the number of HPs for a fixed grid resolution. 
This seems to be true as well for RS at first glance, and we certainly require an exponential number of points in $\dim(\LamS)$ to cover the space well. 
However, in practice, HPO problems often have \textit{low effective dimensionality} \citep{bergstra12}: 
The set of HPs that have an influence on performance is often a small subset of all available HPs.
Consider the example illustrated in Figure~\ref{fig:rd-grid-search}, where an HPO problem with HPs $\lambda_1$ and $\lambda_2$ is shown. A GS with resolution $3$ resulting in $9$ HPCs is evaluated, and we discover that only HP $\lambda_1$ has any relevant influence on the performance,
so only 3 of 9 evaluations provided any meaningful information.
%All evaluations of different configurations of $\lambda_2$ have been wasted, i.e., $6$ of $9$ evaluations.
In comparison, RS would have given us 9 different configurations for HP $\lambda_1$, which results in a higher chance of finding the optimum. 
Another advantage of RS is that it can easily be extended by further samples; in contrast, the number of points on a grid must be specified beforehand, and refining the resolution of GS afterwards is more complicated.
Altogether, this makes RS preferable to GS and a surprisingly strong baseline for HPO in many practical settings. 
%Both GS and RS are widely used and implemented in nearly all ML software frameworks. 
Notably, there are sampling methods that attempt to cover the search space more evenly than the uniform sampling of RS, e.g., 
\emph{Latin Hypercube Sampling} \citep{mckay1979}, or Sobol sequences \citep{antonov1979}. 
However, these do not seem to significantly outperform naive i.i.d.\ sampling \citep{bergstra12}.

\subsubsection{Evolution Strategies}
\label{sssec:evolutionary_strategies}

\begin{figure}[htb]
    \centering
    \includegraphics[width=1.0\textwidth]{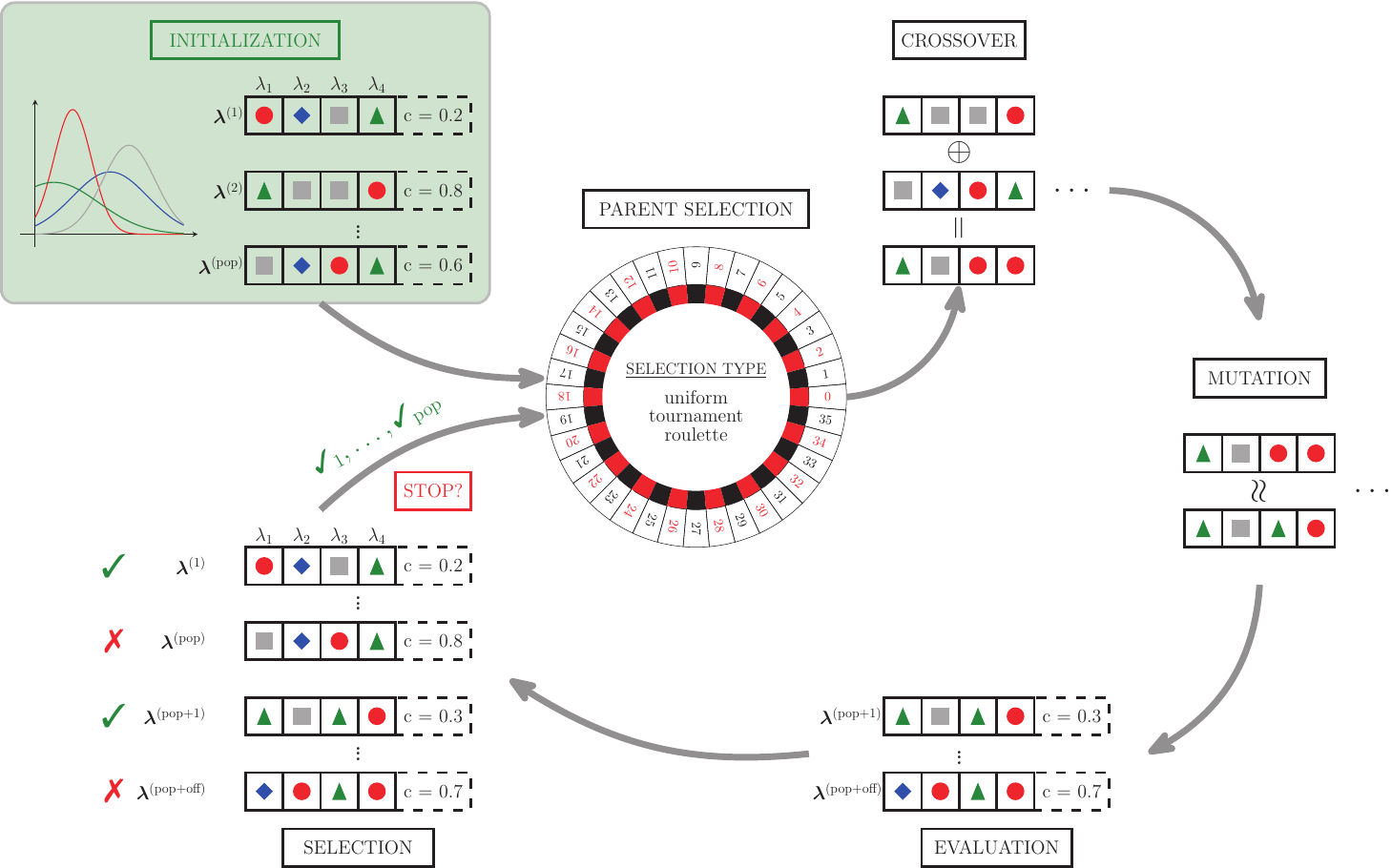}
    \caption{%
        Schematic representation of a single iteration of an ES as a four dimensional discrete problem.
        Parameter values are symbolized by geometric shapes.
    }
    \label{fig:ea}
\end{figure}

Evolution strategies (ES) are a class of stochastic population-based optimization methods inspired by the concepts of biological evolution, belonging to the larger class of \textit{evolutionary algorithms}.
They do not require gradients, making them generally applicable in black-box settings such as HPO.
In ES terminology, an \textit{individual} is a single HPC, the \textit{population} is a currently maintained set of HPCs, and the \textit{fitness} of an individual is its (inverted) generalization error $c(\lambdav)$.
\textit{Mutation} is the (randomized) change of one or a few HP values in a configuration.
\textit{Crossover} creates a new HPC by (randomly) mixing the values of two other configurations.
An ES follows iterative steps to find individuals with high fitness values (c.f.\ Figure~\ref{fig:ea}):
\begin{enumerate*}[label=(\roman*)]
    \item An initial population is sampled at random.
    \item The fitness of each individual is evaluated.
    \item A set of individuals is selected as parents for reproduction.\footnote{Either completely at random, or with a probability according to their fitness, the most popular variants being roulette wheel and tournament selection.}
    \item The population is enlarged through \textit{crossover} and \textit{mutation} of the parents.
    \item The offspring is evaluated.
    \item The \textit{top-k} fittest individuals are selected.\footnote{In ES-terminology this is called a ($\mu+\lambda$) elite survival selection, where $\mu$ denotes the population size, and $\lambda$ the number of offspring; other variants like ($\mu,\lambda$) selection exist.}
    \item Steps (ii) to (v) are repeated until a termination condition is reached.
\end{enumerate*}
For a more comprehensive introduction to ES, see \citet{beyerEvolution2002}.

ES were limited to numeric spaces in their original formulation, but they 
can easily be extended to handle mixed spaces by treating components of different types independently, e.g., by adding a normally distributed random value to real-valued HPs while adding the difference of two geometrically distributed values to integer-valued HPs \citep{li2013mixed}. 
By defining mutation and crossover operations that operate on tree structures or graphs, it is even possible to perform optimization of preprocessing pipelines \citep{OlsonGECCO2016, escalante2009particle} or neural network architectures \citep{real_regularized_2019} using evolutionary algorithms.
The properties of ES can be summarized as follows:
ES have a low likelihood to get stuck in local minima, especially if so-called nested ES are used \citep{beyerEvolution2002}. They can be straightforwardly modified to be robust to noise \citep{beyerRobust2006}, and can also be easily extended to multi-objective settings \citep{coello2007evolutionary}.
Additionally, ES can be applied in settings with complex search spaces and can therefore work with spaces where other optimizers may fail \citep{he2021automl}.
ES are more efficient than RS and GS but still often require a large number of iterations to find good solutions, which makes them unsatisfactory for expensive optimization settings like HPO.

\subsubsection{Bayesian Optimization} 
\label{sssec:model_based_optimization}
Bayesian optimization (BO) has become increasingly popular as a global optimization technique for expensive black-box functions, and specifically for HPO \citep{jones1998efficient,hutter2011sequential,snoek_practical_2012}.

BO is an iterative algorithm whose key strategy is to model the mapping $\lambdav~\mapsto~\clam$ based on observed performance values found in the archive $\archive$ via (non-linear) regression. This approximating model is called a \textit{surrogate model}, for which a Gaussian process or a random forest are typically used. 
BO starts on an archive $\archive$ filled with evaluated configurations, typically sampled randomly, using Latin Hypercube Sampling or the Sobol sampling \citep{bossek_initial_2020}.
%
%This archive is typically generated from an initial design configuration, typically sampled randomly, or using Latin Hypercube Sampling or values derived from the Sobol sequence \citep{bossek_initial_2020}, which are in turn evaluated on $\GEhresa$.
BO then uses the archive to fit the surrogate model, which for each $\lambdav$ produces both an estimate of performance $\chlam$ as well as an estimate of prediction uncertainty $\shlam$, which then gives rise to a predictive distribution for one test HPC or a joint distribution for a set of HPCs. 
Based on the predictive distribution, BO establishes a cheap-to-evaluate acquisition function $\ulam$ that encodes a trade-off between \textit{exploitation} and \textit{exploration}: The former means that the surrogate model predicts a good, low $c$ value for a candidate HPC $\lambdav$, while the latter implies that the surrogate is very uncertain about $\clam$, likely because the surrounding area has not been explored thoroughly.

Instead of working on the true expensive objective, the acquisition function $\ulam$ is then optimized in order to generate a new candidate $\lambdav^+$ for evaluation. The optimization problem $\ulam$
inherits most characteristics from $\clam$; so it is often still multi-modal and defined on a mixed, hierarchical search space. Therefore, $\ulam$ may still be quite complex, but it is at least cheap to evaluate.
This allows the usage of more budget-demanding optimizers on the acquisition function.
If the space is real-valued and the combination of surrogate model and acquisition function supports it, even gradient information can be used. 

Among the possible optimization methods are: iterated local search (as used by \citet{HutterHLS09}), evolutionary algorithms (as in \citet{white2019bananas}), ES using derivatives (as used by \citet{sekhon1998genetic,roustant2012dicekriging}), and a focusing RS called \textit{DIRECT} \citep{jones2009direct}. 

The true objective value $c(\lambdav^+)$ of the proposed HPC $\lambdav^+$ -- generated by optimization of $\ulam$ -- is finally evaluated and added to the archive $\archive$. The surrogate model is updated, and BO iterates until a predefined budget is exhausted, or a different termination criterion is reached. These steps are summarized in Algorithm~\ref{alg:bo}.
BO methods can use different ways of deciding which $\lambdav$ to return, referred to as the \emph{identification step} by \citet{jalali2017comparison}. This can either be the best observed $\lambdav$ during optimization, the best (mean, or quantile) predicted $\lambdav$ from the archive according to the surrogate model \citep{picheny2013benchmark,jalali2017comparison}, or the best predicted $\lambdav$ overall \citep{scott2011correlated}.
The latter options serve as a way of smoothing the observed performance values and reducing the influence of noise on the choice of $\lamh$.

\begin{algorithm}[ht]
  \caption{BO for a black-box objective $c(\lambdav)$.}
  \label{alg:bo}
  Generate $\lambdav^{(1)},\dots,\lambdav^{(k)}$ with sampling scheme or fixed design \\
  Initialize archive  $\archivet[0] = ((\lambdav^{(1)}, c(\lambdav^{(1)})),\dots,(\lambdav^{(k)}, c(\lambdav^{(k)})))$  \\
  \For{$\text{t}=1,2,3,\ldots$ until termination}{%,t_{\text{max}}$}{
    {\bf 1:} Fit surrogate model $(\chlam, \shlam)$ on $\archivet[t-1]$ \\
    {\bf 2:} Build acquisition function $u(\lambdav)$ from $(\chlam, \shlam)$\\
    {\bf 3:} Obtain proposal $\lamp$ by optimizing $u$: $\lambdav^+ \in \argmax_{\lambdav \in \LamS} u(\lambdav)$\\
    {\bf 4:} Evaluate $c(\lambdav^+)$ \\
    {\bf 5:} Obtain $\archivet$ by augmenting $\archivet[t-1]$ with $(\lambdav^+, \clamp)$
 }
 \KwResult{$\lamh$: Best-performing $\lambdav$ from archive or according to surrogates prediction.}
 \vspace{.5cm}
\end{algorithm}
 \vspace{.5cm}
 
\begin{figure}[b!]
    \centering
    \includegraphics[width = 1.0\textwidth]{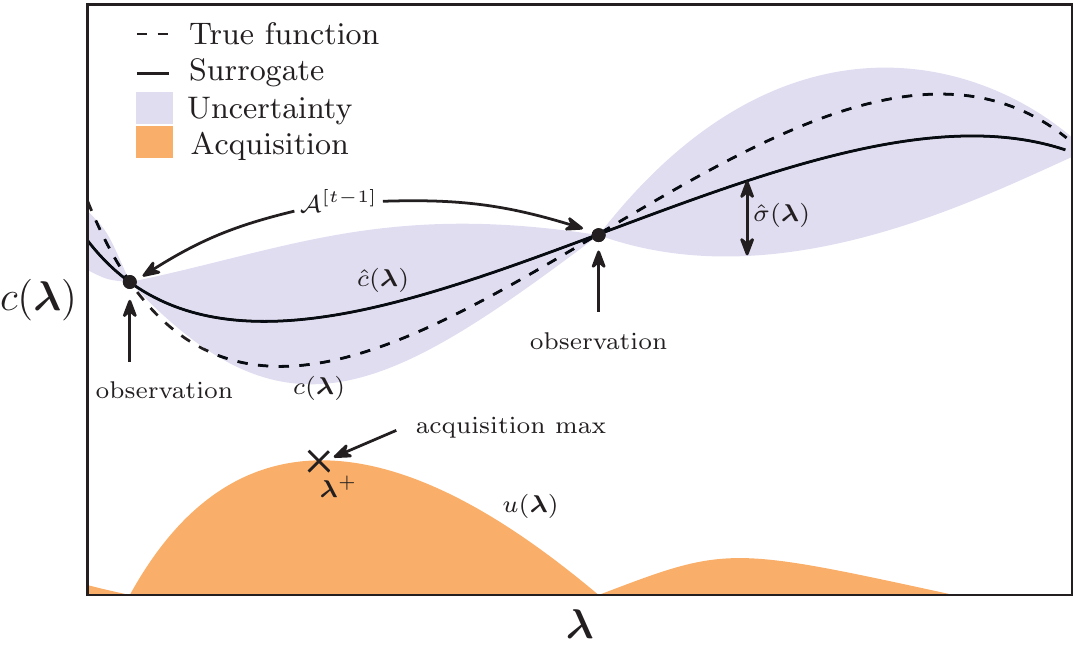}
    \caption{Illustration of how BO generates the proposal by maximizing an acquisition function (figure inspired by \citet{hutter_book19a}).}
    \label{fig:bo}
\end{figure}

\paragraph{Surrogate model}
The choice of surrogate model has great influence on BO performance and is often linked to properties of $\LamS$. 
If $\LamS$ is purely real-valued, Gaussian process (GP) regression \citep{ki2006gaussian} -- sometimes referred to as Kriging -- is used most often.
In its basic form, BO with a GP does not support HPO with non-numeric or conditional HPs, and tends to show deteriorating performance when $\LamS$ has more than roughly ten dimensions. Dealing with integer-valued or categorical HPs requires special care \citep{garrido2020dealing}.
Extensions for mixed-hierarchical spaces that are based on special kernels \citep{swersky2014raiders} exist, and the use of random embeddings has been suggested for high-dimensional spaces \citep{wang2016bayesian,nayebi2019}.
Most importantly, standard GPs have runtime complexity that is cubic in the number of samples, which can result in a significant overhead when the archive $\archive$ becomes large.

\citet{mcintire_sparse_2016} propose to use an adapted, sparse GP that restrains training data from uninteresting areas.
Local Bayesian optimization \citep{eriksson2019scalable} is implemented in the TuRBO algorithm and has been successfully applied to various black-box problems. 

Random forests, most notably used in SMAC \citep{hutter2011sequential}, have also shown good performance as surrogate models for BO. Their advantage is their native ability to handle discrete HPs and, with minor modifications, e.g.,\ in \citet{hutter2011sequential}, even dependent HPs without the need for preprocessing. Standard random forest implementations are still able to handle dependent HPs by treating infeasible HP values as missing and performing imputation.
Random forests tend to work well with larger archives and introduce less overhead than GPs. SMAC 
uses the standard deviation of tree predictions as a heuristic uncertainty estimate $\shlam$ \citep{hutter2011sequential}. However, 
more sophisticated alternatives exist to provide unbiased estimates \citep{sexton2009standard}.
Since trees are not distance-based spatial models, the uncertainty estimator does not increase the further we extrapolate away from observed training points.  
This might be one explanation as to why tree-based surrogates are outperformed by GP regression on purely numerical search spaces \citep{eggensperger2013towards}.

Neural networks (NNs) have shown good performance in particular with nontrivial input spaces, and they are thus increasingly considered as surrogate models for BO \citep{snoek2015scalable}.
Discrete inputs can be handled by one-hot encoding or by automatic techniques, e.g., entity embedding where a dense representation is learned from the output of a simple, direct encoding, such as one-hot encoding by the NN. \citep{Hancock2020}. 
NNs offer efficient and versatile implementations that allow the use of gradients for more efficient optimization of the acquisition function. Uncertainty bounds on the predictions can be obtained, for example, by using Bayesian neural networks (BNNs), which combine NNs with a probabilistic model of the network weights or adaptive basis regression where only a Bayesian linear regressor is added to the last layer of the NN \citep{snoek2015scalable}. 

\paragraph{Acquisition function}
The acquisition function balances out the surrogate model's prediction $\chlam$ and its posterior uncertainty $\shlam$ to ensure both exploration of unexplored regions of $\LamS$, as well as exploitation of regions that have performed well in previous evaluations.
%
% Based on our experience, we recommend setting $\kappa = 2$ if at least one HP is discrete, and $\kappa = 1$ in a fully numeric setting. 
%%%
%
A very popular acquisition function is the \emph{expected improvement} (EI) \citep{jones1998efficient}:
\begin{align}
    \label{eq:EI}
    \begin{split}
    % u_{\operatorname{EI}}(\lambdav) & = \E(\max\left\{c_{\min} - C(\lambdav), 0\right\})
    u_{\operatorname{EI}}(\lambdav) & = \Eop_{C \sim \mathcal{N}\left(\chlam,\, \shlam^2\right)}\left[\max\left\{c_{\min} - C(\lambdav), 0\right\}\right]\\
    & = \left( c_{\min} - \chlam \right) \Phi \left( \frac{c_{\min} - \chlam}{\shlam} \right) + \shlam \phi \left( \frac{c_{\min} - \chlam}{\shlam} \right) \ ,    
    \end{split}
\end{align}
where $c_{\min}$ denotes the best observed outcome of $c$ so far, 
and $\Phi$ and $\phi$ are the cumulative distribution function and density of the standard normal distribution, respectively. 
The EI was introduced in connection with GPs that have a Bayesian interpretation, expressing the posterior distribution of the true performance value given already observed values as a Gaussian random variable $C(\lambdav)$ with $C(\lambdav) \sim \mathcal{N}(\chlam, \shlam^2)$. Under this condition, Eq.~$\eqref{eq:EI}$ can be analytically expressed as above, and the resulting formula is often heuristically applied to other surrogates that supply $\chlam$ and $\shlam$.

A further, very simple acquisition function is the \emph{lower confidence bound} (LCB) \citep{jones2001taxonomy}:
\begin{align}
    \label{eq:LCB}
    u_{\operatorname{LCB}}\left(\lambdav\right) = (-1) \cdot \left(\chlam -\kappa \cdot \shlam\right),
\end{align}
here negated to yield a maximization problem for Algorithm~$\ref{alg:bo}$.
The LCB treats local uncertainty as an additive bonus at each $\lambdav$ to enforce exploration, with $\kappa$ being a control parameter that is not easy to set. 

\paragraph{Adaptive Explore-Exploit Tradeoffs}
In a theoretical analysis without a limit on evaluations, \citet{srinivas_gaussian_2010} suggest to increase the $\kappa$-parameter of LCB over time to encourage exploration in later phases of optimization.
% Janek: An optimal dynamic choice of kappa can also be found in "Optimization as estimation with Gaussian processes in bandit settings. In International Conference on Artificial Intelligence and Statistics (AISTATS), 2016.". There is also a nice connection to max-value entropy search, but this might get a bit too much for this paper
However, in the context of practical HPO with a finite budget for evaluations, it seems plausible to \textit{decrease} $\kappa$ over time to enforce exploration in the beginning and exploitation at the end in a similar cooldown scheme as in simulated annealing, such as e.g.\ suggested by \citet{zheng2016parameterized}, which uses a further cyclical scheme to escape local minima. 
\citet{sasena2002exploration} propose a cooldown scheme for expected improvement: They use the ``generalized expected improvement'' $u_{\operatorname{GEI}}(\lambdav) = \Eop\left[\max\left\{c_{\min} - C(\lambdav), 0\right\}^g\right]$, where larger values for exponent $g$ also enforce more exploration.
They suggest to start with a large $g$ value in the beginning and to gradually decrease it to enforce exploitation at the end.
%over the course of the optimization, which lowers the influence of uncertainty towards the end.
\citet{jasrasaria2018dynamic} propose to dynamically give exploitation more weight as a function of mean model-uncertainty in what they call ``contextual improvement''. This has a similar effect of encouraging late exploitation, as model uncertainty generally decreases over the course of optimization.
Finally, \citet{de2021greed} show that a very simply $\epsilon$-greedy BO strategy can perform well or even better than established acquisition functions. They simply propose 
HPCs that maximize  the predicted mean, but interleave random HPCs with $\epsilon$ probability.
They show that this strategy performs well in settings with a low evaluation budget or with many dimensions, which is consistent with the other proposed methods, which adaptively emphasize exploitation more when remaining budget is low.

\paragraph{Multi-point proposal}
In its original formulation, BO only proposes one candidate HPC per iteration and then waits for the performance evaluation of that configuration to conclude. 
However, in many situations, it is preferable to evaluate multiple HPCs in parallel by proposing multiple configurations at once, or by asynchronously proposing HPCs while other proposals are still being evaluated.

While in the sequential variant, the best point can be determined unambiguously from the full information of the acquisition function. In the parallel variant, many points must be proposed at the same time without information about how the other points will perform. The objective here is to some degree to ensure that the proposed points are sufficiently different from each other.

The proposal of $n_{\text{batch}}>1$ configurations in one BO iteration is called \emph{batch proposal} or \emph{synchronous parallelization} and works well if the runtimes of all black-box evaluations are somewhat homogeneous.
If the runtimes are heterogeneous, one may seek to spontaneously generate new proposals whenever an evaluation thread finishes in what is called \emph{asynchronous parallelization}. This offers some advantages to synchronous parallelization, but is more complicated to implement in practice.

The simplest option to obtain $n_{\text{batch}}$ proposals is to use the LCB criterion in Eq.~(\ref{eq:LCB}) with different values for $\kappa$.
For this so-called qLCB (also referred to as qUCB) approach, \citet{hutter_parallel_2012} propose to draw $\kappa$ from an exponential distribution with rate parameter $1$. This can work relatively well in practice but has the potential drawback of creating proposals that are too similar to each other \citep{bischl_moimbo_2014}. 
\citet{bischl_moimbo_2014} instead propose to maximize both $\chlam$ and $\shlam$ simultaneously, using multi-objective optimization, and to choose $n_{\text{batch}}$ points from the approximated Pareto-front.
Further ways to obtain $n_{\text{batch}}$ proposals are constant liar, Kriging believer (both described in \citealp{ginsbourger_kriging_2010}), and q-EI \citep{chevalier_fast_2013}.
Constant liar sets fake constant response values for the first points proposed in the batch to generate additional one via the normal EI principle and the approach; Kriging believer does the same but uses the GP model's mean prediction as fake value instead of a constant.
The qEI optimizes a true multivariate EI criterion and is computationally expensive for larger batch sizes, but \citet{balandat2020botorch} implement methods to efficiently calculate the qEI (and qNEI for noisy observations) through MC simulations.

\paragraph{Efficient Performance Evaluation}
While BO models only optimize the HPC prediction performance in its standard setup, there are several extensions that aim to make optimization more efficient by considering runtime or resource usage.
These extensions mainly modify the acquisition function to influence the HPCs that are being proposed. \citet{snoek_practical_2012} suggests the \textit{expected improvement per second} (EIPS) as a new acquisition function.
The EIPS includes a second surrogate model that predicts the runtime of evaluating a HPC in order to compromise between expected improvement and required runtime for evaluation.
Most methods that trade off 
between runtime and information gain fall under the category of multi-fidelity methods, which is further discussed in Section~\ref{sssec:hyperband}. Acquisition functions that are especially relevant here consider information gain-based criteria like \emph{Entropy Search} \citep{henning2012entropy} or \emph{Predictive Entropy Search} \citep{hernandez2016predictive}. These acquisition functions can be used for selective subsample evaluation \citep{klein2017fast}, reducing the number of necessary resampling iterations \citep{swersky2013multi}, and stopping certain model classes, such as NNs, early. %

\subsubsection{Multifidelity and Hyperband}
\label{sssec:hyperband}

The multifidelity (MF) concept in HPO refers to all tuning approaches that can efficiently handle a learner $\inducer(\D, \lambdav)$ with a fidelity HP $\lamfid$ as a component of 
$\lambdav$, which influences the computational cost of the fitting procedure in a monotonically increasing manner. Higher $\lamfid$ values imply a longer runtime of the fit. 
This directly implies that the lower we set $\lamfid$, the more points we can explore in our search space, albeit with much less reliable information w.r.t.\ their true performance. 
If $\lamfid$ has a linear relationship with the true computational costs, we can directly sum the $\lamfid$ values for all evaluations to measure the computational costs of a complete optimization run.
We assume to know box-constraints of $\lamfid$ in form of a lower and upper limit, so $\lamfid \in [\lamfidl, \lamfidu]$, where the upper limit implies the highest fidelity returning values closest to the true objective value at the highest computational cost. 
Usually, we expect higher values of $\lamfid$ to be better in terms of predictive performance yet naturally more computationally expensive. However, overfitting can occur at some point, for example when $\lamfid$ controls the number of training epochs when fitting an NN.
Furthermore, we assume that the relationship of the fidelity to the prediction performance changes somewhat smoothly. Consequently, when 
evaluating multiple HPCs with small $\lamfid$, this at least indicates their true ranking.
Typically, this implies a sequential fitting procedure, where $\lamfid$ is, for example, the number of (stochastic) gradient descent steps or the number of sequentially added (boosting) ensemble members. 
A further, generally applicable option is to subsample the training data from a small fraction to 100\% before training and to treat this as a fidelity control \citep{klein2017fast}.
HPO algorithms that exploit such a $\lamfid$ parameter -- usually by spending budget on cheap HPCs with low $\lamfid$ values earlier for exploration, and then concentrating on the most promising ones later -- are called \emph{multifidelity methods}. 
One can define two versions of the MF-HPO problem.
(a) If overfitting can occur with higher values of $\lamfid$ (e.g.,\ if it encodes training iterations), simply minimizing $\min_{\lambdav \in \LamS} \clam$ is already appropriate.
(b) If the assumption holds that a higher fidelity always results in a better model (e.g.,\ if $\lamfid$ controls the size of the training set), we are interested in finding the configuration $\lams$ for which the inducer will return the best model given the full budget, so $\min_{\lambdav \in \LamS, \lamfid= \lamfidu} \clam$.
Of course, in both versions, the optimizer can make use of cheap HPCs with low settings of $\lamfid$ on its path to its result. 

Hyperband \citep{li_2018} can best be understood as repeated execution of the \emph{successive halving} (SH) procedure \citep{Jamieson2016NonstochasticBA}.
SH assumes a fidelity-budget $B$ for the sum of $\lamfid$ for all evaluations. It starts with a given, fixed number of candidates $\lami$ that we denote with $p^{[0]}$ and \enquote{races them down} in stages $t$ to a single best candidate by repeatedly evaluating all candidates with increased fidelity in a certain schedule. 
Typically, this is controlled by the $\etahb$ control multiplier of Hyperband with $\etahb > 1$ (typically set to 2 or 3): After each batch evaluation $t$ of the current population of size $p^{[t]}$, we reduce the population to the best $\frac{1}{\etahb}$ fraction and set the new fidelity for a candidate evaluation to $\etahb \times \lamfid$. Thus, promising HPCs are assigned a higher fidelity overall, and sub-optimal ones are discarded early on.
The starting fidelity $\lamfid^{[0]}$ and the number of stages $s+1$ are computed in a way such that each batch evaluation of an SH population has approximately $B/(s+1)$ amount of fidelity units spent. Overall, this ensures that approximately, but not more than, $B$ fidelity units are spent in SH:

\begin{equation}
\label{eq:hbbudget}
    \sum^{s}_{t=0} \left\lfloor p^{[0]}\etahb^{-t}\right\rfloor \lamfid^{[0]}\etahb^{t} \leq B.
\end{equation}

\begin{figure}[htb]
    \centering
    \includegraphics[width = 1.0\textwidth]{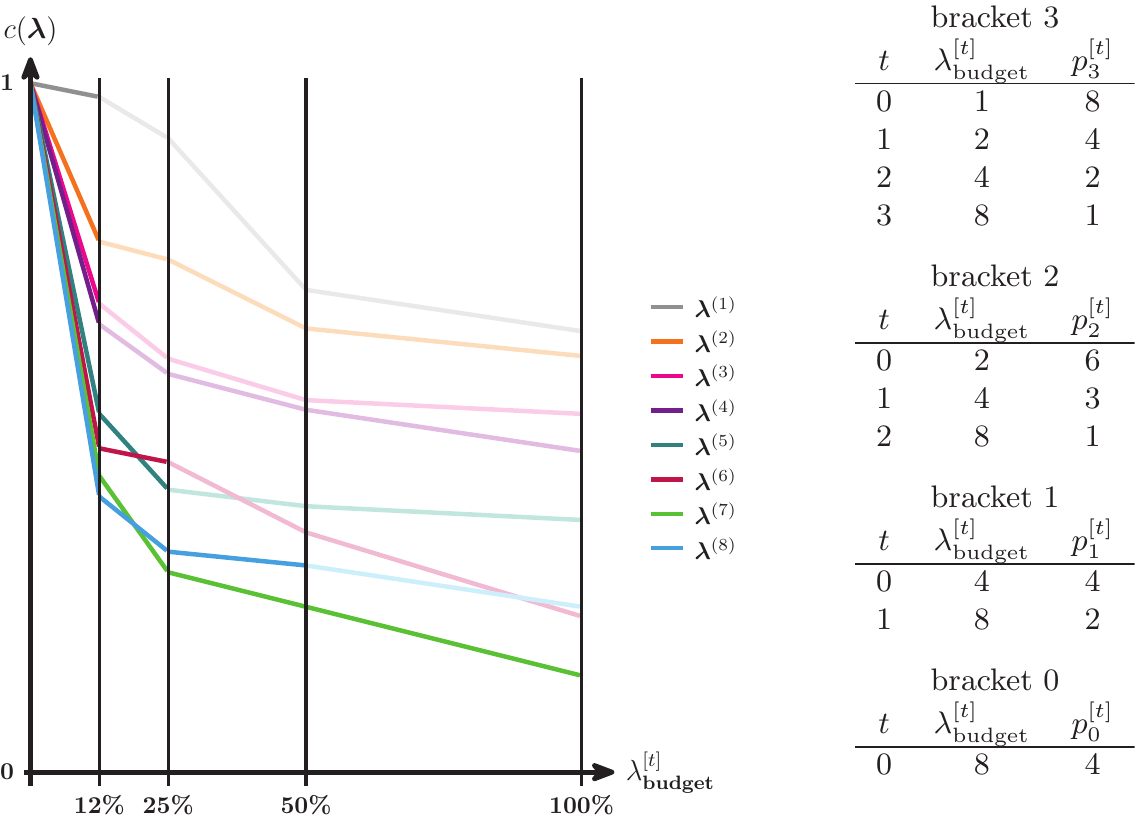}
    \caption{Right: Bracket design of HB with $\lamfidu = 8$ and $\etahb = 2$ (resulting in four brackets). Left: Exemplary bracket run (figure inspired by \citet{hutter_book19a}). Faint lines represent future performance of HPCs that were discarded early.
    %that would have been observed, had the evaluation continued.
    }
    \label{fig:successive_halving} 
\end{figure}

However, the efficiency of SH strongly depends on a sensible choice of the number of starting configurations and the resulting schedule.
If we assume a fixed fidelity-budget for HPO, the user has the choice of running either (a) more configurations but with less fidelity, or (b) fewer configurations, but with higher fidelity. While the former naturally explores more, the latter schedules evaluations with stronger correlation to the true objective value and more informative evaluations. As an example, consider how $\lambda^{(6)}$ is discarded in favor of $\lambda^{(8)}$ at 25\% in Figure~\ref{fig:successive_halving}. Because their performance lines would have crossed close to 100\%, $\lambda^{(6)}$ is ultimately the better configuration. However, in this case, the superiority of $\lambda^{(6)}$ was only observable after full evaluation. 
%Thus, the correct trade-off between how many configurations to evaluate and how aggressively to discard bad performers early depends on the correlation between the smaller budgets and the larger budgets.
As we often have no prior knowledge regarding this effect, HB simply runs SH for different numbers of starting configurations $p_s^{[0]}$, and each SH run or schedule is called a \emph{bracket}. As input, HB takes $\etahb$ and the maximum fidelity $\lamfidu > \etahb$. HB then constructs the target fidelity budget $B$ for each bracket by considering the most explorative bracket: Here, the number of batch evaluations $s_{\text{max}} + 1$ is chosen to be $\left\lfloor\log_{\etahb}(\lamfidu)\right\rfloor + 1$ for which $\lamfid^{[0]} = \lamfidu\etahb^{-s_{\text{max}}} \in (\etahb^{-1}, \etahb),\, \lamfid^{[s_{\text{max}}]} = \lamfidu$, and we collect these values in $\bm{r} = (\lamfidu\etahb^{-s_{\text{max}}}, \lamfidu\etahb^{-s_{\text{max}}+1},\lamfidu\etahb^{-s_{\text{max}}+2}, \dots, \lamfidu) \in \R^{s_{max}+1}$. 
Since we want to spend approximately the same total fidelity and reduce the candidates to one winning HPC in every batch evaluation, the fidelity budget of each bracket is $B = (s_{\text{max}}+1)\lamfidu$. 
%This $B$ is used for all brackets. 
For every $s \in \{0, \dots, s_{\text{max}}\}$, a bracket is defined by setting the starting fidelity $\lamfid^{[0]} \ge \lamfidl$ of the bracket to $\bm{r}^{(1 + s_{\text{max}}-s)}$, resulting in $s_{\text{max}} + 1$ brackets and an overall fidelity budget of $(s_{\text{max}} + 1)B$ spent by HB. Consequently, every bracket $s$ consists of $s + 1$ batch evaluations, and the starting population size $p_s^{[0]}$ is the maximum value that fulfills Eq.~(\ref{eq:hbbudget}). The full algorithm is outlined in Algorithm~\ref{alg:hyperband}, and the bracket design of HB with $\lamfidu = 8$ and $\etahb = 2$ is shown in Figure~\ref{fig:successive_halving}.\\
Starting configurations are usually sampled uniformly, but \cite{li_2018} also show that any stationary sampling distribution is valid. Because HB is a random-sampling-based method, it can trivially handle hierarchical HP spaces in the same manner as RS.

\begin{algorithm}[ht]
\captionsetup{singlelinecheck=off}
  \caption[Hyperband algorithm]{Hyperband algorithm \citep{li_2018} where \\
\noindent
\begin{minipage}[t]{.01\textwidth}
\strut\vspace*{-\baselineskip}\newline
  $\;$
\end{minipage}
\begin{minipage}[t]{.03\textwidth}
\strut\vspace*{-\baselineskip}\newline
  $\bullet$
\end{minipage}
\begin{minipage}[t]{.91\textwidth}
\strut\vspace*{-\baselineskip}\newline
      $\textrm{get\_HPCs}(p)$ uses a stationary sampling distribution to generate the initial HPC population of size $p$,
\end{minipage}
\noindent
\begin{minipage}[t]{.01\textwidth}
\strut\vspace*{-\baselineskip}\newline
  $\;$
\end{minipage}
\begin{minipage}[t]{.03\textwidth}
\strut\vspace*{-\baselineskip}\newline
  $\bullet$
\end{minipage}
%\begin{minipage}[t]{.91\textwidth}
%\strut\vspace*{-\baselineskip}\newline
 %     $\textrm{HPC\_with\_fidelity}(\lambdav, \lamfid)$ returns $\lambdav$ where its fidelity HP is set to $\lamfid$,
%\end{minipage}
% \noindent
% \begin{minipage}[t]{.01\textwidth}
% \strut\vspace*{-\baselineskip}\newline
%   $\;$
% \end{minipage}
% \begin{minipage}[t]{.03\textwidth}
% \strut\vspace*{-\baselineskip}\newline
%   $\bullet$
% \end{minipage}
\begin{minipage}[t]{.91\textwidth}
\strut\vspace*{-\baselineskip}\newline
      $\textrm{top\_k}(\Lambda_s, C, k)$ selects the $k$ HPCs in $\Lambda_s$ associated to the $k$ best performances in $C$ as the next HPC population.
\end{minipage}
}

  \label{alg:hyperband}
  {\bf input:} maximum fidelity per HPC $\lamfidu, \etahb$ \\
  {\bf initialization:} $s_{\text{max}} = \left\lfloor\log_{\etahb}(\lamfidu)\right\rfloor,\; B = (s_{\text{max}}+1)\lamfidu$ \\
  $\quad\quad \bm{r} = (\lamfidu\etahb^{-s_{\text{max}}}, \lamfidu\etahb^{-s_{\text{max}}+1},\lamfidu\etahb^{-s_{\text{max}}+2}, \dots, \lamfidu)$ \\
  \For{$s = s_{\text{max}}, s_{\text{max}}-1, \dots, 0 $}{
    $p_s = \left\lceil\frac{B}{\lamfidu}\frac{\etahb^s}{s+1}\right\rceil$\\  $\Lambda_s^{[0]} = \textrm{get\_HPCs}(p_s) \quad (=\{\lambdav^{(1)}_s, \dots, \lambdav^{(p_s)}_s\}, \, \lambdav^{(i)}_s \in \LamS$) \\
    // Successive Halving inner loop \\
  \For{$t = 0, \dots, s $}{
    $p_s^{[t]} = \left\lfloor p_s\etahb^{-t}\right\rfloor$\\
    Set $\lamfid$ components of entries of $\Lambda_s^{[t]}$ to $\bm{r}^{(1 + s_{\text{max}}-s+t)} \quad (= (\lamfidu\etahb^{-s})\cdot\etahb^{t})$\\
    $C^{[t]} = \{\clam : \lambdav \in \Lambda_s^{[t]}\}$ \\
    $\Lambda_s^{[t+1]} = \textrm{top\_k}(\Lambda_s^{[t]}, C^{[t]}, \left\lfloor p_s^{[t]}/\etahb\right\rfloor)$
 }
 }
 \KwResult{HPC with best performance}
 \vspace{.5cm}
\end{algorithm}
 \vspace{.5cm}

\paragraph{Multifidelity Bayesian Optimization}
The idea behind Hyperband -- trying to discard HPCs that do not perform well early on -- is somewhat orthogonal to the idea behind BO, i.e. intelligently proposing HPCs that are likely to improve performance or to otherwise gain information about the location of the optimum. 
It is therefore natural to combine these two methods. 
This has first been achieved with BOHB by \citet{falkner_icml18a}, who progressively increase $\lamfid$ of suggested HPCs as in Hyperband. 
However, instead of proposing HPCs randomly, they use a model-based approach equivalent to maximizing expected improvement. 
They show that BOHB performs similar to HB in the low-budget regime, where it is superior to normal BO methods, but outperforms HB and perform similar or better to BO when enough budget for tens of full-budget evaluations are available. 
\ifarxiv
% we can't cite abohb in wiley because not peer-reviewed
BOHB was later extended to A-BOHB \citep{tiao2020model} to efficiently perform asynchronously parallelized optimization by sampling possible outcomes of evaluations currently under way.
\fi

Hyperband-based multi-fidelity methods have a control parameter that functions similar to $\etahb$ described above, which determines the fraction of configurations that are discarded at every $\lamfid$ value for which evaluations are performed. However, the optimal proportion of configurations to discard may vary depending on how strong the correlation is between performance values at different fidelities. 
An alternative approach is to use the surrogate model from BO to make adaptive decisions about what $\lamfid$ values to use, or what HPCs to discard. 
Algorithms following this approach typically use a method first proposed by \citet{swersky_nips13a}, who use a surrogate model for both the performance, as well as the resources used to evaluate an HPC. Entropy search \citep{henning2012entropy} is then used to maximize the information gained about the maximum for when $\lamfid=\lamfidu$ per unit of predicted resource expenditure. 
Low-fidelity HPCs are evaluated whenever they contribute disproportionately large amounts of information to the maximum compared to their needed resources. 
A special challenge that needs to be solved by these methods is the modeling of performance with varying $\lamfid$, which often has a different influence than other HPs and is therefore often considered as a separate case.
\ifarxiv
An early HPO method building on this concept is freeze-thaw BO \citep{swersky2014freeze}, which considers optimization of iterative ML methods such as deep learning that can be suspended (``frozen'') and continued (``thawed''). 
\fi
Another HPO method that specifically considers the training set size as fidelity HP is FABOLAS \citep{klein_aistats17}, which actively decides the training set size for each evaluation by trading off computational cost of an evaluation with a lot of data against the information gain on the potential optimal configuration.

In general, there could be other proposal mechanisms instead of random sampling as in Hyperband or BO as in BOHB. For example, \citet{AwadMH21} showed that differential evolution can perform even better; however the evolution of population members across fidelities needs to be adjusted accordingly.  

\subsubsection{Iterated Racing}\label{sssec:racing}  

The iterated racing (IR, \citealp{birattari2010f}) procedure is a general algorithm configuration method that optimizes for a configuration of a general (not necessarily ML) algorithm that performs well over a given distribution of (arbitrary) problems. 
In most HPO algorithms, HPCs are evaluated using a resampling procedure such as CV, so a noisy function (error estimate for single resampling iterations) is evaluated multiple times and averaged.
In order to connect racing to HPO, we now define a problem as a single holdout resampling split for a given ML data set, as suggested in \citet{autoweka,lang_2015_autom}, and we will from now on describe racing only in terms of HPO.

The fundamental idea of \emph{racing} \citep{maron1994hoeffding} is that HPCs that show particularly poor performance when evaluated on the first problem instances (in our case: resampling folds) are unlikely to catch up in later folds and can be discarded early to save computation time for more interesting HPCs.
This is determined by running a (paired) statistical test w.r.t.\ HPC performance values on folds.
This allows for an efficient and dynamic allocation of the number of folds in the computation of $\clam$ -- a property of IR that is unique, at least when compared to the algorithms covered in this article.

Racing is similar to HB in that it discards poorly-performing HPCs early. Like HB, racing must also be combined with a sampling metaheuristic to initialize a race.
Particularly well-suited for HPO are iterated races
%, implemented in R-package \rpkg{irace}
\citep{lopez_2016}, and we will use the terminology of that implementation to explain the main control parameters of IR. 
IR starts by racing down an initial population of randomly sampled HPCs and then uses the surviving HPCs of the race to stochastically initialize the population of the subsequent race to focus on interesting regions of the search space.

\begin{figure}[htb]
    \centering
    \includegraphics[width=1.0\textwidth]{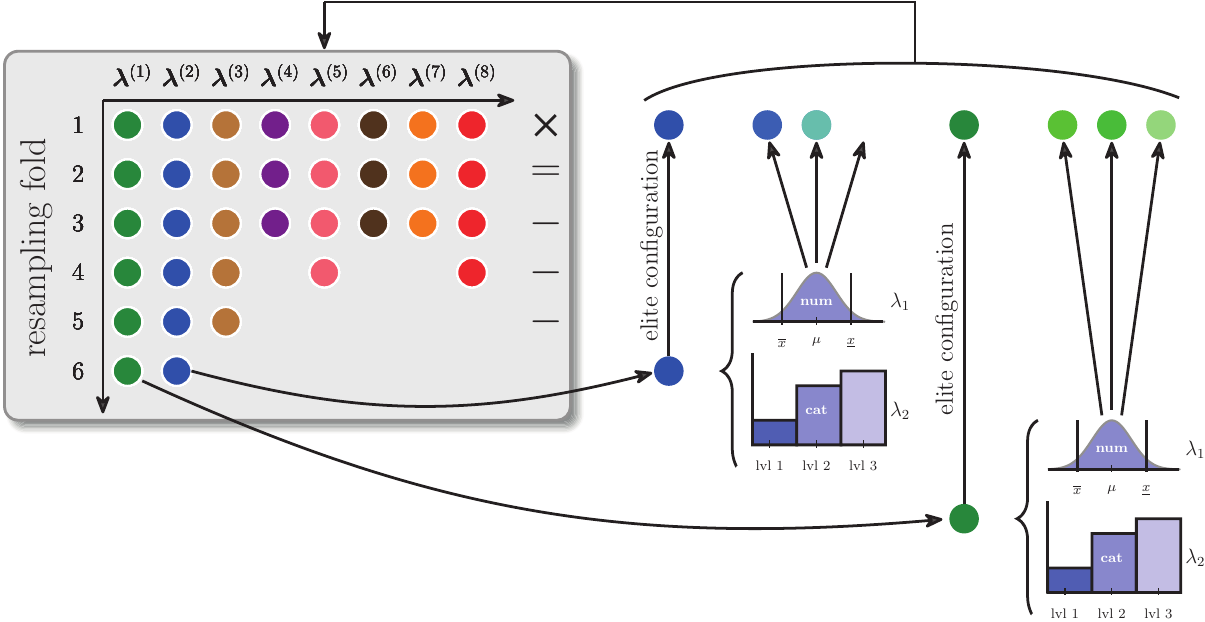}
    \caption{
    Scheme of the iterated racing algorithm (figure based on \citet{lopez_2016}).}
    \label{fig:racing}
\end{figure}

Sampling is performed by first selecting a parent configuration $\lambdav$ among the $N^{\textrm{elite}}$ survivors of the previous generation, according to a categorical distribution with probabilities $p_{\lambdav} = 2 (N^{\textrm{elite}} - r_{\lambdav} + 1) / (N^{\textrm{elite}}(N^{\textrm{elite}}+1))$, where $r_{\lambdav}$ is the rank of the configuration $\lambdav$. 
A new HPC is then generated from this parent by mutating numeric HPs using a truncated normal distribution, always centered at the numeric HP value of the parent.
Discrete parameters use a discrete probability distribution. 
This is visualized in Figure~\ref{fig:racing}.
The parameters of these distributions are updated as the optimization continues: The standard deviation of the Gaussian is narrowed to enforce exploitation and convergence, and the categorical distribution is updated to more strongly favor the values of recent ancestors. 
IR is able to handle search spaces with dependencies by sampling HPCs that were inactive in a parent configuration from the initial (uniform) distribution.
This algorithmic principle of having a distribution that is centered around well-performing candidates, is continuously sampled from and updated, is close to an estimation-of-distribution algorithm (EDA), a well-known template for ES \citep{larranaga2001estimation}. 
Therefore, IR could be described as an EDA with racing for noise handling. 

IR has several control parameters that determine how the racing experiments are executed. 
We only describe the most important ones here; many of these have heuristic defaults set in the implementation introduced by \citet{lopez_2016}. 
%The budget parameter $B$ (\texttt{maxExperiments} in \rpkg{irace}) determines how many performance evaluations are to be executed in total, which coincides with the number of HPC evaluations on single holdout folds. 
$N^\textrm{iter}$ (\texttt{nbIterations}) determines the number of performed races, defaulting to $\lfloor 2+\log_2 \dim(\LamS)\rfloor$ (with $\dim(\LamS)$ the number of HPs being optimized).
Within a race, each HPC is first evaluated on $T^\textrm{first}$ (\texttt{firstTest}) folds before a first comparison test is made. Subsequent tests are then made after every $T^\textrm{each}$ (\texttt{eachTest}) evaluation. 
IR can be performed as \emph{elitist}, which means surviving configurations from a generation are part of the next generation. 
%BB: komischer satz hier
% Otherwise, survivors are discarded. 
The statistical test that discards individuals can be the Friedman test or the $t$-test \citep{birattari2002racing}, the latter possibly with multiple testing correction. 
\citet{lopez_2016} recommend the $t$-test when when performance values for evaluations on different instances are commensurable and the tuning objective is the mean over instances, which is usually the case for our resampled performance metric where instances are simply resampling folds.
%or when the tuning objective is an order statistic. 
%The $t$-test is recommended if the objective is the mean of a cost function.

\subsection{HPO - A Bilevel Inference Perspective}
\label{ssec:seclevel}
\begin{figure}[htb]
    \centering
    \includegraphics[width = 0.9\textwidth]{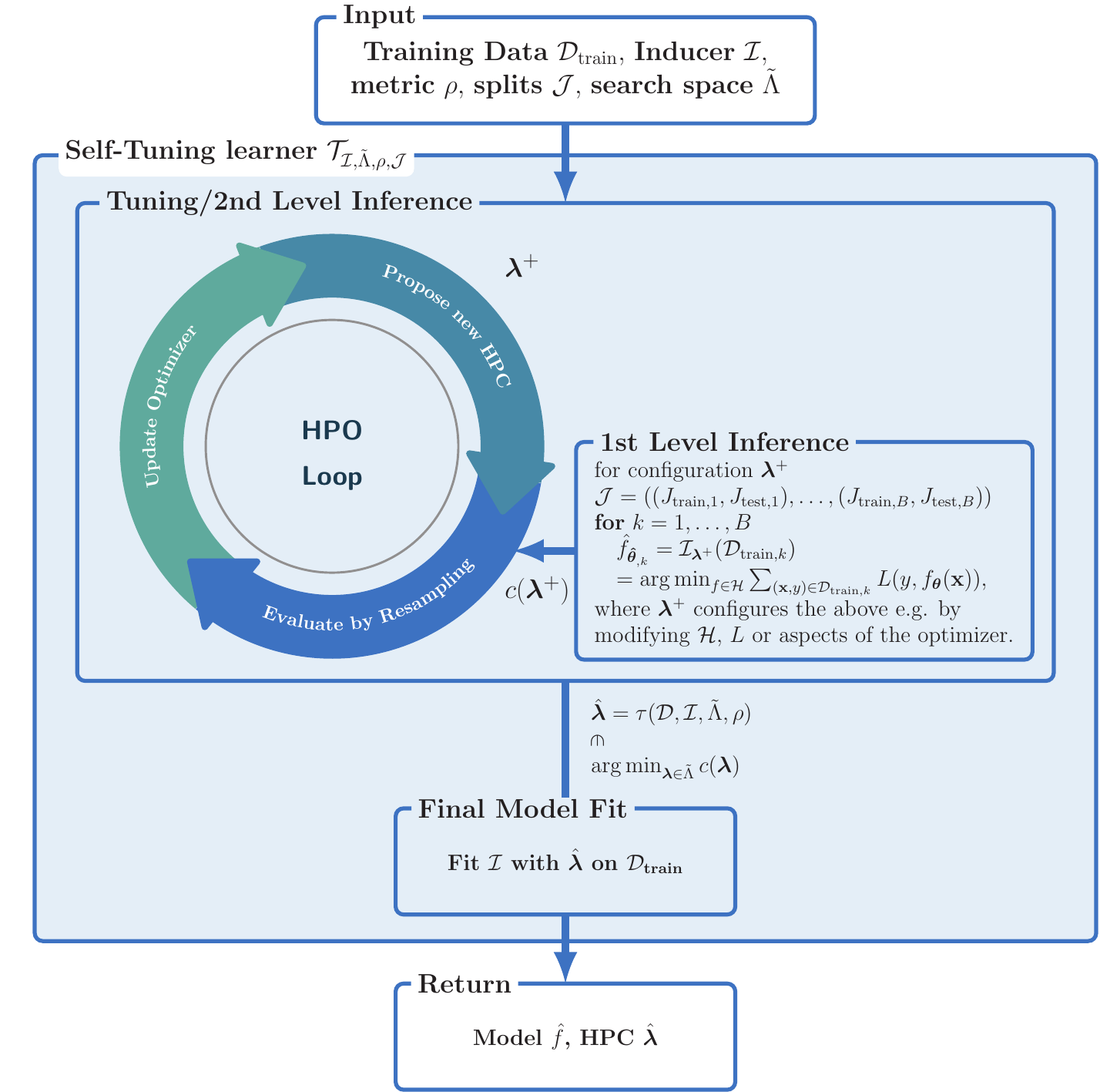}
    \caption{Self-Tuning learner with integrated HPO wrapped around the inducer.}
    \label{fig:learner_autotuned}
\end{figure}

As discussed in Section~\ref{ssec:hpo_definiton}, HPO produces an approximately optimal HPC $\lamh = \tau(\D,\inducer,\LamS,\rho)$ by optimizing it w.r.t.\ the resampled performance $\clam =\GEhresa$.
%, where in each training step in during resampling, .
This is still risk minimization w.r.t.\ (hyper)parameters, where we search for optimal parameters~$\lamh$ so that the risk of our predictor~$\hat{f}_{\lamh, \thetah}$ becomes minimal when measured on validation data via
\begin{equation}
    \rhoL (\yv, \F_{\lamh, \thetah}) = \sumim L\left(\yi, \Fi_{\lamh, \thetah}\right)\textrm{,}
\end{equation}
where $\hat{f}_{\lamh, \thetah} = \inducer(\Dtrain, \lamh)$ and $\F_{\lamh, \thetah}$ is the prediction matrix of $\fh$ on validation data, for a pointwise loss function.
The above is formulated for a single holdout split $(\Jtrain, \Jtest)$ in order to demonstrate the tight connection between (first level) risk minimization and HPO; Eq.~$\eqref{eq:ges}$ provides the generalization for arbitrary resampling with multiple folds. 
This is somewhat obfuscated and complicated by the fact that we cannot evaluate Eq.~\eqref{eq:ges} in one go, but must rather fit one or multiple models $\fh$ during its computation (hence also its black-box nature).
It is useful to conceptualize this as a bilevel inference mechanism; while the parameters~$\thetah$ of $f$ for a given HPC are estimated in the first level, in the second level we infer the HPs~$\lamh$. 
However, both levels are conceptually very similar in the sense that we are optimizing a risk function for model parameters which should be optimal for the the data distribution at hand. 
In case of the second level, this risk function is not $\risket$, but the harder-to-evaluate generalization error $\GEh$.
An intuitive, alternative term for HPO is \emph{second level inference} \citep{guyon_model_2010}, visualized in Figure~\ref{fig:learner_autotuned}.
 
% %This can also be interpreted as a game: In order to minimize the upper level objective $\GEhresa$ (the resampled generalization error) player \enquote{HPO} makes a move $\lamp$, anticipating that player \enquote{ML} will react with move $\fh$ that minimizes the lower level objective $\riskef$ with $\fh=\inducer(\Dtrain, \lamp)$. 
 
There are mainly two reasons why such a bilevel optimization is preferable to a direct, joint risk minimization of parameters and HPs \citep{guyon_model_2010}: 
\begin{itemize}
    \item Typically, learners are constructed in such a way that optimized first-level parameters can be more efficiently computed for fixed HPs, e.g.,
    often the first-level problem is convex, while the joint problem is not.
    \item Since the generalization error is eventually optimized for the bilevel approach, the resulting model should be less prone to overfitting.
\end{itemize}
Thus, we can define a learner with integrated tuning as a mapping $\tunerfull : \allDatasets \rightarrow \Hspace$, $\D \rightarrow \inducer_{\tau(\D,\inducer,\LamS,\rho)}(\D)$, which maps a data set $\D$ to the model $\fh_{\lamh}$ that has the HPC set to $\lamh$ as optimized by $\tau$ on $\D$ and is then itself trained on the whole of $\D$; all for a given inducer $\inducer$, performance measure $\rho$, and search space $\LamS$. 
%his tuner can be identified as a learner itself, a mapping from data $\Dtrain$ to a model $\fh$, or 
Algorithmically, this learner has a 2-step training procedure (see Figure~\ref{fig:learner_autotuned}), where tuning is performed before the final model fit.
% This learner now has no, or at least fewer, HPs than the original and wrapped learner $\inducer$.
This \enquote{self-tuning} learner $\tuner$ \enquote{shadows} the tuned HPs of its search space $\LamS$ from the original learner and integrates their configuration into the training procedure
\footnote{The self-tuning learner actually also adds new HPs that are the control parameters of the HPO procedure.}. 
If such a learner is cross-validated, we naturally arrive at the concept of \emph{nested CV}, which is discussed in the following Section~{\ref{ssec:nested}}.

\subsection{Nested Resampling and Meta-Overfitting}
\label{ssec:nested}

As discussed in Section~\ref{ssec:evaluation_of_ml}, the evaluation of any learner should always be performed via resampling on independent test sets to ensure non-biased estimation of its generalization error. 
This is necessary because evaluating $\fh$ on the data set $\D$ that was used for its construction would lead to an optimistic bias.
In the general HPO problem as in Eq.~\eqref{eq:hpo_objective}, we already minimize this generalization error by resampling:
\begin{equation}
\lamh \in \argmin_{\lambdav \in \LamS} \clam = \argmin_{\lambdav \in \LamS}\GEhresa .
\end{equation}
If we simply report the estimated $\clamh$ value of the returned best HPC, this also creates an optimistically biased estimator of the generalization error, as we have violated the fundamental \enquote{untouched test set} principle by optimizing on the test set(s) instead. 

\begin{figure}[htb]
    \centering
    \includegraphics[width=1.0\textwidth]{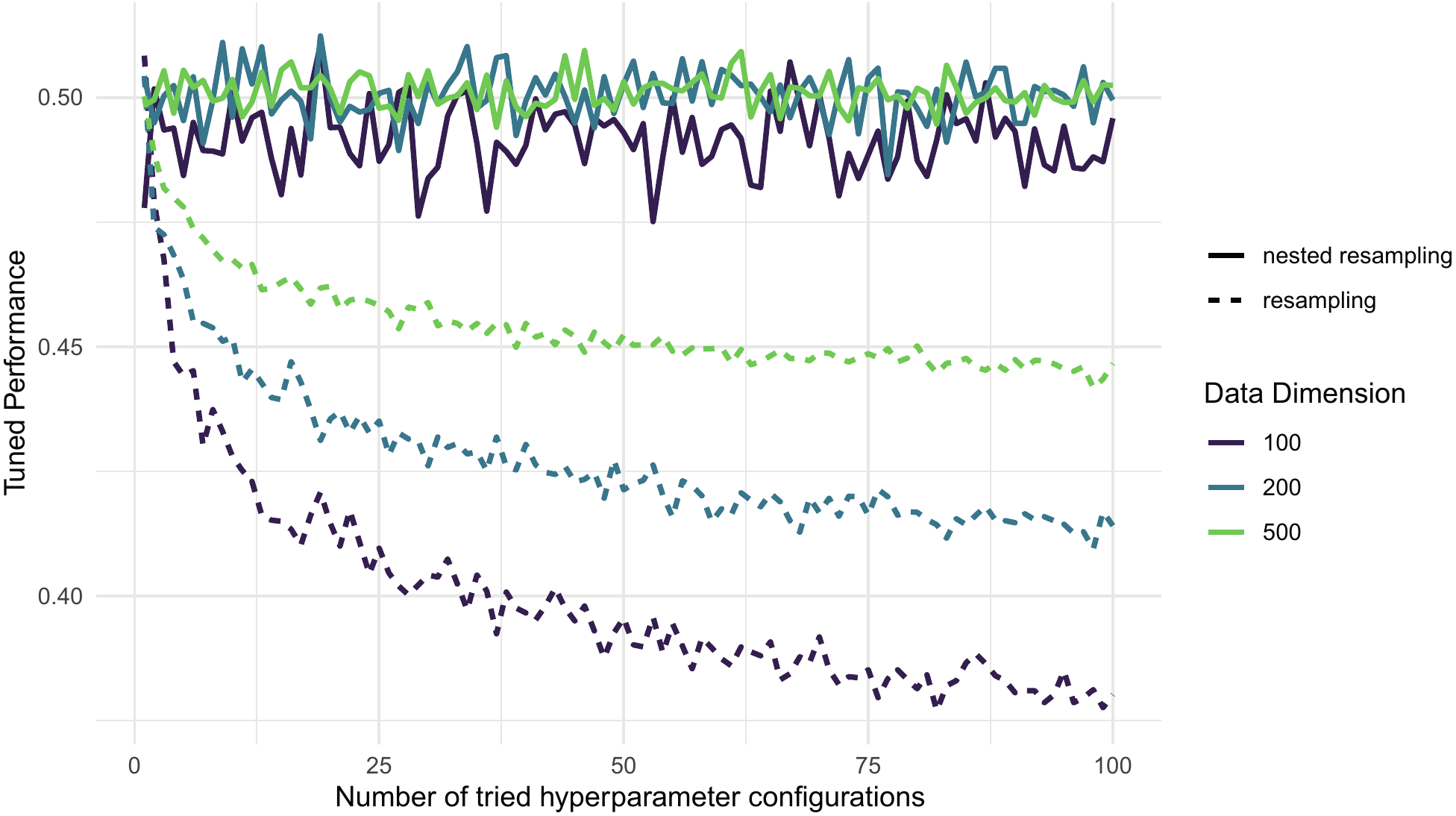}
    \caption{While nested resampling delivers correct results for the performance around $0.5$, taking the tuning result directly results in a biased, optimistic estimator, especially on smaller data sets.}
    \label{fig:overtuning}
\end{figure}
To better understand the necessity of an additional resampling step, we consider the following example in Figure~\ref{fig:overtuning}, introduced by \citet{bischl2012}.
Assume a balanced binary classification task and an inducer $\Ilam$ that ignores the data. Hence, $\lambdav$ has no effect, but rather \enquote{predicts} the class labels in a balanced but random manner.
Such a learner always has a true misclassification error of $\GEfull=0.5$ (using $\rho_{CE}$ as a metric), and any normal CV-based estimator will provide an approximately correct value as long as our data set is not too small.
We now \enquote{tune} this learner, for example, by RS -- which is meaningless, as $\lambdav$ has no effect.
The more tuning iterations are performed, the more likely it becomes that some model from our archive will produce partially correct labels simply by random chance, and the (only randomly) \enquote{best} of these is selected by our tuner at the end. 
The more we tune, the smaller our data set, or the more variance our GE estimator exhibits, the more expressed this optimistic bias will be.

\begin{figure}[htb]
    \centering
    \includegraphics[width = 1.00\textwidth]{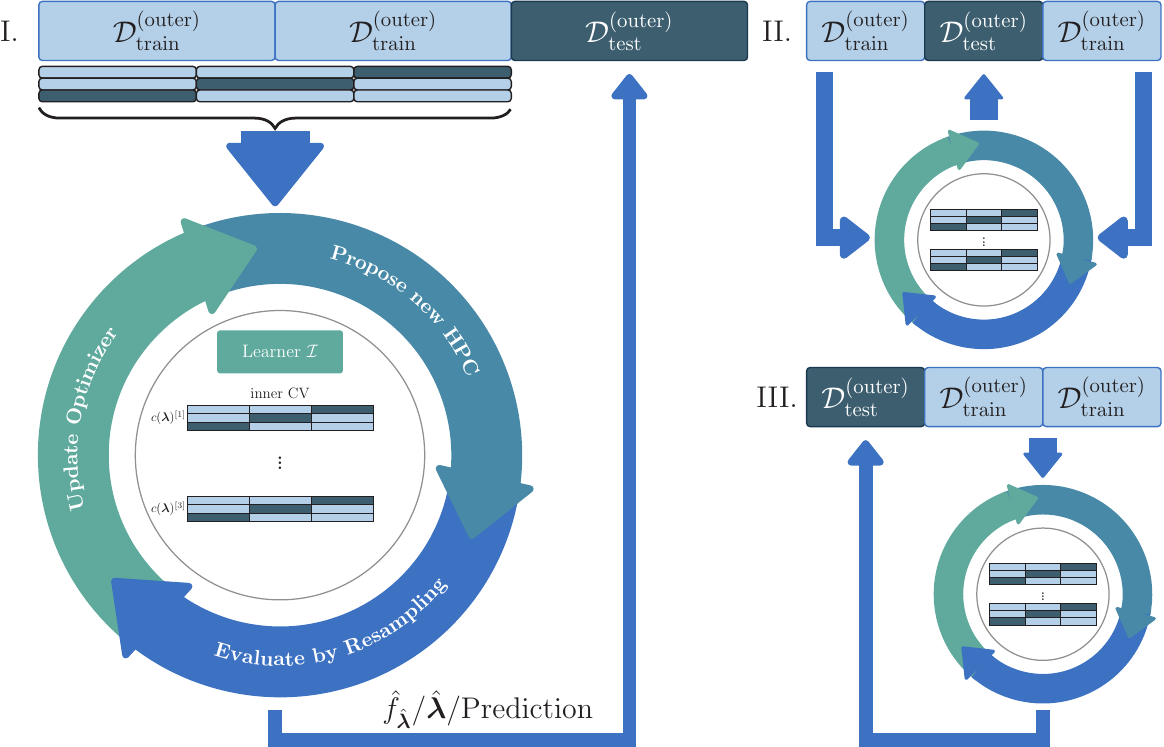}
    \caption{Nested CV with 3 inner and outer folds: each HPC is evaluated on an inner CV, while the resulting tuned model is evaluated on the outer test set.}
    \label{fig:nested_cv}
\end{figure}

To avoid this bias, we introduce an additional outer resampling loop around this inner HPO-resampling procedure -- or as discussed in Section~\ref{ssec:seclevel}, we simply regard this as cleanly cross-validating the self-tuned learner $\tunerfull$. 
This is called \emph{nested resampling}, which is illustrated in Figure~\ref{fig:nested_cv}.

The procedure works as follows: 
In the outer loop, an outer model-building or training set is selected, and an outer test set is cleanly set aside.
Each proposed HPC $\lamp$ during tuning is evaluated via inner resampling on the outer training set.
%a 3-fold CV, that is, averaging the performances of each inner test set.
The best performing HPC $\lamh$ returned by tuning is then used to fit a final model for the current outer loop on the outer training set, and this model is then cleanly evaluated on the test set.
This is repeated for all outer loops, and all outer test performances are aggregated at the end.

Some further comments on this general procedure:
\begin{enumerate*}[(i)]
\item Any resampling scheme is possible on the inside and outside, and these schemes can be flexibly combined based on statistical and computational considerations. Nested CV and nested holdout are most common.
\item Nested holdout is often called the \emph{train-validation-test} procedure, with the respective terminology for the generated three data sets resulting from the 3-way split.
\item Many users often wonder which \enquote{optimal} HPC $\lamh$ they are supposed to report or study if nested CV is performed, with multiple outer loops, and hence multiple outer HPCs $\lamh$. 
However, the learned HPs that result from optimizations within CV are considered temporary objects that merely exist in order to estimate $\GEhresa$.
The comparison to first-level risk minimization from Section~\ref{ssec:seclevel} is instructive here: The formal goal of nested CV is simply to produce the performance distribution on outer test sets; the $\lamh$ can be considered as the fitted HPs of the self-tuned learner $\tuner$.
If the parameters of a final model are of interest for further study, the tuner $\tuner$ should be fitted one final time on the complete data set. 
This would imply a final tuning run on the complete data set for second-level inference.
\end{enumerate*}

Nested resampling ensures unbiased outer evaluation of the HPO process, but, as CV for the first level, it is only a process that is used to estimate performance -- it does not directly help in constructing a better model.
%Note that in the example above, our inner error estimators were seriously biased, but our HPC selection was still correct, as in this hypothetical scenario, all configurations are equally bad. 
The biased estimation of performance values is not a problem for the optimization itself, as long as all evaluated HPCs are still ranked correctly.
But after a considerably large amount of evaluations, wrong HPCs might be selected due to stochasticity or overfitting to the splits of the inner resampling. 
This effect has been called either overtuning, meta-overfitting or oversearching
\citep{ng1997preventing,quinlan1995oversearching}.
At least parts of this problem seem directly related to the problem of multiple hypothesis testing. However, it has not been analysed very well yet, and unlike regularization for (first level) ERM, not many counter measures are currently known for HPO.

\subsection{Threshold Tuning}
\label{ssec:threshold_tuning}

Most classifiers do not directly output class labels, but rather probabilities or real-valued decision scores, although many metrics require predicted class labels.
A score is converted to a predicted label by comparing it to a threshold $t$ so that $\yh = [\fx \ge t]$, where we use the Iverson bracket $[]$ with $\yh=1$ if $\fx \ge t$ and $\yh=0$ in all other cases. 
For binary classification, the default thresholds are $t = 0.5$ for probabilities and $t = 0$ for scores.
%Depending on the used performance measure (c.f. Performance measures for classification based on class labels in Table~\ref{tab:measures}) a different threshold $t$ can be optimal.
However, depending on the metric and the classifier, different thresholds can be optimal.
\emph{Threshold tuning} \citep{sheng2006thresholding} is the practice of optimizing the classification threshold $t$ for a given model to improve performance. Strictly speaking, the threshold constitutes a further HP that must be chosen carefully. However, since the threshold can be varied freely after a model has been built, it does not need to be tuned jointly with the remaining HPs and can be optimized in a separate, subsequent, and cheaper step for each proposed HPC $\lamp$. 

After models have been fitted and predictions for all test sets have been obtained when $\clamp$ is computed via resampling, the vector of joint test set scores $\tilde{\F}$ can be compared against the joint vector of test set labels $\yv$ to optimize for an optimal threshold
%\begin{eqnarray}
$   \hat{t} = \argmin_{t} \rho(\tilde{\yv}, [\tilde{\F} \ge t])$, 
%\end{eqnarray}
where the Iverson bracket is evaluated component-wise and $t \in [0,1]$ for a binary probabilistic classifier and $t \in \R$ for a score classifier.
Since $t$ is scalar and evaluations are cheap, $\hat{t}$ can be found easily via a line search.
This two-step approach ensures that every HPC is coupled with its optimal threshold. 
$\clamp$ is then defined as the optimal performance value for $\lamp$ in combination with $\hat{t}$.

The procedure can be generalized to multi-class classification. 
Class probabilities or scores $\pikxh , k= 1,\ldots,g$ are divided by threshold weights $w_k$, $k = 1, \ldots, g$, and the $k$ that yields the maximal $\frac{\pikxh}{w_k}$ is chosen as the predicted class label. 
The weights $w_k, k = 1,\ldots,g$ are optimized in the same way as $t$ in the binary case.
Generally, threshold tuning can be performed jointly with any HPO algorithm.
In practice, threshold tuning is implemented as a post-processing step of an ML pipeline (Sections~\ref{ssec:linear_pipeline} \& \ref{ssec:complex_pipelines}). 

%\paragraph{Classifier calibration} is similar to threshold tuning but instead of $t$, directly adapts the classification probabilities $\pikxh , k= 1,\ldots,g$.
%A classifier is called \textit{calibrated} when the predicted probability of each class matches the expected empirical frequency of that class \citep{friedman_additive_2000}. 
%In practice, many ML algorithms produce badly calibrated models, for example (boosted) regression trees \citep{niculescu2005obtaining}.  
%Calibration methods like \textit{Logistic Correction} \citep{friedman_additive_2000}, \textit{Platt Scaling} \citep{platt_probabilistic_1999}, or \textit{Isotonic Regression} \citep{zadrozny2001,zadrozny2002} try to overcome this deficiency by changing the probabilities to match the empirical frequencies.

%% file: sec_05_pipelining_preprocessing_and_automl.tex
ML typically involves several data transformation steps before a learner can be trained.
If one or multiple preprocessing steps are executed successively, the data flows through a linear graph, also called pipeline.
Subsection~\ref{ssec:linear_pipeline} explains why a pipeline forms an ML algorithm itself, and why its performance should be evaluated accordingly through resampling.
Finally, Subsection~\ref{ssec:complex_pipelines} introduces the concept of flexible pipelines via hierarchical spaces.

\subsection{Linear Pipelines}\label{ssec:linear_pipeline}

\begin{figure}[htb]
    \begin{subfigure}[b]{0.67\textwidth}
    \includegraphics[width=1.0\textwidth]{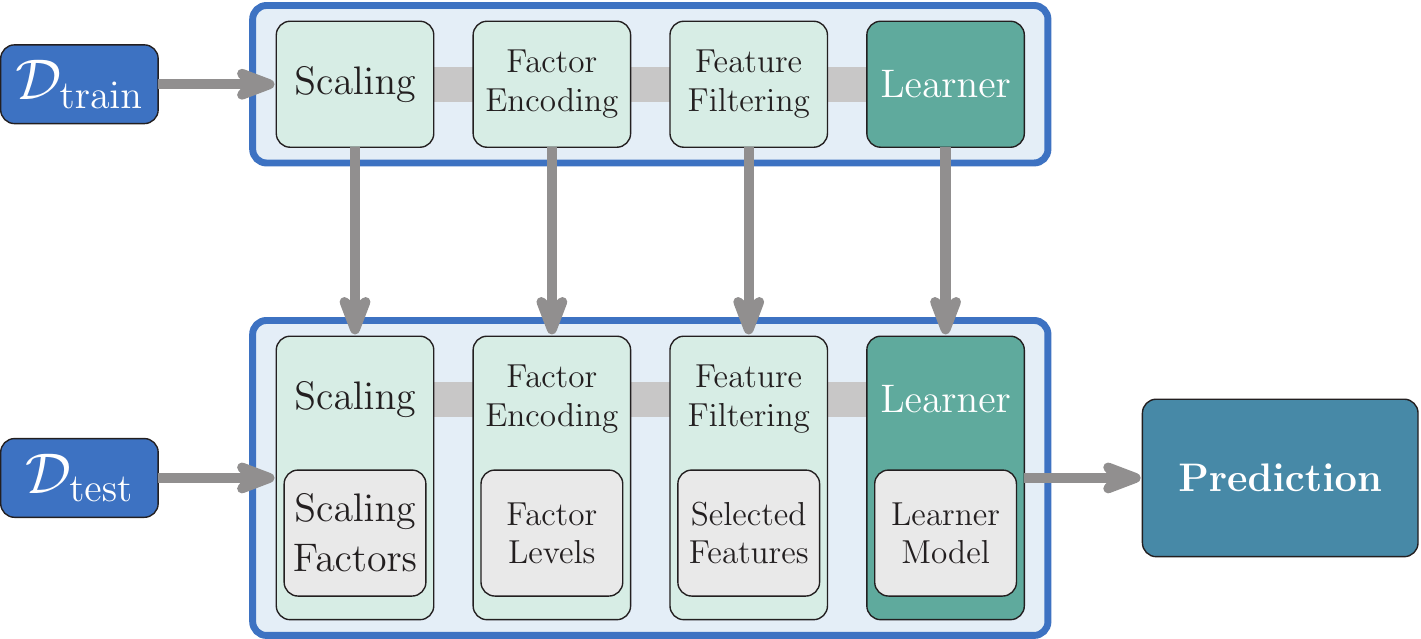}
    \end{subfigure}
    \begin{subfigure}[b]{0.32\textwidth}
    \includegraphics[width = 1.0\textwidth]{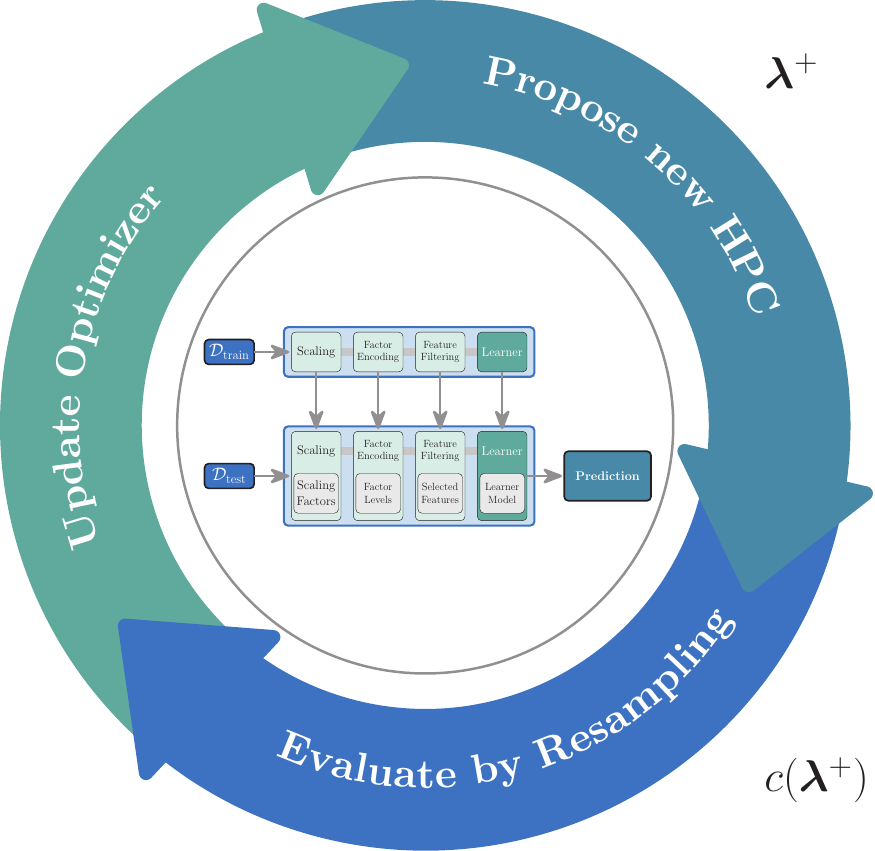}
    \end{subfigure}
    \caption{Example of a linear pipeline, including node parameters.}
    % replace imputation with filtering so that we have a hyperparameter
    \label{fig:linear_pipeline}
\end{figure}

In the following, we will extend the HPO problem for a specific learner towards configuring a full pipeline including preprocessing.
A \emph{linear ML pipeline} is a succession of preprocessing methods followed by a learner at the end, all arranged as nodes in a linear graph. Each node acts in a very similar manner as the learner, and has an associated training and prediction procedure. 
During training, each node learns its parameters (based on the training data) and sends a potentially transformed version of the training data to its successor.
Afterwards, a pipeline can be used to obtain predictions: The nodes operate on new data according to their model parameters.
Figure~\ref{fig:linear_pipeline} shows a simple example. 
Pipelines are important to properly embed the full model building procedure, including preprocessing,
into cross-validation, so every aspect of the model is only inferred from the training data.
This is necessary to avoid overfitting and biased performance evaluation \citep{bischl2012,hornung2015measure}, as it is for basic ML.
As each node represents a configurable piece of code, each node can have HPs, and the HPs of the pipeline are simply the joint set of all HPs of its contained nodes.
Therefore, we can model the whole pipeline as a single HPO problem with the combined search space $\LamS = \LamS_{\text{op}, 1} \times \cdots \times \LamS_{\text{op}, k} \times \LamS_\inducer$.

\subsection{Operator Selection and AutoML}\label{ssec:complex_pipelines}

More flexible pipelining, and especially the selection of appropriate nodes in a data-driven manner via HPO, can be achieved by representing our pipeline as a directed acyclic graph.
Usually, this implies a single source node that accepts the original data and a single sink node that returns the predictions.
Each node represents a preprocessing operation, a learner, a postprocessing operation, or a \emph{directive} operation that directs how the data is passed to the child node(s).

\begin{figure}[htb]
    \centering
    \includegraphics[width=1.0\textwidth]{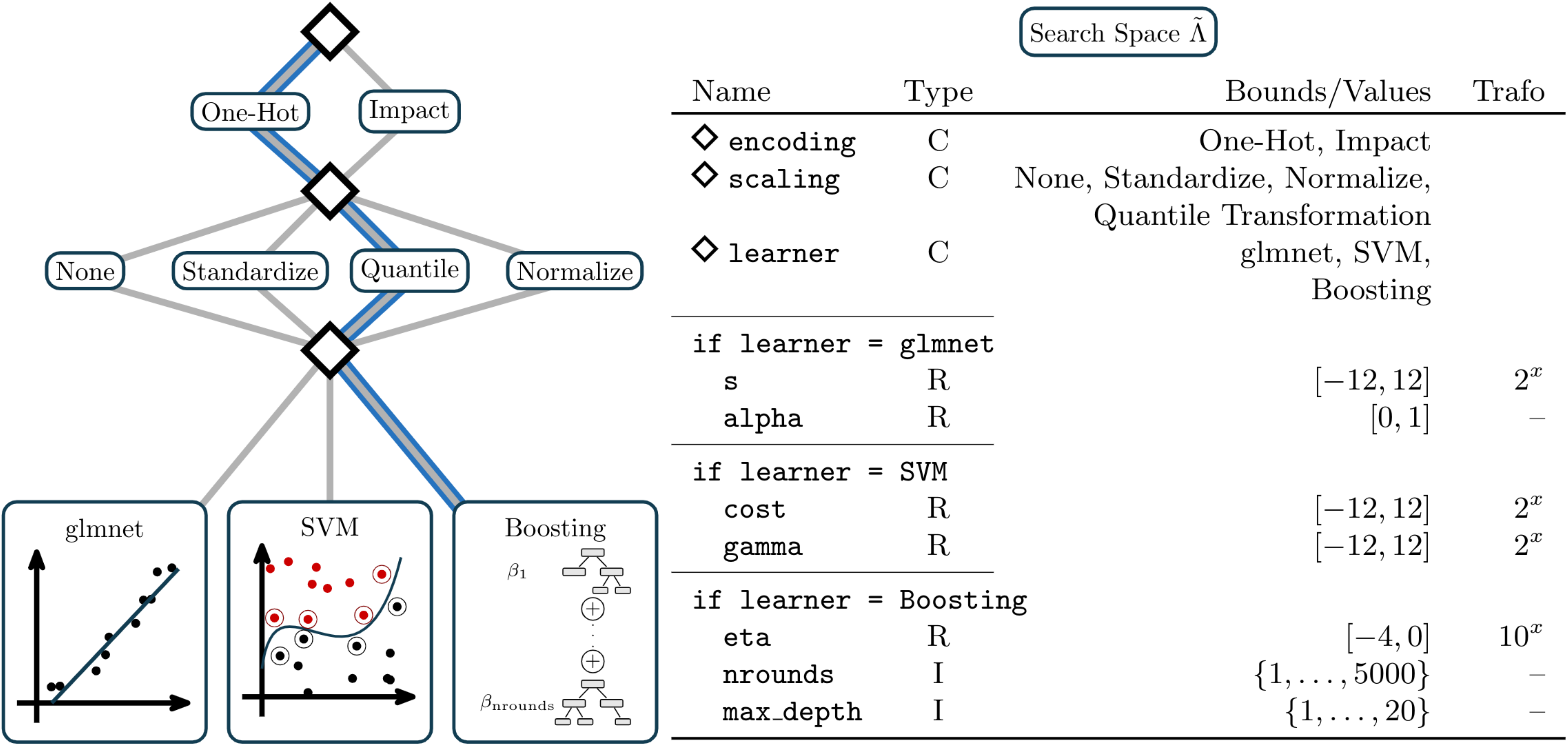}
    \caption{Example for a graph pipeline with operator selection via branches.}
    \label{fig:graph_pipeline}
\end{figure}

One instance of such a pipeline is illustrated in Figure~\ref{fig:graph_pipeline}.

Here, we consider the choice between multiple mutually exclusive preprocessing steps, as well as the choice between different ML algorithms. 
Such a choice is represented by a branching operator, which can be configured through a categorical parameter and can determine the flow of the data, resulting in multiple \enquote{modeling paths} in our graph.

The HP space induced by a flexible pipeline is potentially more complex.
Depending on the setting of the branching HP, different nodes and therefore different HPs are active, resulting in a very 
hierarchical search space.
If we build a graph that includes a sufficiently large selection of preprocessing steps combined with sufficiently many ML models, the result can be flexible enough to work well on a large number of data sets -- assuming it is correctly configured in a data-dependent manner.
Combining such a graph with an efficient tuner is the key principle of AutoML \citep{automlnips,mohr18,OlsonGECCO2016}.

%% file: sec_06_practical_aspects_of_hpo.tex
In this section, we discuss practical aspects of HPO, which are more qualitative in nature and less often discussed in an academic context. %Topics discussed here are less academic in nature than in previous sections.
Some of these recommendations are more \enquote{rules of thumb}, based solely on experience, while others are at least partially confirmed by empirical benchmarks -- keeping in mind that empirical benchmarks are only as good as the selection of data sets.
Even if a proper empirical study cannot be cited for every piece of advice in this section, we still propose that such a compilation is highly valuable for HPO beginners.

\subsection{Choosing Resampling and Performance Metrics}\label{ssec:choosing_resampling}
A resampling procedure is usually chosen based on two fundamental properties of the available data: (i)~the number of observations, i.e., to what degree do we face a small sample size situation, and (ii)~whether the i.i.d.\ assumption for our data sampling process is violated.

For smaller data sets, e.g.,\ $n<500$, repeated CV with a high number of repetitions should be used to reduce the variance while keeping the pessimistic bias small \citep{bischl2012}. 
The larger the data set, the fewer splits are necessary. Consequently, for data sets of \enquote{medium} size with $500 \leq n \leq 50000$, usually 5- or 10-fold CV is recommended. Beyond that, 
simple holdout might be sufficient.
Note that even for large $n$, sample size problems can occur, for example, when data is imbalanced. In that case, repeated resampling might still be required to obtain properly accurate performance estimates. 
\textit{Stratified} sampling, which ensures that the relative class frequencies for each train/test split are consistent with the original data set, helps in such a case.

One fundamental assumption about our data is that observations are i.i.d., i.e., $\obs \iid \Pxy.$
This assumption is often violated in practice.
A typical example is \textit{repeated measurements}, where observations occur in \enquote{blocks} of multiple, correlated data, e.g., from different hospitals, cities or persons.
In such a scenario, we are usually interested in the ability of the model to generalize to new blocks. We must then perform CV with respect to the blocks, e.g., \enquote{leave one block out}. 
A related problem occurs if data are collected sequentially over a period of time.
In such a setting, we are usually interested in how the model will generalize in the future, 
and the rolling or expanding window forecast must be used for evaluation \citep{BERGMEIR201870} instead of regular CV.
However, discussing these special cases is out of scope for this work.

In HPO, resampling strategies must be specified for the inner as well as the outer level of nested resampling. 
The outer level is simply regular ML evaluation, and all comments from above hold. We advise readers to study further material such as \citet{japkowicz_shah_2011}.
The inner level concerns the evaluation of $\clam$ through resampling during tuning. While the same general comments from above apply, in order to reduce runtime, repetitions can also be scaled down. 
We are not particularly interested in very accurate numerical performance estimates at the inner level, and we must only ensure that HPCs are properly ranked during tuning to achieve correct selection, as discussed in Section~\ref{ssec:nested}.
%It is challenging to translate this into concrete, practical advice. 
Hence, it might be appropriate to use a 10-fold CV on the outside to ensure proper generalization error estimation of the tuned learner, but to use only 2 folds or simple holdout on the inside. 
In general, controlling the number of resampling repetitions on the inside should be considered an aspect of the tuner and should probably be automated away from the user (without taking away flexible control in cases of i.i.d.\ violations, or other deviations from standard scenarios). However, not many current tuners provide this, although racing is one of the attractive exceptions.

The choice of the performance measure should be guided by the costs that suboptimal predictions by the model and subsequent actions in the real-word context of applying the model would incur.
Often, popular but simple measures like accuracy do not meet this requirement. 
Misclassification of different classes can imply different costs. For example, failing to detect an illness may have a higher associated cost than mistakenly admitting a person to the hospital.
There exists a plethora of performance measures that attempt to emphasize different aspects of misclassification with respect to prediction probabilities and class imbalances, c.f. \citet{japkowicz_shah_2011}%
\ifarxiv
 and many listed in Table~\ref{tab:measures} in Appendix~\ref{app:evaluation}.
\else
.
\fi
For other applications, it might be necessary to design a performance measure from scratch or based on underlying business key performance indicators (KPIs).

While a further discussion of metrics is again out of scope for this article, two pieces of advice are pertinent. 
First, as HPO is a black-box, no real constraints exist regarding the mathematical properties of $\rho$ (or the associated outer loss). 
For first-level risk minimization, on the other hand, we usually require differentiability and convexity of $L$. If this is not fulfilled, we must approximate the KPI with a more convenient version. 
%All ML frameworks also allow the optimization of custom losses, either regarding their mathematical definition or in their implementation.
Second, for many applications, it is quite unclear whether there is a single metric that captures all aspects of model quality in a balanced manner.
In such cases, it can be preferable to optimize multiple measures simultaneously, resulting in a multi-criteria optimization problem \citep{horn2016multi}.

\subsection{Choosing a Pipeline and Corresponding Search Space}\label{ssec:choosing_pipeline}

For HPO, it is necessary to define the \emph{search space} $\LamS$ over which the optimization is to be performed. 
This choice can have a large impact on the performance of the tuned model.
A simple search space is a (lower dimensional) Cartesian product of individual HP sets that are either numeric (continuous or integer-valued) or categorical. 
Encoding categorical values as integers is a common mistake that degrades the performance of optimizers that rely on information about distances between HPCs, such as BO. 
The search intervals of numeric HPs typically must be bounded within a region of plausibly well-performing values for the given method and data set.

Many numeric HPs are often either bounded in a closed interval (e.g.,~$[0, 1]$) or bounded from below (e.g.,~$[0,\infty)$).
The former can usually be tuned without modifications.
HPs bounded by a left-closed interval should often be tuned on a \emph{logarithmic} scale with a generous upper bound, as the influence of larger values often diminishes. 
For example, the decision whether $k$-NN uses $k=2$ vs.\ 3 neighbors will have a larger impact than whether it uses $k=102$ vs.\ $k=103$ neighbors. 
The logarithmic scale can either be defined in the tuning software or must be set up manually by adjusting the algorithm to use \emph{transformations}: 
If the desired range of the HP is $[a, b]$, the tuner optimizes on $[\log a, \log b]$, and any proposed value is transformed through an exponential function before it is passed to the ML algorithm.
The logarithm and exponentiation must refer to the same base here, but which base is chosen does not influence the tuning process.

The size of the search space will also considerably influence the quality of the resulting model and the necessary budget of HPO.
If chosen too small, the search space may not contain a particularly well-performing HPC.
Choosing too wide HP intervals or including inadequate HPs in the search space can have an adverse effect on tuning outcomes in the given budget.
If $\LamS$ is simply too large, it is more difficult for the optimizer to find the optimum or promising regions within the given budget.
Furthermore, restricting the bounds of an HP may be beneficial to avoid values that are \textit{a priori} known to cause problems due to unstable behavior or large resource consumption.
If multiple HPs lead to poor performance throughout a large part of their range -- for example, by resulting in a degenerate ML model or a software crash -- the fraction of the search space that leads to fair performance then shrinks exponentially in the number of HPs with this problem. 
This effect can be viewed as a further manifestation of the so-called \emph{curse of dimensionality}. 

Due to this curse of dimensionality and the  considerable runtime costs of HPO, we would like to tune as few HPs as possible.
If no prior knowledge from earlier experiments or expert knowledge exists, it is common practice to leave other HPs at their software default values with the assumption that the developers of the algorithm chose values that work well under a wide range of conditions, which is not necessarily given and its often not documented how these defaults were specified.
Recent approaches have studied how to empirically find optimal default values, tuning ranges and HPC prior distributions based on extensive meta-data \citep{wistuba_dsaa15a,wistuba_pkdd15a,automlnips,pfisterer2018learning,van2018hyperparameter,probst2019tunability,NEURIPS2019_6ea3f187,gijsberg2021meta}.
%However, they emphasize that for some HPs, tuning generally yields performance gains.

%%%%%

%\citet{probst2019tunability} show that it is possible to empirically find \enquote{optimal defaults} that generally work well on a diverse portfolio of data sets as well as to also learn tuning ranges based on meta-data.
%However, they emphasize that for some HPs, tuning generally yields performance gains.

It is possible to optimize several learning algorithms in combination, as described in Section~\ref{ssec:complex_pipelines}, but this introduces HP dependencies.
The question then arises of which of the large number of ML algorithms (or preprocessing operations) should be considered. 
However, \citet{delgado2014we} showed that in many cases, only a small but diverse set of learners is necessary to choose one ML algorithm that performs sufficiently well.

\subsection{Choosing an HPO Algorithm}\label{ssec:choosing_hpo}

The number of HPs considered for optimization has a large influence on the difficulty of an HPO problem.
Particularly large search spaces typically arise from optimizing over large pipelines or multiple learners (Section~\ref{sec:preproc}).
With very few HPs (up to about 2-3) and well-functioning discretization, GS may be useful due to its interpretable, deterministic, and reproducible nature; however, it is not recommended beyond this \citep{bergstra12}.
BO with GPs work well for up to around 10 HPs. However, more HPs typically require more function evaluations -- which in turn is problematic runtime-wise, since GPs scale cubically with the number of dimensions. On the other hand, BO with RFs have been used successfully on search spaces with hundreds of HPs \citep{autoweka} and can usually handle mixed hierarchical search spaces better. 
Pure sampling-based methods, such as RS and Hyperband, work well even for very large HP spaces as long as the \enquote{effective} dimension (i.e., the number of HPs that have a large impact on performance) is low, which is often observed in ML models \citep{bergstra12}. 
Evolutionary algorithms (and those using similar metaheuristics, such as racing) can also work with truly large search spaces, and even with search spaces of arbitrarily complex structure if one is willing do use non-custom mutation and crossover operators. Evolutionary algorithms may also require fewer objective evaluations than RS. Therefore, they occupy a middle ground between (highly complex, sample-efficient) BO and (very simple but possibly wasteful) RS. 

Another property of algorithms that is especially relevant to practitioners with access to large computational resources is \emph{parallelizability}, which is discussed in Subsection~\ref{ssec:parallelizability}.
Furthermore, HPO algorithms differ in their simplicity, both in terms of algorithmic principle and in terms of usability of available implementations, which can often have implications for their usefulness in practice. 
While more complex optimization methods, such as those based on BO, are often more sample-efficient or have other desirable properties compared to simple methods, they also have more components that can fail. % (including the surrogate model or the acquisition function to trade off exploration and exploitation).
When performance evaluations are cheap and the search space is small, it may therefore be beneficial to fall back on simple and robust approaches such as RS, Hyperband, or any tuner with minimal inference overhead.
The availability of any implementation at all (and the quality of that implementation) is also important; there may be optimization algorithms based on beautiful theoretical principles that have a poorly maintained implementation or are based on out-of-date software. 
The additional cost of having to port an algorithm to the software platform being used, or even implement it from scratch, could be spent on running more evaluations with an existing algorithm.

One might wish to select an HPO algorithm that performed best on previous benchmarks.
However, no single benchmark exists which includes all relevant scenarios and whose results generalize to all possible applications. 
Specific benchmark results can therefore only indicate how well an algorithm works for a selected set of data sets, a predefined budget, specific parallelization, specific learners, and search spaces.
Even worse, extensive comparison studies are missing in the current literature, although
efforts have been made to establish unified benchmarks. \cite{eggensperger2013towards} showed that (i) BO with GPs were a strong approach for small continuous spaces with few evaluations, (ii) BO with Random Forests performed well for mixed spaces with a few hundred evaluations, and (iii) for large spaces and many evaluations, ES were the best optimizers.
%\footnote{\url{ https://github.com/automl/HPOBench}}

\subsection{Choosing an Implementation}\label{ssec:choosing_implementation}

In addition to the choice of algorithm, the choice of implementation is also important in practice.
We note that a thorough comparison of all available software packages and frameworks is not within the scope of this article.
Nevertheless, we provide a brief list of popular implementations and frameworks as reasonable starting points for newcomers.

Simple and established tuning algorithms for HPO are usually shipped with general purpose ML frameworks such as \texttt{Scikit-learn} \citep{scikit_learn}, \texttt{Tensorflow}\footnote{\url{https://www.tensorflow.org/}}, \texttt{PyTorch} \citep{pytorch_neurips19a}, \texttt{MXNet} \citep{chen_arxiv15a}, \texttt{mlr3} \citep{lang_mlr3_2019,mlr3pipelines}, \texttt{tidymodels}\footnote{\url{https://www.tidymodels.org/}} or \texttt{h2o.ai}\footnote{\url{https://github.com/h2oai/h2o-3}}.
Although modern state-of-the-art algorithms often build on, extend, or connect to such an ML framework, they are usually developed in independent software projects.

For Python, there exist a plethora of HPO toolkits, e.g., \texttt{Spearmint} \citep{snoek_nips12a}, \texttt{SMAC}\footnote{\url{https://github.com/automl/SMAC3}} \citep{hutter_lion11a}, \texttt{BoTorch} \citep{balandat2020botorch}, \texttt{Dragonfly} \citep{kandasamy2020dragonfly}, %Ray Tune \citep{liaw2018tune},
or \texttt{Or\'ion}\footnote{\url{https://github.com/Epistimio/orion}}.
Multiple HPO methods are supported by toolkits like \texttt{Hyperopt} \citep{hyperopt}, Optuna \citep{akiba_kdd2019} or Weights \& Biases\footnote{\url{https://www.wandb.com/}}.
A popular framework that combines modern HPO approaches with the \texttt{Scikit-learn} toolbox for ML in Python is \texttt{auto-sklearn} \citep{feurer_nips2015a}.

The availability of HPO implementations in R is comparably smaller.
%Many of the very recently developed algorithms are not (yet) connected.
However, more established HPO tuners are either already shipped with the ML framework, or can be found, e.g., in the packages \texttt{mlrMBO}\footnote{\url{https://github.com/mlr-org/mlrMBO}}, \texttt{irace} \citep{lopez_2016}, \texttt{DiceOptim} \citep{roustant2012dicekriging}, or \texttt{rBayesianOptimization}\footnote{\url{https://github.com/yanyachen/rBayesianOptimization}}.

\ifarxiv
See Appendix~\ref{app:software} for more information

%%%%

\subsection{When to Terminate HPO}
\label{ssec:choosing_term}

Choosing an appropriate budget for HPO or dynamically terminating HPO based on collected archive data is a difficult practical challenge that is not readily solved. 
The user typically has these options:
\begin{enumerate*}[(i)]
\item A certain amount of runtime is specified before HPO is started, solely based on practical considerations and intuition. 
This is what currently happens nearly exclusively in practice.
A simple rule-of-thumb might be scaling the budget of HPC evaluations with search space dimensionality in a linear fashion like $50 \times l$ or $100 \times l$, which would also define the number of full budget units in a multi-fidelity setup. 
The downside is that while it is usually simpler to define when a result is needed, it is nearly impossible to determine whether this computational budget is enough to solve the HPO problem reasonably well.
\item The user can set a lower bound regarding the required generalization error, i.e., what model quality is deemed good enough in order to put the resulting model to practical use. Notably, if this lower bound is set too optimistically, the HPO process might never terminate or simply take too long.
\item If no (significant) progress is observed for a certain amount of time during tuning, HPO is considered to have converged or stagnated and stopped. 
This criterion bears the risk of terminating too early, especially for large and complex search spaces.
Furthermore, this approach is conceptually problematic, as it mainly makes sense for iterative, local 
optimizers. No HPO algorithm from this article, except for maybe ES, belongs to this category, as they all contain a strong exploratory, global search component. 
This often implies that search performance can flatline for a longer time and then abruptly change.
\item For BO, a small maximum value of the $\operatorname{EI}$ acquisition function (see Eq.~\ref{eq:EI}) indicates that further progress is estimated to be unlikely \citep{jones1998efficient}. This criterion could even be combined with other, no-BO type tuners. It implies that the surrogate model is trustworthy enough for such an estimate, including its local uncertainty estimate. This is somewhat unlikely for large, complex search spaces.
\end{enumerate*}
In practice, it is probably prudent to set up a combination of all mentioned criteria (if possible) as an \enquote{and} or \enquote{or} combination, with somewhat liberal thresholds to 
continue the optimization for (ii), (iii), (iv), and (v), and an absolute bound on the maximal runtime for (i).

With many HPO tuners, it is possible to continue the optimization even after it has been terminated. 
One can even use (part of) the archive as an initialization of the next HPO run, e.g., as an initial design for BO or as the initial population for an ES.

Some methods, e.g.,\ ES and under some circumstances BO, may get stuck in a subspace of the search space and fail to explore other regions. 
While some means of mitigation exist, such as interleaving proposed points with randomly generated points, there is always the possibility of re-starting the optimization from scratch multiple times, and selecting the best performance from the combined runs as the final result. 
The decision regarding the termination criterion is itself always a matter of cost-benefit calculation of potential increased performance against cost of additional evaluations. One should also consider the possibility that, for more local search strategies like ES, terminating early and restarting the optimization can be more cost-effective than letting a single optimization run continue for much longer. 
For more global samplers like BO and HB, it is much less clear whether and how such an efficient restart could be executed. 

\subsection{Warm-Starts}\label{ssec:run_time_considerations}

HPO may require substantial computational resources, as a large optimization search space may require many performance evaluations, and individual performance evaluations can also be very computationally expensive. 
One way to reduce the computational time are warm-starts, where information from previous experiments are used as a starting solution.

\paragraph{Warm-Starting Evaluations}
Certain model classes may offer specific methods that reduce the computational resources needed for training a single model by transferring model parameters from other, similar configurations that were already trained. 
NNs with similar architectures can be initialized from trained networks to reduce training time -- a concept known as weight sharing. 
Some neural architecture search algorithms are specifically designed to make use of this fact \citep{elsken_neural_2019}. 

\paragraph{Warm-Starting Optimization}

Many HPO algorithms do not take any input regarding the relative merit of different HPCs beyond box-constraints on the search space $\LamS$. However, it is often the case that some HPCs work relatively well on a large variety of data sets \citep{probst2019tunability}. At other times, HPO may have been performed on other problems that are similar to the problem currently being considered. In both cases, it is possible to warm-start the HPO process itself -- for example, by choosing the initial design as HPCs that have performed well in the past or by choosing HPCs that are known to perform well in general \citep{Lindauer_Hutter_2018}. 
This can also be regarded as a transfer learning mechanism for HPO.
Large systematic collections of HPC performance on different data sets, such as those collected in \citet{binder20} or on OpenML \citep{van2013openml}, can be used to build these initial designs \citep{pfisterer2018learning, automlnips}.

\subsection{Control of Execution}\label{ssec:parallelization}

HP tuning is usually computationally expensive, and running optimization in parallel is one way to reduce the overall time needed for a satisfactory result.
During parallelization, \textit{computational jobs} (which can be arbitrary function calls) are mapped to a pool of \textit{workers} (usually distributed among individual physical CPU cores, or remote computational nodes).
In practice, a \textit{parallelization backend} orchestrates starting up, shutting down, and communication with the workers.
The achieved speedup can be proportional to the number of workers in ideal cases, but less than linear scaling is typically observed, depending on the algorithm used (\textit{Amdahl's law}, \citealp{Rodgers1985}). 
Besides parallelization, ensuring a fail-safe program flow is often also mandatory, and requires special care.

\subsubsection{Job Hierarchy for HPO with Nested Resampling}
\label{ssec:parallelisation_hpo}

HPO can be parallelized at different granularity or \textit{parallelization levels}.
These levels result from the nested loop outlined in Algorithm~\ref{alg:nested_loops} and described in detail in the following (from coarse to fine granularity):
\begin{algorithm}[htb]
\ForEach {outer resampling iteration}{ \label{line:loop_outer}
    Initialize archive $\mathcal{A} = \{\}$\;
    \ForEach{tuning iteration}{ \label{line:loop_tuning}
        $\{\lambdav^{+,1}. \ldots, \lambdav^{+,n_{\text{batch}}}\}$ = propose\_points($\mathcal{A}$)\;
        \ForEach{$i \in \{1, \ldots, n_{\text{batch}}\}$}{\label{line:loop_configs}
            $\bm{c} = \{\}$\;
            \ForEach{inner resampling iteration}{\label{line:loop_inner}
                $\bm{c} = \bm{c} \cup \operatorname{eval}(\lambdav^{+,i}$, $\mathcal{D}_\text{inner\_train}, \mathcal{D}_\text{inner\_test})$\;\label{line:eval_model}
            }
            $\mathcal{A}$ = $\mathcal{A} \cup (\lambdav^{+,i}, \agr (\bm{c}))$\;
        }
    }
    $\lams$ = get\_best($\mathcal{A}$)\;
    eval($\lams$, $\mathcal{D}_\text{train}, \mathcal{D}_\text{test}$)\;\label{line:eval_model_two}
}
\caption{Nested resampling as nested loops.}
\label{alg:nested_loops}
\end{algorithm}
\begin{enumerate}[(a)]
    \item One iteration of the \textit{outer resampling loop} (Line~\ref{line:loop_outer}), i.e., tuning an ML algorithm on the respective training set of the outer resampling split. The result is the outer test set performance of the best HPC found and trained on the outer training set. \label{enum:job_outer_resampling}
    \item The execution of one \textit{tuning iteration} (Line~\ref{line:loop_tuning}), i.e., the proposal and evaluation of a batch of new HPCs. The result is an updated optimization archive. \label{enum:job_tuning_iteration}
    \item One \textit{evaluation} of a single proposed HPC (Line~\ref{line:loop_configs}), i.e., one inner resampling with configuration $\lambdav^{+,i}$.
    The result is an aggregated performance score. \label{enum:job_config_evaluation}
    \item One iteration of the \textit{inner resampling} loop (Line~\ref{line:loop_inner}). \label{enum:job_resampling}
    The result is the performance score for a single inner train / test split.
    \item The \textit{model fit} for the evaluation of the model itself is sometimes also parallelizable (Line~\ref{line:eval_model} and~\ref{line:eval_model_two}). For example, the individual trees of a random forest can be fitted independently and as such are an obvious target for parallelization.
\end{enumerate}
Note that the $k_{\text{inner}}$ resampling iterations created in \ref{enum:job_resampling} are independent between the $n_{\text{batch}}$ HPC evaluations created in \ref{enum:job_config_evaluation}. Therefore, they form $n_{\text{batch}} \cdot k_{\text{inner}}$ independent jobs that can be executed in parallel.

An HPO problem can now be distributed to the workers in several ways.
For example, if one wants to perform a 10-fold (outer) CV of \emph{BO} that proposes one point per tuning iteration, with a budget of 50 HP evaluations done via a 3-fold CV (inner resampling), one can decide to
\begin{enumerate*}[(i)]
 \item spawn 10 parallel jobs (level~\ref{enum:job_outer_resampling}) with a long runtime once, or to
 \item spawn 3 parallel jobs (level~\ref{enum:job_resampling}) with a short runtime 50 times.
\end{enumerate*}

Consider another example with an \emph{ES} running for 50 generations with an offspring size of 20.
This translates to 50 tuning iterations with 20 HPC proposals per iteration. 
If this is evaluated using nested CV, with 10 outer and 3 inner resampling loops, then the parallelization options are:
\begin{enumerate*}[(i)]
    \item spawn 10 parallel jobs with a long runtime once (level~\ref{enum:job_outer_resampling}),
    \item spawn 20 parallel jobs (level~\ref{enum:job_config_evaluation}) $10 \cdot 50$ times with a medium runtime,
    \item spawn 3 parallel jobs (level~\ref{enum:job_resampling}) with a short runtime $10 \cdot 50 \cdot 20$ times, or 
    \item spawn $20 \cdot 3 = 60$ parallel jobs (level~\ref{enum:job_resampling} and~\ref{enum:job_config_evaluation} combined) with a short runtime $10 \cdot 50$ times.
\end{enumerate*}

\subsubsection{Parallelizability}\label{ssec:parallelizability}

These examples demonstrate that the choice of the parallelization level also depends on the choice of the tuner.
Parallelization for RS and GS is so straightforward that it is also called \enquote{embarrassingly parallel}, since all HPCs to be evaluated can be determined in advance as they have only one (trivial) tuning iteration (level~\ref{enum:job_tuning_iteration}). 
Algorithms based on iterations over batch proposals (ES, BO, IR, Hyperband) are limited in how many parallel workers they can use and profit from: they can be parallelized up to their (current) batch size, which decreases for HB and IR during a bracket/race.
The situation looks very different for BO; by construction, it is a sequential algorithm with a batch size of 1. Multi-point proposals (see~\ref{sssec:model_based_optimization}) must be used to parallelize multiple configurations in each iteration (level~\ref{enum:job_config_evaluation}).
However, parallelization does not scale very well for this level. If possible, a different parallelization level should be chosen to achieve a similar speedup but avoid the problems of multi-point proposal.

Therefore, if a truly large number of parallel resources is available, then the best optimization method may be RS due to its relatively boundless parallelizability and low synchronization overhead. With fewer parallel resources, more efficient algorithms have a larger relative advantage.

\subsubsection{Parallelization Tweaks}

The more jobs generated, the more workers can be utilized in general, and the better the tuning tends to scale with available computational resources. 
However, there are some caveats and technical difficulties to consider. 
First, the process of starting a job, communicating the inputs to the worker, and communicating the results back often comes with considerable overhead, depending on the parallelization backend used. 
To maximize utilization and minimize overhead, some backends therefore allow chunking multiple jobs together into groups that are calculated sequentially on the workers.
Second, whenever a batch of parallel jobs is started, the main process must wait for all of them to finish, which leads to \emph{synchronization overhead} if nodes that finish early are left idling because they wait for longer running jobs to finish. 
This can be a particular problem for HPO, where it is likely that the jobs have heterogeneous runtimes. 
For example, fitting a boosting model with 10 iterations will be significantly faster than fitting a boosting model with 10,000 iterations.
Unwittingly chunking many jobs together can exacerbate synchronization overhead and lead to a situation where most workers idle and wait for one worker to finish a relatively large chunk of jobs.
While there are approaches to mitigate such problems, as briefly discussed in Section~\ref{sssec:model_based_optimization} for BO, it is often sufficient -- as a rule of thumb -- to aim for as many jobs as possible, as long as each individual job has an expected average runtime of $\geq$ 5 minutes. 
Additionally, randomizing the order of jobs can increase the utilization of search strategies such as GS.

\paragraph{Budget and Overhead}
For an unbiased comparison, it is mandatory that all tuners are granted the same budget.
However, if the budget is defined by a fixed number evaluations, it is unclear how to count multi-fidelity evaluations and how to compare them against \enquote{regular} evaluations.
We recommend comparing evaluations by wall-clock time, and to explicitly report overhead for model inference and point proposal. 

\paragraph{Anytime Behavior} In practice, it is unclear how long tuners will and should be run. Thus, when benchmarking tuners, the performance of the optimization process should be considered at many stages.
To achieve this objective, one should not (only) compare the final performance at the maximum budget, but rather compare the whole optimization traces from initialization to termination. 
Furthermore, tuning runs must be terminated if the wall-clock time is exceeded, which may leave the tuner in an undefined state and incapable of extracting a final result.

\paragraph{Parallelization}
The effect of parallelization must be factored in\footnote{At least if the budget is defined by available wall-clock time}. 
While some tuners scale linearly with the number of workers without a degradation in performance (e.g., GS), other tuners (e.g., BO) show diminishing returns from excessive parallelism.
A fair comparison w.r.t.\ a time budget is also hampered by technical aspects like cache interference or difficult to control parallelization of the ML algorithms to tune.

\subsection{Simple Sequential Approach to Model Selection}
In practical model selection, it is often the case that practitioners care not only about predictive performance, but also about other aspects of the model such as its simplicity and interpretability.
If a practitioner considers only a low number of different model classes, then a practical alternative to AutoML on a full pipeline space of different ML models is to construct a manual sequence of preferred models of increasing complexity, to separately tune them, and to compare their performance.
An informed choice can then be made about how much additional complexity is acceptable for a certain amount of improved predictive performance. 
\citet{hastie_elements_2009} suggest the \enquote{one-standard-error rule} and use the simplest model within one standard error of the performance of the best model.
The specific sequence of learners will differ for each use case but might look something like this: (1) linear model with only a single feature or tree stump, (2) L1/L2 regularized linear model or a CART tree, (3) component-wise boosting of a GAM, (4) random forest, (5) gradient tree boosting, and (6) full AutoML search space with pipelines. 
If even more focus is put on predictive performance (multi-level), stacking a diverse set of models can give a final (often rather small) boost \citep{wolpert1992stacked}.

\subsection{Benchmarking HPO Algorithms} \label{ssec:comparison_of_hpo}

While few well-designed large-scale benchmarks of HPO exist \citep{klein2019meta,EggenspergerLHH18,eggensperger2013towards}, there is still an ongoing endeavour of the community for better and more comprehensive studies.
Usually, tuners are compared on fixed sets of data sets and search spaces or cheaper approximations via empirical performance models / surrogates \citep{EggenspergerLHH18}. It is still unclear how an ideal and representative benchmark set for HPO that also enables efficient benchmarking might look like.

%% file: sec_07_related_fields.tex
Besides HPO, there are many related scientific fields that face very similar problems on an abstract level and nowadays resort to techniques that are very similar to those described in this work. Although detailed coverage of these related areas is out of scope for this paper, we nevertheless briefly summarize and compare them to HPO.

\paragraph{Neural Architecture Search}
A specific type of HPO is neural architecture search (NAS) \citep{elsken_neural_2019}, where the task is to find a well-performing architecture of a deep NN for a given data set. Although NAS can also be formulated as an HPO problem \citep[e.g.,][]{ZimLin2021a}, it is usually approached as a bi-level optimization problem that can be simultaneously solved while training the NN \citep{liu2018darts}. Although it is common practice to optimize architecture and HPCs in sequence, recent evidence suggests that they should be jointly optimized \citep{zela_automl18,JMLRv2120_056}.

\paragraph{Algorithm Selection and Traditional Meta-Learning}
In HPO, we actively \emph{search} for a well-performing HPC through iterative optimization. 
In the related problem of algorithm selection \citep{Rice76}, we usually train a meta-learning model offline to select an optimal element from a finite set of algorithms or HPCs \citep{BischlKKLMFHHLT16}. This model can be trained on empirical meta-data of the candidates' performances on a large set of problem instances or data sets.
While this allows an instantaneous prediction of a model class and/or a configuration for a new data set without any time investment, these meta-models are rarely sufficiently accurate in practice, and modern approaches use them mainly to warmstart HPO \citep{FeurerSH15} in a hybrid manner.

\paragraph{Algorithm Configuration}
Algorithm configuration (AC) is the general task of searching for a well-performing parameter configuration of an arbitrary algorithm for a given, finite, arbitrary set of problem instances.
We assume that the performance of a configuration can be quantified by some scoring metric for any given instance \citep{hutter2006performance}. 
Usually, this performance can only be accessed empirically in a black-box manner and comes at the cost of running this algorithm.
This is sometimes called offline configuration for a set of instances, and the obtained configuration is then kept fixed for all future problem instances from the same domain. Alternatively, algorithms can also be configured per instance. 
AC is much broader than HPO, and has a particularly rich history regarding the configuration of discrete optimization solvers \citep{xu2008satzilla}. 
In HPO, we typically search for a well-performing HPC \emph{on a single data set}, which can be seen as a case of per-instance configuration.
The case of finding optimal pipelines or default configurations \citep{pfisterer2018learning} across a larger domain of ML tasks is much closer to traditional AC across sets of instances.
The field of AutoML as introduced in Section~\ref{sec:preproc} originally stems from AC and was initially introduced as the \textit{Combined Algorithm Selection and Hyperparameter Optimization} (CASH) problem \citep{autoweka}.

%#\citet{EggenspergerLHH18} argues that algorithm configuration is a generalization of HPO by optimizing across a set of tasks (e.g., data sets) to find a robust configuration across all of these. To this end, racing (as explained in Section~\ref{sssec:racing}) is used to select configurations by evaluating promising HPCs on many tasks and poorly performing ones to reject early on \citep{HutterHLS09}.

\paragraph{Dynamic Algorithm Configuration}
In HPO (and AC), we assume that th e HPC is chosen \emph{once} for the entire training of an ML model. 
However, many HPs can actually be adapted while training. 
A well-known example is the learning rate of an NN optimizer, which might be adapted via heuristics \citep{KingmaB14} or pre-defined schedules \citep{LoshchilovH17}. 
However, both are potentially sub-optimal. 
A more adaptive view on HPO is called dynamic algorithm configuration \citep{BiedenkappBEHL20}.
This allows a policy to be learned (e.g., based on reinforcement learning) for mapping a state of the learner to an HPC, e.g., the learning rate \citep{DanielTN16}. 
We note that dynamic algorithm configuration is a generalization of algorithm selection and configuration (and thus HPO) \citep{speck_icaps21}.

\paragraph{Learning to Learn and to Optimize}
\citet{ChenHCDLBF17} and \citet{LiM17} considered methods beyond simple HPO for fixed hyperparameters, and proposed replacing entire components of learners and optimizers. 
For example, \citet{ChenHCDLBF17} proposed using an LSTM to learn how to update the weights of an NN. 
In contrast, \citet{LiM17} applied reinforcement learning to learn where to sample next in black-box optimization such as HPO. 
Both of these approaches attempt to learn these meta-learners on diverse sets of applications. 
While it is already non-trivial to select an HPO algorithm for a given problem, such meta-learning approaches face the even greater challenge of generalizing to new (previously unseen) tasks.

%% file: sec_08_conclusion_challenges_and_open_issues.tex
This work has sought to provide an overview of the fundamental concepts and algorithms for HPO in ML.
Although HPO can be applied to a single ML algorithm, state-of-the-art systems typically optimize the entire predictive pipeline -- including pre-processing, model selection, and post-processing. By using a structured search space, even complex optimization tasks can be efficiently optimized by HPO techniques.
This final section now outlines several open challenges in HPO that we deem particularly important. 

\paragraph{General vs. Narrow HPO Frameworks} 
For HPO tools, there is a general trade-off between handling many tasks (such as auto-sklearn~\citep{automlnips}) and a specializing for few and narrow-focused tasks.
The former has the advantage that it can be applied more flexibly, but comes with the disadvantages that (i) it requires more development time to initially set it up and (ii) that the search space is larger such that the efficiency might be sub-optimal compared to a task-specific approach. 
The latter has the advantage that a more specialized search space can lead to a higher efficiency on a specific task but might not be applicable to a different task. When a specialized tool for an HPO problem can be found, it can often preferred to generalized tools.

\paragraph{Interactive HPO}
It is still unclear how HPO tools can be fully integrated into the exploratory and prototype-driven workflow of data scientists and ML experts. 
On the one hand, it is desirable to support these experts in tuning HPs to avoid this tedious and error-prone task. 
On the other hand, this can lead to lengthy periods of waiting that interrupt the workflow of the data scientist, ranging from a few minutes on smaller data sets and simpler search spaces, or for hours or even days for large-scale data and complex pipelines. 
Therefore, the open problems remains of how HPO approaches can be designed so that users can monitor progress in an efficient and transparent manner and how to enable interactions with a running HPO tool in case the desired progress or solutions are not achieved.

\paragraph{HPO for Deep Learning and Reinforcement Learning}
For many applications of reinforcement learning and deep learning, especially in the field of natural language processing and large computer vision models, expensive training often prohibits several several training runs with different HPCs.
Iterative HPO for such computationally extremely expensive tasks is infeasible even with efficient approaches like BO and multifidelity variants.
There are three ways to address this issue. 
First, gradient-based approaches can directly make use of gradient-information on how to update HPs~\citep{maclaurin_icml15a,franceschi_icml17a,thiede-nn19a}. However, gradient-based HPO is only applicable to a few HPs and requires specific gradient information which is often only available in deep NNs.
Second, some models do not have to be trained from scratch by applying transfer learning or few-shot learning, e.g., \citet{finn_icml17a}. This allows cost-effective applications of these models without incredibly expensive training. Using meta-learning techniques for HPO is an option and can also be further improved if gradient-based HPO or NAS is feasible~\citep{franceschi_icml18a,elsken-cvpr20a}.
Third, dynamic configuration approaches allow application of HPO on the fly while training. A very prominent example is population-based training (PBT)~\citep{li-kdd19a}, which uses a population of training runs with different settings and applies ideas from evolutionary algorithms (especially mutation and tournament selection) from time to time while the model training makes progress. The disadvantage is that this method requires a substantial amount of computational resources for training several models in parallel. This can be reduced by combining PBT with ideas from bandits and BO~\citep{parkerholder-neurips20a}.

\paragraph{Overtuning and Regularization for HPO}
As discussed in \ref{ssec:nested}, long HPO runs can lead to biased performance estimators, which in the end can also lead to incorrect HPC selection. It seems plausible that this effect is increasingly exacerbated the smaller the data and the less iterations of resampling the user has configured.\footnote{Issues like imbalance of a classification task and non-standard resampling and evaluation metrics can complicate this matter further.}
It seems even more plausible that better testing of results (on even more separate validation data) can mitigate or resolve the issue, but this makes data usage even less efficient in HPO, as we are already operating with 3-ways splits. 
HPO tools should probably more intelligently control resampling by increasing the number of folds more and more the longer the tuning process runs. However, determining how to ideally set up such a schedule seems to be an under-explored issue.

\paragraph{Interpretability of HPO Process}
Many HPO approaches mainly return a well-performing HPC and leave users without insights into decisions of the optimization process. 
Not all data scientists trust the outcome of an AutoML system due to the lack of transparency \citep{DrozdalWWDYZMJS20}.
Consequently, they might not deploy an AutoML model despite all performance gains. In addition, larger performance gains could potentially be generated (especially by avoiding meta-configuration problems of HPO) if a user is presented with a better understanding of the functional landscape of the objective $\clam$, the sampling evolution of the HPO algorithm, the importance of HPs, or their effects on the resulting performance.
Hence, in future work, HPO may need to be combined more often with approaches from interpretable ML and sensitivity analysis. 

\paragraph{Multi-Criteria HPO and Model Simplicity}
Until here, we mainly considered the scenario of having one well-defined metric for predictive performance available to guide HPO. However, in practical applications, there is often an unknown trade-off between multiple metrics at play, even when only considering predictive performance (e.g., consider an imbalanced multiclass task with unknown misclassification costs). Additionally, there often exists an inherent preference towards simple, interpretable, efficient, and sparse solutions. Such solutions are easier to debug, deploy and maintain, as well as assist in overcoming the \enquote{last-mile} problem of integrating ML systems in business workflows \citep{chui2018notes}.
Since the optimal trade-off between predictive performance and simplicity is usually unknown a-priori, an attractive approach is multi-criteria HPO in order to learn the HPCs of all Pareto optimal trade-offs \citep{horn2016multi, mmpaper}.
Such multi-criteria approaches introduce a variety of additional challenges for HPO that go beyond the scope of this paper.
Recently, model distillation approaches have been proposed to extract a simple(r) model from a complex ensemble \citep{fakoor2020fast}.

\paragraph{Optimizing for Interpretable Models and Sparseness}
A different challenge regarding interpretability is to bias the HPO process towards more interpretable models and return them if their predictive performance is not significantly and/or relevantly worse than that of a less understandable model.
To integrate the search for interpretable models into HPO, it would be beneficial if the interpretability of an ML model could be quantitatively measured. This would allow direct optimization against such a metric \citep{molnar2019quantifying}, e.g.,\ using multi-criteria HPO methods.
% A practical alternative is to construct a manual sequence of preferred models of increasing complexity, to separately tune them and compare their performance. Then an informed choice can be made how much additional complexity is acceptable for a certain amount of improved predictive performance. \citet{hastie_elements_2009} suggest the \enquote{one-standard-error rule} and use the simplest model within one standard error of the performance of the best model. The specific sequence of learners will differ for each use/case but might look something like this: (1) linear model with only a single feature or tree stump, (2) L1/L2 regularized linear model or a CART tree, (3) component-wise boosting of a GAM, (4) random forest, (5) gradient tree boosting, and (6) full AutoML search space with pipelines. If even more focus is put on predictive performance (multi-level), stacking of a diverse set of models can give a final, often rather small, boost \citep{wolpert1992stacked}.

%\paragraph{Multitask HPO}

\paragraph{HPO Beyond Supervised Learning}
This paper discussed HPO for supervised ML.
Most algorithms from other sub-fields of ML, such as clustering (unsupervised) or anomaly detection (semi- or unsupervised), are also configurable by HPs.
In these settings, the HPO techniques discussed in this paper can hardly be used effectively, since performance evaluation, especially with a single, well-defined metric, is much less clear.
%%% some final encouraging words
Nevertheless, HPO is ready to be used.
With the seminal paper by \citet{bergstra12}, HPO again gained exceptional attention in the last decade. 
Furthermore, this development sparked tremendous progress both in terms of efficiency and in applicability of HPO. 
Currently, there are many HPO packages in various programming languages readily available that allow easy application of HPO to a large variety of problems.

%%%

%% file: sec_09_end.tex
\section*{Funding Resources} %Funding Information goes also into submission system
The authors of this work take full responsibilities for its content.
This work was supported by the Federal Statistical Office of Germany; the Deutsche Forschungsgemeinschaft (DFG) within the Collaborative Research Center SFB 876, A3; the German Federal Ministry of Education and Research (BMBF) under Grant No. 01IS18036A; and the Bavarian Ministry for Economic Affairs, Infrastructure, Transport and Technology through the Center for Analytics-Data-Applications (ADA-Center) within the framework of “BAYERN DIGITAL II”.

\section*{Acknowledgments} 
We would like to thank Eyke Hüllermeier, Lars Kothoff, Jürgen Branke and Eduardo C. Garrido-Merch\'an and for their valuable feedback on the manuscript.

%% file: appendix_A_learners_and_search_spaces.tex
\section{Learners and Search Spaces} \label{app:important_ml_algorithms}

In this appendix, we list suggested HP spaces for a variety of popular ML algorithms. All covered learners can be used for both classification and regression.
%, see \citet{hastie_elements_2009} and \citet{murphy_machine_2012}. 
%See Appendix~\ref{app:software} for software implementations of these algorithms in R and Python.
%
% A plethora of ML algorithms for supervised learning exists. 
% In this paper we restrict ourselves to the most prominent representatives of penalized linear models, support vector machines, gradient boosting, random forests, and neural networks in order to discuss a variety of different learners while staying within scope.
% All of them can be used for both classification and regression. For more details on the algorithms, see \citet{hastie_elements_2009} and \citet{murphy_machine_2012}.
%
% The overview has a particular focus on hyperparameters: for each considered algorithm, we discuss its relevant parameters in relation to HPO and give concrete recommendations on how to configure the search space. See Appendix~\ref{app:software} for software implementations of these algorithms in R and Python.
%
In Tables~\ref{tab:hp_knn} to~\ref{tab:hp_ann}, important HP names, package default values, and recommended tuning ranges are given. Although the names and default values relate to the particular R implementation, they can usually be easily matched with names in other implementations.
HP data types are listed as:  logical~(L), integer-valued~(I), real-valued~(R), or categorical~(C).
Some HPs are typically optimized on a logarithmic scale, which yields a higher resolution for smaller values compared to larger ones. %removed reference to main paper% (see Section~\ref{ssec:choosing_pipeline}). 
This can be achieved by using a transformation function before a proposed HP is passed on to the learning algorithm, indicated by the column ``Trafo'' in the HP tables. In this case, the transformation must be applied to the values taken from ``Typical range''. The default values, on the other hand, should be interpreted as default HP values after transformation. 

It should be noted that the selection of learners, HPs, and especially their recommended ranges is somewhat subjective, because
\begin{enumerate*}[(i)]
  \item many learners and implementations exist and we can only list a small subset due to space constraints, and because
  \item it is an open problem which parameters should be tuned and in what ranges. %removed reference to main paper%, see also Section~\ref{ssec:choosing_pipeline} for further discussion.
\end{enumerate*}
The ranges presented here are mostly based on experience and should work well for a wide range of data sets, but -- especially for non-standard situations -- adaptation may be necessary.

\subsection{Nearest Neighbors}
\label{ssec:nearest_neighbors} %Theresa

\paragraph{Concept} 
The $k$-Nearest-Neighbors ($k$-NN) algorithm \citep{cover_nearest_1967} is a simple method where model fitting only consists of storing the data. To predict a new data point, a distance function, e.g.,\ Euclidean distance, to all existing points in the training set is calculated, and the $k$ \enquote{nearest neighbors} are determined.
For classification, posterior probabilities are simply estimated by computing class proportions among the nearest neighbors; for regression, the prediction is a simple average.
%the value of the new point can be predicted by taking the mean of the values of the $k$ nearest neighbors.

$k$-NN can be extended by introducing a kernel function that weights the nearest neighbors according to their distances to the prediction point 
\citep{hechenbichler_weighted_2004}.

\paragraph{Hyperparameters} 
%Important hyperparameters of the $k$-NN algorithm are the distance measure and the number $k$ of neighbors.
See Table~\ref{tab:hp_knn}.
$k$ mainly controls how local the $k$-NN model becomes: small values make $k$-NN more flexible but susceptible to overfitting, while larger values create a smoother albeit potentially underfitted prediction function.
%When using the distance-weighted $k$-NN, the type of kernel function is another hyperparameter.

\begin{table}[!ht]
\begin{adjustbox}{center}
\begin{tabular}{lc>{\raggedleft}p{3cm}rrp{3.8cm}} 
\toprule
    HP & Type & Typical Range & Trafo & Default & Description \\
    \midrule
    \texttt{k} & I & $[\log(1),\log(50)]$ & $\left\lfloor e^x\right\rfloor$ & 7 &  number of nearest neighbors\\
    \texttt{distance} & R & $[1,5]$ & -- & 2 &  parameter $p$ of the Minkowski distance $\lVert x-y \rVert_p$ \\
    \texttt{kernel} & C & \{rectangular, optimal, epanechnikov, biweight, triweight, cos, inv, gaussian, rank\} & -- & optimal &  distance weighting function; ``rectangular'' corresponds to ordinary $k$-NN \\ 
    \bottomrule
\end{tabular}
\end{adjustbox}
\caption{Important HPs for $k$-NN in the \rpkg{kknn} package.}
\label{tab:hp_knn}
\end{table}

\subsection{Regularized Linear Models}
\label{ssec:generalized_linear_models} 

\paragraph{Concept} Methods in this category extend (generalized) linear regression models. The regression coefficients are ``shrunk'' in order to prevent overfitting and therefore improve prediction accuracy. 
%This is called \textit{regularization}.
In the simplest case, the least squares criterion of ordinary regression is extended by a penalty term (using the notation of \citealp{friedman2010}):
\begin{equation}
    %\min_{\mathbf{\theta} \in \mathbb{R}^p} \left\{ \frac{1}{n} \lVert \mathbf{y} - \mathbf{X}\mathbf{\theta}\rVert_2^2 + \lambda P(\mathbf{\theta}) \right\},
    \min_{\mathbf{\theta} \in \mathbb{R}^p} \left\{ \frac{1}{2n} \lVert \mathbf{y} - \mathbf{X}\mathbf{\theta}\rVert_2^2 + \lambda_{\text{reg}} \left( \frac{1}{2}(1-\alpha) \lVert \mathbf{\theta} \rVert_2^2 + \alpha \lVert \mathbf{\theta} \rVert_1 \right) \right\}\textrm{,}
\end{equation}
with $\lambda_{\text{reg}}$ as the \emph{regularization parameter}. For $\alpha = 1$, this is known as the \textit{lasso} \citep{tibshirani_regression_1996}, for $\alpha = 0$ it is \textit{ridge regression} \citep{hoerl_ridge_1970}. For $\alpha \in (0,1)$, the \textit{elastic net} is obtained \citep{zou_regularization_2005}. 

While both the ridge and the lasso penalty cause shrinkage of the regression coefficients, the lasso penalty also features selection by shrinking some coefficients to be \textit{exactly} zero, leading to sparse models. 
In the case of a group of highly correlated features, the lasso tends to pick only one feature from this group. Moreover, in the $p > n$ case, the lasso can select at most $n$ non-zero coefficients. The elastic net overcomes these limitations by combining the lasso with the ridge penalty. 

%If there is a group of highly correlated features, the lasso tends to pick only one feature from the group and does not care which one isselected
%\citet{zou_regularization_2005} combined both lasso and ridge regression into the \textit{elastic net}. Here, the penalty consists of a linear combination of both the lasso and the ridge penalties.
%Like the lasso, the elastic net performs automatic feature selection and thus often leads to sparse models. 
Lasso and ridge regression as well as the elastic net can be extended to a regularized generalized linear model (GLM, see \citet{nelder_generalized_1972} and \citet{dunn_generalized_2018}), simply by plugging in the objective of the GLM instead of least squares. 

\paragraph{Hyperparameters} See Table~\ref{tab:hp_glmnet}. $\alpha \in [0,1]$ controls the respective weighting of the lasso and ridge penalties. 
 $\lambda_{\text{reg}}$ controls the strength of the regularization in general. 

%To obtain a cross-validated model, the function \texttt{cv.glmnet()} can be used, which calls the \texttt{glmnet()} function repeatedly (see Section~\ref{ssec:evaluation_of_ml} for cross-validation). 
% (resp.\ \texttt{predict.cv.glmnet()} for a cross-validated \texttt{glmnet} model).

\begin{table}[h!]
\begin{adjustbox}{center}
\begin{tabular}{lcrrrp{5cm}} 
\toprule
    HP & Type & Typical Range & Trafo & Default & Description \\
    \midrule
    \texttt{s} & R & $[-12, 12]$ & $2^x$ & 1 & controls regularization, but can also be optimized internally \\
    \texttt{alpha}                & R & ${[0, 1]}$ & - & 1 & determines mixture of the lasso and ridge penalties. $\alpha=1$ is lasso \\ \bottomrule
\end{tabular}
\end{adjustbox}
\caption{Important HPs for elastic net in the \rpkg{glmnet} package.}\label{tab:hp_glmnet}
\end{table}

\subsection{Support Vector Machines}
\label{ssec:support_vector_machines} %Theresa

\paragraph{Concept} 
The basic idea of the \textit{Support Vector Machine} (SVM, \citet{boser_training_1992,cortes_support_vector_1995}) can best be understood by considering the case of binary classification: The SVM places a hyperplane in the feature space between both classes
%in order to separate the data points according to the two classes (i.e., the data points are on different sides of the hyperplane, depending on class). The hyperplane is placed 
such that the margin, i.e., the distance of the training points that lie closest to the hyperplane (the \textit{support vectors}), is maximized, allowing for few margin violations.
This geometric interpretation can equivalently be formulated as regularized risk minimization under hinge loss and ridge penalty, meaning the linear SVM is similar to a regularized GLM.
Class prediction for a new data point can be performed by checking which side of the hyperplane the new data point lies on. Posterior probabilities are usually obtained by re-scaling the decision value through a logistic function \citep{platt_probabilistic_1999}. 

In order to achieve non-linear separation, data points are implicitly mapped to a higher dimensional feature space by a kernel function, where the same linear procedure is employed.
%they may become (better) linearly separable.
%The optimization problem can be re-formulated to be expressed in terms of a so-called kernel function
%\[ 
%K\left(\xi,\mathbf{x}^{(j)}\right) = \phi\left(\xi\right)^T \phi\left(\mathbf{x}^{(j)}\right).
%\]

The SVM can be (somewhat non-trivially) extended to multiclass classification as well as to regression (Support Vector Regression, SVR, see \citet{drucker_support_1996,scholkopf_new_2000}).

\paragraph{Hyperparameters} 
See Table~\ref{tab:hp_svm}.
The SVM is mainly influenced by its regularization control parameter, its type of kernel (e.g., linear, polynomial, sigmoid, radial basis function, etc.), and its kernel HPs. 
%must be set as a hyperparameter. Depending on the type of kernel, other hyperparameters have to be set that control the exact functional form of the selected kernel.

\begin{table}[h!]
\begin{adjustbox}{center}
\begin{tabular}{lc>{\raggedleft}p{3cm}rrp{4cm}} \toprule
    HP & Type & Typical Range & Trafo & Default & Description \\
    \midrule
    \texttt{cost} & R & $[-12,12]$ & $2^x$ & 1 & cost of margin violations \\
    \texttt{kernel} & C & \{radial, sigmoid, polynomial, linear\}& -- & radial & type of kernel function \\
    \texttt{degree} & I & $\{2,\ldots,5\}$ & -- & 3 & degree of polynomial kernel (only if kernel is of type polynomial)\\
    \texttt{gamma} & R & $[-12,12]$ & $2^x$ & $1/p$ & controls shape of kernel function, e.g.,\ for radial the inverse width (not needed if kernel is of type linear) \\ \bottomrule
\end{tabular}
\end{adjustbox}
\caption{Important HPs for SVM in the \rpkg{e1071} package (LIBSVM). Note that by default, \texttt{gamma} is set to a value that depends on the number of features $p$.}\label{tab:hp_svm}
\end{table}

\subsection{Decision Trees}
\label{ssec:decision_trees} %Anne-Laure

\paragraph{Concept} Although many variants of decision trees exist, we only focus on the widely used CART algorithm here \citep{breiman1984classification}.
A decision tree is a decision rule constructed from the training set by recursively partitioning (\enquote{splitting}) the space of features in such a way that the resulting nodes are as homogeneous as possible regarding the target variable of interest. 

\paragraph{Hyperparameters} See Table~\ref{tab:hp_cart}. CART is mainly influenced by
\begin{enumerate*}[(i)]
  %\item the type of splits allowed (e.g., only binary splits or multi-splits), 
  \item the splitting criterion / loss function used to select the best feature and split point for each internal node, such as Gini or entropy;
  
  %variable and the best split within this variable at each stage, e.g., the maximization of the Gini impurity reduction, 
  \item the stopping criteria for splitting, which determine the size of the tree and hence regularize it;
  %to further split or to stop the partitioning process or, in the same vein, the criterion used to prune the tree once a maximal tree has been constructed, and 
  \item the procedure used to assign a predicted value to each leaf node. 
  %As a result of this variety, different variants of decision trees have different sets of hyperparameters.
\end{enumerate*}

%However, it can be said that, quite generally, most important hyperparameters of decision trees are related to (iii), i.e., to the stopping/pruning criterion. In a broad sense, they determine the size of the tree. Typical stopping criteria include the maximum number of layers or the maximum number of terminal nodes the tree can have, the minimum number of observations in a parent node or in the resulting child nodes for the split to take place, or the minimum improvement (in terms of the homogeneity of the nodes) a split should generate.

\begin{table}[h!]
\begin{adjustbox}{center}
\begin{tabular}{lcrrrp{3.5cm}} \toprule
    HP & Type & Typical Range & Trafo & Default &  Description \\ \midrule
    \texttt{minsplit} & I & $\{1,\ldots,7\}$ & $2^x$ & 20 & minimum number of observations a parent node must have to be split \\
    \texttt{minbucket} & I & $\{0,\ldots,6\}$ & $2^x$ & $\lfloor \text{minsplit}/3 \rceil$ & minimum  number  of  observations in a leaf node \\
%   \texttt{maxdepth} & I & $\{0,\ldots,10\}$& $2^x$ & 30 & maximum depth of any node of the final tree\\
   \texttt{cp} & R & ${[-4,-1]}$ & $10^{x}$ & 0.01 & complexity parameter that prevents splits that reduce the overall loss by a fraction less than this\\ \bottomrule
\end{tabular}
\end{adjustbox}
\caption{Important HPs for CART in the \rpkg{rpart} package.}
\label{tab:hp_cart}
\end{table}

\subsection{Random Forests}
\label{ssec:ranom_forests} %Anne-Laure 
\paragraph{Concept} 
The random forest (RF) algorithm \citep{breiman2001random} is an ensemble technique. It aggregates a large number of decision trees in order to reduce prediction error and therefore smoothes out prediction variance through bagging. It also further de-correlates the contained trees by expanding them to a large depth and subsampling the candidate features to be considered (\enquote{tried}) at each split.

%combined with high variability of the individual trees. While in bagged decision trees \citep{breiman1996bagging} the trees differ only in the specific bootstrapped samples they are built on, %The number of variables in these randomly selected subsets is often denoted as \texttt{mtry}. 
%Dissimilarities between trees generally lead to more robust ensembles \citep{breiman2001random}, which is the basis of the good performance of random forests compared to single decision trees. 

%The existing variants of RF differ in 
%the type of decision trees used to build the forest and 
%the procedure used to aggregate the predictions made by the individual decision tree.
%the allowed values for the hyperparameters mentioned below and---within the allowed values---the default values. 
%Several reviews can be found in the literature \citep{boulesteix2012overview}.

\paragraph{Hyperparameters} See Table~\ref{tab:hp_rf}.
RFs have a relatively large number of HPs. 
These include the parameters controlling the construction of the single decision trees and also the parameters controlling the structure and size of the forest. 
The latter include the number of trees in the forest and parameters determining its randomness: the number %\texttt{mtry} 
of variables considered as candidate splitting variables at each split, or the sampling scheme used to generate the data sets on which the trees are built (the proportion of randomly drawn observations, and whether they are drawn with or without replacement).
Note that the choice of the number of trees results from a compromise between performance and computation time rather than from the search for an optimal value; except for relatively rare exceptions, employing more trees will lead to better performance at the cost of requiring more computation and memory \citep{probst2017tune}.
The number of trees is thus not a tuneable HP in the classical sense.

RF has been shown to work reasonably well with default parameters for a wide range of applications and may require less tuning compared to other algorithms \citep{delgado2014we, probst2019tunability}. 
However, in some cases, HP tuning can lead to a substantial performance improvement \citep{probst2019hyperparameters}.

\begin{table}[h!]
\begin{adjustbox}{center}
\begin{tabular}{p{2.4cm}cr>{\raggedleft}p{3cm}p{2.7cm}} \toprule
    HP & Type & Typical Range & Default & Description \\
    \midrule
   \texttt{mtry} & I & $\{1,\ldots, p\}$ & $\left\lfloor\sqrt{p}\right\rfloor$ & number of features considered as candidate splitting variables at each split \\
    \texttt{replace} & L & & \texttt{TRUE} & are observations drawn with or without replacement? \\
    \texttt{sample.fraction} & R & ${[0.1,1]}$ & 1 (if replace=\texttt{TRUE}) or 0.632 & proportion of randomly drawn observations\\
    \texttt{num.trees} & I & $\{1,\ldots,2000\}$ & 500 & number of trees\\
    \bottomrule
\end{tabular}
\end{adjustbox}
\caption{In addition to the individual tree parameters in Table~\ref{tab:hp_cart}, the most important HPs for RF in the \rpkg{ranger} package. Note that, by default, \texttt{mtry} is set to a value that depends on the number of features $p$.} 
\label{tab:hp_rf}
\end{table}

\subsection{Boosting}
\label{ssec:gradient_boosting} %Theresa

\paragraph{Concept} 
\textit{Boosting} is an ensemble learning technique and refers to the general idea of improving the prediction performance of a \enquote{weak} learning algorithm by sequentially training multiple models on re-weighted versions of the data and then combining their predictions \citep{schapire_strength_1990,freund_boosting_1995}.

A popular example is the \textit{AdaBoost} algorithm for classification \citep{freund_decision_theoretic_1997}.
%which iteratively updates weights of the training samples. AdaBoost starts by training a model with each training sample weighted equally by $1/n$. The weights of the training samples are then updated such that samples that were misclassified by this model obtain higher weights. In the next step, a second model is trained on the basis of the re-weighted samples, and afterwards the weights of the samples are once again updated according to whether they were misclassified. This procedure is repeated several times. In the end, the predictions of all classifiers are combined, giving higher influence to classifiers with less misclassification error.
\citet{friedman_additive_2000} show that AdaBoost can be reformulated and generalized to \textit{gradient boosting} \citep{friedman_greedy_2001}. Gradient boosting uses a backfitting procedure to fit the next learner, which is added to the ensemble in a greedy stagewise manner by fitting it against the pseudo-residuals of the current ensemble. 
This not only generalizes the boosting principle for arbitrary losses of classification and regression (as long as they are differentiable), but also allows boosting in nearly any supervised ML task, e.g., survival analysis.

%the stagewise fitting of additive models to more robust 
%equivalent to fitting an additive model in a stagewise fashion using an exponential loss function. However, this 
%loss function is not robust in noisy settings. Extending 

%The loss function must be differentiable such that it can be minimized via gradient descent. This procedure can be implemented for both classification and regression. 

In principle, the idea of (gradient) boosting works for any base learner, but the most frequent choices are (shallow) decision trees or (regularized) linear models.

\paragraph{Hyperparameters} See Table~\ref{tab:hp_boost}. The HPs of gradient boosting can be roughly divided into two categories: 
\begin{enumerate*}[(i)]
    \item choice of base learner and its HPs, e.g., HPs of decision trees in Table~\ref{tab:hp_cart}, and 
    \item HPs of the boosting process, such as the choice of the loss function and the number of boosting iterations $M$. 
\end{enumerate*}
Several HPs from both categories are related to regularization, and tuning them is therefore necessary to avoid overfitting, e.g., 
(i) choosing $M$ not too large or using early stopping; 
(ii) adapting the \textit{learning rate}, which controls the contribution of each boosting iteration and interacts with $M$. Typically, a smaller learning rate increases the optimal value for $M$ \citep{friedman_greedy_2001}. 

%Most important are 
%One such hyperparameter is the number of boosting rounds $M$: Overfitting may be prevented by
%Solely focusing on $M$ is usually not the best approach for regularization, and
%However, one should also consider other regularization hyperparameters. 

\begin{table}[h!]
\begin{adjustbox}{center}
\begin{tabular}{p{3cm}crrrp{3cm}} \toprule
    HP & Type & Typical Range & Trafo & Default &  Description \\
    \midrule
    %\texttt{booster} & C & \{gbtree, gblinear, dart\}  & gbtree & type of base model or boosting method \\
    \texttt{eta} &  R & $[-4, 0]$ & $10^x$ & 0.3 & learning rate (also called $\nu$) shrinks contribution of each boosting update \\
    % \texttt{objective} & C & \{reg:squarederror, reg:logistic, binary:logistic, $\ldots$\} & reg:squarederror & loss function \\
   \texttt{nrounds} & I & $\{1,\ldots,5000\}$ & -- &  -- & number of boosting iterations. Can also be optimized with early stopping.\\
    %\texttt{gamma} & R & $[-7,6]$ & $2^x$ & 0 & minimum loss reduction required to make a further partition on a leaf node of the tree\\
    \texttt{max\_depth} & I & $\{1,\ldots,20\}$ & -- & 6 & maximum depth of a tree\\
    \texttt{colsample\_bytree} & R & $[0.1,1]$ & -- & 1 & subsample ratio of columns for each tree \\
    \texttt{colsample\_bylevel} & R & $[0.1,1]$ & -- & 1 & subsample ratio of columns for each depth level \\
    \texttt{lambda} & R & $[-10,10]$ & $2^x$ & 1 & $L_2$ regularization term on weights\\
 
    \texttt{alpha} & R & $[-10,10]$ & $2^x$ & 0 & $L_1$ regularization term on weights\\
    \texttt{subsample} & R & $[0.1,1]$ & -- & 1 & subsample ratio of the training instances\\
   \bottomrule
   \end{tabular}
\end{adjustbox}
\caption{Important HPs for gradient boosting in \rpkg{xgboost} package. The HPs given here assume that a tree base learner is used.}\label{tab:hp_boost}
\end{table}

\subsection{Neural Networks}
\label{ssec:artificial_neural_networks} %Martin 

\paragraph{Concept} Artificial neural networks (NNs) are inspired by biological networks of neurons that communicate with each other by sending action potential signals. 
A NN generally consists of non-linearly transformed weighted sums of inputs, called \textit{neurons}, that are typically organized in layers.
They are trained by stochastic gradient descent variants on small \textit{batches} of training samples.
Modern NNs are adapted to many data modalities, such as, e.g., computer vision and natural language processing tasks. Pre-defined architectures for a variety of these tasks exist.

\paragraph{Hyperparameters} See Table~\ref{tab:hp_ann}. The HPs of NNs can be split into two categories:
\begin{enumerate*}[(i)]
    \item optimization and regularization parameters, and 
    \item architectural parameters that define the type, amount, structure, and connections of neurons. 
\end{enumerate*}

For the first category of HPs, the optimization techniques introduced in this paper can be used straightforwardly.
The second category of HPs is much more difficult to optimize, and the field of \textit{neural architecture search} (NAS) provides highly customized strategies to find well-working network architectures \citep{elsken_neural_2019}.
However, we can define a function that can construct a network architecture given specific simple HPs while not being able to construct all possible network architectures.
A simple example would be an HP that controls the number of hidden layers.
However, especially for non-tabular data, NNs usually have more complex architectures.
In such cases, one could parameterize the layout of a single cell within an architecture of stacked cells \citep{zoph_learning_2018}.

\begin{table}[h!]
\begin{adjustbox}{center}
\begin{tabular}{p{2.5cm}crrrp{3.4cm}} \toprule
    HP & Type & Typical Range & Trafo & Default & Description \\
    \midrule
    \texttt{learning\_rate} & R & ${[-5,0]}$ & $10^x$ & -- & gradient descent learning rate, also called $\eta$ or $\alpha$\\
    %\texttt{optimizer} & C & \{sgd, rmsprop, adam\} & -- & -- & optimization algorithm\\
    \texttt{regularizer\_l2} & R & ${[-7,-4]}$ & $10^x$ & -- & L$_2$ penalty on weights.\\
    \texttt{epochs} & I & $\{1,\dots\}$ & -- & -- & Very problem dependent, usually optimized with early stopping.\\ \bottomrule
    % FIXME (\{1,\dots\})
    % [FD:] Schwer zu verstehen ;). Wie geht man vor?
    % Jakob: Irgendwie inkonsequent, dass wir hier keine obere Grenze angeben.
    % Jakob: Aber bzgl. "wie geht man vor" ist doch "early stopping" die antwort direkt daneben?
    
    %\texttt{batch\_size} & I & $\{0,\ldots, 11\}$ & $2^x$ & ?? &  training batch size\\ \bottomrule
\end{tabular}
\end{adjustbox}
\caption{Optimizer HPs for a NN, as used in the \rpkg{keras} package. Unlike other methods presented here, NNs in \rpkg{torch} and \rpkg{keras} are constructed programmatically, so their configuration in many regards does not consist of HPs in the form of explicit function arguments.}
% FIXME:
% Marius Lindauer: That's a surprisingly short list of hyperparameters. In fact, there is a bunch of regularization techniques with many hyperparameters you could apply here. See for example https://openreview.net/pdf?id=2d34y5bRWxB
% bernd.bischl: we kept this delibaretly short. vielleicht zu sehr,
% bernd.bischl: auf jeden fall etwas zitieren, wo mehr dazu drin steht
% bernd.bischl: MB
\label{tab:hp_ann}
\end{table}

%% file: appendix_B_preprocessing.tex
\section{Preprocessing}\label{app:preprocessing}

Before applying a learner, it may be helpful or even necessary to modify the data before training in a step called \emph{preprocessing}.
Preprocessing serves mainly two purposes:
\begin{enumerate*}[(i)]
    \item it either transforms the data to fit the technical requirements of a specific learner, e.g., by converting categorical features to numerical features or by imputing missing values; or %which the learner cannot handle,
    \item it modifies the data to improve predictive performance, e.g., by bringing features on a common scale or by extracting components of a date column.
\end{enumerate*}
%In the following, the most frequently used preprocessing methods are briefly presented.

%Preprocessing methods are not learning algorithms themselves, but because they are used in combination with learning algorithms, they do add additional dimensions to the HP search space: both by adding their own HPs to the space, and potentially by adding the option of whether and what preprocessing is done to begin with.

\subsection{Missing Data Imputation}
\label{sec:imputation}
Imputation describes the task of replacing missing values with feasible ones, since many ML algorithms do not natively handle missing values.
%,serves two goals:
%\begin{enumerate*}[(i)]
%    \item enable a learner to process the data, since many ML algorithms do not handle missing %values natively, and
    %\item allow to draw inference from the data, e.g.\ extract estimates of the effects from the %fitted model.
%\end{enumerate*}
%Our focus is on the first goal. 
%The reason for missingness is often unknown in practice, and 
In practice, data is often not missing fully at random, and therefore the fact that part of the data is missing itself might be informative for predicting the target label. 
Hence, it is generally recommended to preserve the information of whether a certain value has been imputed.
%the missingness information while imputing missing values for ML.
%
For categorical features, we often simply introduce a new level for missing values.
For numerical or integer features, the simplest techniques impute with a constant mean or median value per feature column or sample from the empirical univariate distribution of the feature.
To preserve the missing information in this case, an additional binary dummy column 
%(which is 1 if the value was missing before imputation and 0 otherwise) 
can be added per imputed feature column.
When tree-based methods are used, it is often preferred to impute missing values with very large or very small values outside the range of the training data \citep{Ding10a}, 
as the tree can then split between available and imputed data, thereby preserving the information about the fact that data is missing without the need for an extra column.
%An additional binary column is not required for this imputation method.
Much more elaborate methods exist, often based on multiple imputation, but these are mainly geared towards ensuring proper inference. 
For purely predictive tasks, it is still somewhat unclear how beneficial these more elaborate imputation methods are. 
More detailed discussions are available in \citet{Garciarena2017,Jadhav2019,Woznica_2020}.

%A typical hyperparameter of the imputation process would be the choice of the imputation strategy or whether to include the additional binary column that indicates missingness.

%such as fitting machine learning models on available data based on the complete observations of other features. 
%A popular imputation method used in some fields is Multiple Imputation by Chained Equations (MICE) \citep{White2010}.
%Instead, they are of practical importance for goal~(ii) where inference is drawn from the data.
%It should be mentioned that in this case it is often assumed that values are \textit{missing completely at random} (MCAR, c.f.\ \citet{Little1988}), which might be violated.
%Furthermore, analysis of effect estimates and corresponding $p$-values requires special caution if imputation is involved as  many imputation methods have a biasing, variance-minimizing effect.

\subsection{Encoding of Categorical Features}
\label{sec:featenc}
Many ML algorithms operate on purely numerical inputs, and hence, we often need an additional numerical encoding procedure for categorical features.
%It is therefore often necessary to transform non-numeric features into a numerical representation. 
%While this can be done in different ways, 
\textit{Dummy encoding}, also called \textit{one-hot encoding}, is arguably the most popular choice. %encoding.
It replaces a categorical $k$-level feature with $k$ or $k-1$ binary indicator features. 
The latter case avoids multicollinearity of the newly created features by using the $k$-th level as a reference level, encoding it by setting all indicator features to $0$.
For data sets with many categorical features and data sets with categorical features with many levels (\textit{high-cardinality} features), this leads to a considerable increase in dimensionality, resulting in increased runtime and memory requirements.
%the number of features, resulting in computational difficulties or even rendering a model fit mathematically infeasible. 
In such cases, methods like \textit{impact encoding} -- also called \textit{target encoding} \citep{sweeney_transformation_1972} -- replace the categorical feature with only a single numeric feature, based on the average outcome value for all observations for which that level has been observed. \citet{pargent2021regularized} have shown that target encoding using generalized linear mixed models (GLMMs) \citep{micci2001preprocessing} tends to outperform other common encoding methods in settings where categorical feature cardinality is high.

Problems arise if factor levels occur during prediction that have not been encountered during training.
To some extent, this can be addressed by merging infrequent factor levels to a new level with value \enquote{other}, or by stratifying on the levels during resampling, but can never be completely ruled out.
Since the new level is unknown to the model, it can be considered a special case of missing data and could therefore be handled by imputation.
%Accordingly, the new factor level can be replaced by following similar strategies as for imputation, e.g.\ sampling from seen levels.
%For impact encoding, the value for the new factor level is set to the mean impact value.
%The \rpkg{mlr3pipelines} package \citep{mlr3pipelines} offers such strategies to make models robust against this occurrence.
A typical HP of the feature encoding process would be the threshold that decides when to perform target encoding instead of dummy encoding, depending on the number of levels of a feature.

Several benchmarks of encoding methods have been performed, typically with different emphasis on setting, such as the work by \citet{pargent2021regularized} and the references contained therein.

% FIXME
% bernd.bischl: janek. du wolltest hier nochmal schauen, vielleicht muss man den threshold an sich auch nochmal in 2 sätzen erklären
% bernd.bischl: BB

% PARAMS
% TODO: Janek: Trade Off zwischen globalen lokalem mittlerwert

% \subsection{Target Transformation}

% Many preprocessing methods only concern feature data, but it may also be beneficial to transform target values. 
% % The predictions have to be transformed inversely to fit the original data.
% % If the outcome value spans many orders of magnitude, it may, e.g., be useful to log-transform target values. 
% For example, if the distribution of the numeric target variable looks positively skewed (i.e., has a long right tail), a log transformation can make the target more normally distributed.
% Similarly, a binary classification problem can be turned into a regression problem by encoding the target as $\pm{}1$.

% A fact to keep in mind is that transforming the target value often leads to a method that optimizes for a transformed loss function. 
% For example, training a model that minimizes the MSE on a log-transformed target will not lead to an optimal model regarding the MSE on the naively exponentiated predictions. % A possible remedy is to add a correction value based on the estimated predicted uncertainty.

\subsection{Feature Filtering}
\label{sssec:fitering}
Feature filtering is used to select a subset of features prior to model fitting.
Subsetting features can serve two purposes: 
\begin{enumerate*}[(i)]
  \item either to reduce the computational cost, or 
  \item to eliminate noise variables that may deteriorate model performance.
 \end{enumerate*}
Feature filters often assign a numerical score to each feature, usually based on simple dependency measures between each feature and the outcome, e.g.,
%serving as a proxy for how much a feature contributes to the model.
mutual information or correlation.
%or the variable importance obtained by training a random forest.
The 
%threshold for the score or the 
percentage of features to keep is typically an HP.
Note that feature filtering is one of several possibilities to perform feature selection. See \citet{Guyon2003} for a description of feature wrappers and embedded feature selection methods.

See \citet{wah2018feature} for a benchmark assessing filter methods applied to simulated datasets and the correctness of selected features. See also \citet{bommert2020benchmark} as well as \citet{xue2015comprehensive} for a benchmark assessing filters with respect to predictive performance as well as secondary aspects of interest such as runtime. 
\citet{mmpaper} evaluates a variety of feature filters with respect to their similarity and their suitability for constructing a feature filter ensemble, which can subsequently be tuned for a data set at hand with BO.

%Feature filtering is just an approximate process because the measures cannot reflect the true importance of a feature.
%The true contribution of a feature towards the generalization error can only be obtained by evaluating the ML algorithm, which yields a combinatorial optimization problem and is discussed briefly in Section~\ref{ssec:?} % TODO: Find appropriate section

% ML: Out of scope for this paper?
% JR: Move this part to section ssec:?
% In Section~\ref{sssec:fitering} we introduced filter methods that reduce the number of features based on a proxy measure.
% Instead, we can directly try to find the optimal subset of features that maximizes the generalization error using so called \emph{wrapper methods}.
% Therefore we choose a certain subset of features and evaluate the performance of this subset and the ML algorithm through resampling.
% Finding the subset that optimizes the generalization error yields a combinatorial optimization problem.
% Often this problem is solved approximately through forward or backward selection, or through genetic optimization.
% Combining the wrapper approach and HPO yields a joint optimization problem \citep{mmpaper}.
%% FIXME
%% bernd.bischl: drauf achten, dass das paper noch zitiert wird
%% bernd.bischl: MB
%% bernd.bischl: zitieren da wo wir multicrit erwähnen
% Ml algorithms that embed feature selection like decision trees (Section~\ref{sssec:decision_trees}) or lasso (Section~\ref{sssec:generalized_linear_models}) often render additional feature selection unnecessary.

\subsection{Data Augmentation and Sampling}
\label{sssec:data_augmentation}
Data augmentation has the goal of improving predictive performance by adding or removing rows of the training data.
A common use-case for data augmentation is classification with imbalanced class labels.
Here, we can add additional observations by oversampling, i.e., repeating observations of the minority class.
A more flexible alternative is SMOTE (``Synthetic Minority Oversampling Technique'', \citealp{Chawla2002}),
where convex combinations of observations from the minority class are added to the training data.
When the available dataset is very large, it is also possible to undersample in order to correct for class imbalance, i.e., filtering out observations of the majority class.
Also, subsampling the data set as a whole is an old trick to reduce training time during HPO, and modern multifidelity approaches exploit this in a more principled manner.
Sampling and data augmentation methods differ from many other preprocessing methods in that they only affect the training phase of a model. 
Typical HPs include the under-, over-, or subsampling ratio or the number of observations to use for the convex combinations in SMOTE.

For a more thorough description of how to treat imbalanced data in ML, see \citet{krawczyk2016learning}. Current challenges and areas of research are described in \citet{he2013imbalanced}.

\subsection{Feature Extraction}

A crucial step in improving predictive performance of a model is \textit{feature extraction} or \textit{feature engineering}.
The key idea is to add new features to the data that the model can exploit better than the original data.
This is often both domain-specific and model-specific, and therefore is hard to generalize.
For linear models, for example, adding polynomial terms
%simple feature transformations like $\tilde x_1 = x_1^2$, $\tilde x_2 = x_2^2$ and $\tilde x_3 = x_1 \cdot x_2$ 
can provide a better view on the data, effectively allowing the model to capture non-linear effects and higher-order interactions.
Simple tree-based methods might profit from methods like principal component analysis, which rotate all or a subset of numerical features to align feature dimensions with the (orthogonal) directions of major variance. 
On the other hand, such simple extractions techniques very often do not provide competitive benefit in terms of predictive performance when many non-parametric models are included in model selection and HPO \citep{bischl2014benchmarking}.

More beneficial is often domain-specific feature extraction for more complex data types. 
For example, for functional data features, where a collection of columns represents a curve, one can either extract generic characteristics (mean value, maximum value, average slope, number of peaks, etc.)
or run domain-specific extraction, e.g., we might extract mel-frequency cepstral coefficients (MFCCs)  \citep{sahidullah2012design} for an audio signal.
For a practical guide to using feature extraction in predictive modeling, see e.g. \citet{kuhn2019feature}.

All extracted features can either be used instead or in addition to the original features.
As it is often highly unclear how to perform ideal feature extraction for a specific task, it seems particularly attractive if an ML software system allows the user to embed custom extraction code into the preprocessing with exposed custom HPs so that a flexible feature extraction piece of code can be automatically configured by the tuner in a data-dependent manner.

%% file: appendix_C_evaluation_metrics.tex
\newgeometry{margin=2.3cm}
\begin{landscape}
\section{Evalution Metrics}\label{app:evaluation}
\begin{table}[h!]
\small
%\begin{tabular}{ll >{\raggedleft}p{0.8cm}@{,\hspace{0.15em}}>{\raggedright}p{0.8cm} p{7cm}}
\begin{tabular}{llcc p{9cm}}
\toprule
Name & Formula & Direction & Range & Description \\
\midrule
\multicolumn{5}{l}{\textbf{Performance measures for regression}}\\[0.15em]
Mean Squared Error (MSE)        & $\frac{1}{\ntest} \sum_{i = 1}^{\ntest} \left(\yi - \yih\right)^2 $  & min & $[0, \infty)$ &  Mean of the squared distances between the target variable $y$ and the predicted target $\yh$.\\
Mean Absolute Error (MAE)       & $\frac{1}{\ntest} \sum_{i = 1}^{\ntest} \left\lvert\yi - \yih\right\rvert$ & min & $[0, \infty)$  & More robust than MSE, since it is less influenced by large errors. \\
$R^2$        & $1 - \frac{\sum_{i = 1}^{\ntest} \left(\yi - \yih\right)^2}{\sum_{i = 1}^{\ntest} \left(\yi - \bar{y}\right)^2}$    & max & $(-\infty, 1]$  & Compare the sum of squared errors (SSE) of the model to a constant baseline model. \\ 
\multicolumn{5}{l}{\textbf{Performance measures for classification based on class labels}}\\[0.2em]
Accuracy (ACC)            & $\frac{1}{\ntest} \sum_{i = 1}^{\ntest} \I_{\left\{\yi = \yih\right\}}$                                   & max & $[0, 1]$      & Proportion of correctly classified observations. \\
Balanced Accuracy (BA)    & $\frac{1}{g} \sum_{k=1}^g \frac{1}{n_{\text{test},k}} \sum_{\yi: \yi = k} \I_{\left\{\yi = \yih\right\}}$ & max & $[0, 1]$      & Variant of the accuracy that accounts for imbalanced classes. \\

Classification Error (CE) & $\frac{1}{\ntest} \sum_{i = 1}^{\ntest} \I_{\left\{\yi \neq \yih\right\}}$                                & min & $[0, 1]$      & $\mathrm{CE}=1 - \mathrm{ACC}$ is the proportion of incorrect predictions. \\
ROC measures              & $\mathrm{TPR} = \frac{\mathrm{TP}}{\mathrm{TP} + \mathrm{FN}}$                                                & max & $[0, 1]$      & True Positive Rate: how many observations of the positive class 1 are predicted as 1?\\
                          & $\mathrm{FPR} = \frac{\mathrm{FP}}{\mathrm{TN} + \mathrm{FP}}$                                                & min & $[0, 1]$      & False Positive Rate: how many observations of the negative class 0 are falsely predicted as 1? \\
                          & $\mathrm{TNR} = \frac{\mathrm{TN}}{\mathrm{TN} + \mathrm{FP}}$                                                & max & $[0, 1]$      & True Negative Rate: how many observations of the negative class 0 are predicted as 0?\\
                          & $\mathrm{FNR} = \frac{\mathrm{FN}}{\mathrm{TP} + \mathrm{FN}}$                                                & min & $[0, 1]$      & False Negative Rate: how many observations of the positive class 1 were falsely predicted as 0? \\
                          & $\mathrm{PPV} = \frac{\mathrm{TP}}{\mathrm{TP} + \mathrm{FP}}$                                                & max & $[0, 1]$      & Positive Predictive Value: how likely is a predicted 1 a true 1?\\
                          & $\mathrm{NPV} = \frac{\mathrm{TN}}{\mathrm{FN} + \mathrm{TN}}$                                                & max & $[0, 1]$      & Negative Predictive Value: how likely is a predicted 0 a true 0?\\
$F_1$                     & $2 \frac{\mathrm{PPV} \cdot \mathrm{TPR}}{\mathrm{PPV} + \mathrm{TPR}}$                                       & max & $[0, 1]$      & $F_1$ is the harmonic mean of PPV and TPR. Especially useful for imbalanced classes.\\
Cost measure              & $\sum_{i = 1}^{\ntest} C(\yi, \yih)$                                                                          & min & $[0, \infty)$ & Cost of incorrect predictions based on a (usually non-negative) cost matrix $C \in \R^{g,g}$.\\
\multicolumn{5}{l}{\textbf{Performance measures for classification based on class probabilities}}\\[0.2em]
Brier Score (BS)          & $\frac{1}{\ntest} \sum_{i = 1}^{\ntest} \sum_{k = 1}^g \left( \pikxih - \sigma_k(\yi) \right)^2 $             & min & $[0, 1]$      & Measures squared distances of probabilities from the one-hot encoded class labels. \\
Log-Loss (LL)             & $\frac{1}{\ntest} \sum_{i = 1}^{\ntest} \left( - \sum_{k=1}^g \sigma_k(\yi) \log(\pikxih) \right) $           & min & $[0, \infty)$ & A.k.a. Bernoulli, binomial or cross-entropy loss \\

AUC                       &                                                                                                               & max & $[0, 1]$      & Area under the ROC curve.\\
\bottomrule
\multicolumn{5}{p{22.5cm}}{\footnotesize{$\yih$ denotes the predicted label for observation $\xi$. ACC, BA, CE, BS, and LL can be used for multi-class classification with $g$ classes. For AUC, multiclass extensions exist as well. The notation $\I_{\{\cdot\}}$ denotes the indicator function. $\sigma_k(y) = \I_{\{y = k\}}$ is 1 if $y$ is class $k$, 0 otherwise (multi-class one-hot encoding). $n_{\text{test},k}$ is the number of observations in the test set with class $k$. $\pikxh$ is the estimated probability for observation $\xi$ of belonging to class $k$. $TP$ is the number of true positives (observations of class 1 with predicted class 1), $FP$ is the number of false positives (observations of class 0 with predicted class 1), $TN$ is the number of true negatives (observations of class 0 with predicted class 0), and $FN$ is the number of false negatives (observations of class 1 with predicted class 0).}}
\end{tabular}
\caption{Popular performance measures used for ML, assuming an arbitrary test set of size $\ntest$.}
\label{tab:measures}
\end{table}
\end{landscape}
\restoregeometry

%% file: appendix_D_01_software_r.tex
\subsection{Software in R}\label{ssec:software_r}
\subsubsection{Important Machine Learning Algorithms} % (fold)
\label{sub:important_machine_learning_algorithms}

Here we present a curated list of learners that work well in general and are well integrated with ML frameworks in R (see Section~\ref{sub:ml_frameworks}). 
For a comprehensive list of R-packages that offer ML models, categorized by the type of model implemented, see the ``CRAN Task View'' for ML and Statistical Learning\footnote{\url{https://cran.r-project.org/view=MachineLearning}}. 
%%%

\paragraph{$k$-NN} 
The basic $k$-NN algorithm is implemented in the \rpkg{class} package that comes installed with R. The distance-weighted $k$-NN is implemented in the \rpkg{kknn} package \citep{kknnpackage}. For large data sets, one may prefer the approximate algorithms \citep{arya1998optimal} in the \rpkg{FNN} package \citep{fnnpackage}.

\paragraph{Regularized Linear Models} 
The elastic net is implemented in the \rpkg{glmnet} package \citep{friedman2010},
which also offers an internal, fast optimization of its regularization HPs 
via the \texttt{cv.glmnet()} function. Both \rpkg{biglasso} \citep{zeng2018biglasso} and \rpkg{LiblineaR} implement regularized linear models specifically optimized for large datasets.

%Models with very close $\lambda_{\text{reg}}$ hyperparameter values also have very similar coefficient values. This can be used to fit many models with a grid of $\lambda_{\text{reg}}$ values simultaneously \citep{efron2004least}, which the \texttt{glmnet} function applies for quick optimization of the $\lambda_{\text{reg}}$ hyperparameter. This function by default fits multiple models for different $\lambda_{\text{reg}}$ values, chosen on a logarithmic scale from $\lambda_{\text{max}}$ (determined from data, lowest $\lambda_{\text{reg}}$ for which all coefficients are zero) down to $\lambda_{\text{max}} \times r_{\text{min}}$ ($r_{\text{min}}$ being 0.0001 if $n \geq p$, else 0.01). 
%The hyperparameter \texttt{s} is used to control $\lambda_{\text{reg}}$ during prediction with the function \texttt{predict.glmnet()}.
%The optimal value for \texttt{s} can also be determined automatically by letting \rpkg{glmnet}'s function \texttt{cv.glmnet()} perform an internal cross-validation.
%The value for \texttt{s} is then either set to the best $\lambda_{\text{reg}}$ found during the cross-validation, or to the largest $\lambda_{\text{reg}}$ that is not worse than the best $\lambda_{\text{reg}}$ by more than the resampling standard error.
%The latter is an established rule of thumb and the default in \texttt{predict.cv.glmnet()}.

\paragraph{Support Vector Machines} 
A frequently used implementation of the SVM for R can be found in the \rpkg{e1071} package \citep{meyer2019}, which uses the LIBSVM library internally \citep{chang2011}. \rpkg{kernlab} \citep{kernlabpkg} is a more comprehensive package with an emphasis on flexibility.

\paragraph{Decision Trees}
The basic CART algorithm is implemented in the \rpkg{rpart} R package \citep{therneau2019}, while a more flexible implementation of decision trees is included in \rpkg{partykit} \citep{hothorn2015partykit}
% (which implements the CART algorithm for constructing trees), the default minimum number of observations a parent node must have to be split (denoted as minsplit) is 20, while the default minimum number of observations in a child node (denoted as 'minbucket') is minsplit$/3$. 

\paragraph{Random Forests}
Three of the most widely used R implementations of RF are provided in the packages
\begin{enumerate*}[(i)]
    \item \rpkg{randomForest}, implementing the original version \citep{breiman2001random},
    \item \rpkg{party}, implementing (unbiased) conditional inference forests \citep{hothorn2006unbiased}, and
    \item \rpkg{ranger}, providing a newer implementation of RF, including additional variants and a wide range of options, optimized for use on high-dimensional data \citep{wright2018fast}.
\end{enumerate*}

\paragraph{Boosting}
Popular implementations of gradient boosting with tree base learners can be found in \rpkg{gbm} \citep{gbm2020} and implemented more efficiently in \rpkg{xgboost} \citep{chen2016,chen2020}.
Gradient boosting of generalized linear and additive models is implemented in \rpkg{xgboost} as well as in \rpkg{mboost} \citep{mboost2010}, the latter being more efficiently implemented in \rpkg{compboost} \citep{schalk2018compboost}. 

\paragraph{Artificial Neural Networks}
Modern and fast implementations of NNs are provided by the \rpkg{keras} \citep{rkeras} and \rpkg{torch} \citep{rtorch} packages. 
\rpkg{torch} is more low-level than \rpkg{keras}, and some properties that are explicit HPs of \rpkg{keras} need to be implemented by the user through R code in \rpkg{torch}.

% subsection important_machine_learning_algorithms (end)

%\paragraph{Black-Box Optimizers}
%A categorized overview of R packages for optimization can be found in the CRAN task view for optimization\footnote{\url{https://cran.r-project.org/view=Optimization}}.
%Note that only a very small fraction of the listed optimizers meet the requirements of hyperparameter optimization, i.e., black-box optimization, possibly with support for mixed numeric and categorical search spaces.
%In the remaining candidates, there are some general-purpose optimizers that can be applied to hyperparameter optimization by manually writing an objective function that does performance evaluation through resampling.
%One popular example is the \rpkg{irace} package \citep{lopez_2016} which implements iterated F-racing.
%When using an optimizer not specifically geared towards hyperparameter optimization, the burden of implementing the evaluation function lies on the user.
%In most of these cases, it is necessary to write a target function that accepts a hyperparameter configuration~$\lambdav$ as input and does resampling to calculate an aggregated performance score.
%Implementing this correctly requires paying attention to many details, especially juggling with the indices of inner and outer resampling sets, which has proven to be quite error-prone in the past.
%We therefore strongly advise the use of one of the existing frameworks for machine learning -- not only inside the target function to perform the resampling -- but generally for tuning.

\subsubsection{Black-Box and HP Optimizers}

\paragraph{Evolution Strategies}
Popular R packages that implement different ES are \rpkg{rgenoud} \citep{walter2011rgenoud}, \rpkg{cmaes} \citep{trautman2015cmaes}, \rpkg{adagio} \citep{borchers2018adagio}, \rpkg{DEoptim} \citep{mullen2011deoptim}, \rpkg{mco} \citep{mersmann2014mco}, and \rpkg{ecr} \citep{bossek2017ecr} and \rpkg{mosmafs} \citep{mmpaper}.

\paragraph{Bayesian Optimization}
\rpkg{tune}, \rpkg{rBayesianOptimization} \citep{rBayesianOptimizationpkg} and \rpkg{DiceOptim} \citep{roustant2012dicekriging} are R packages that implement BO with Gaussian processes and for purely numerical search spaces.
\rpkg{mlrMBO} \citep{bischl2017mlrMBO} and its successor \href{https://github.com/mlr-org/mlr3mbo}{\texttt{mlr3mbo}} offer to construct a flexible class of BO variants with arbitrary regression models and acquisition functions. These packages can work with mixed and hierarchical search spaces.
%HP:;  hyperparameter spaces by performing missing value imputation on inactive hyperparameter values when fitting the surrogate model.

\paragraph{Other Methods}
Hyperband is implemented in the package \rpkg{mlr3hyperband} \citep{gruber2019hyperband}. Iterated F-racing is provided by \rpkg{irace} \citep{lopez_2016}.

\subsubsection{ML Frameworks} % (fold)
\label{sub:ml_frameworks}

% subsection ml_frameworks (end)
There are many technical difficulties and intricacies to attend to when practically combining a black-box optimizer with different ML algorithms, especially pipelines, to perform HPO.
Although it would be possible to use one of the algorithms shown above and write an objective function that performs performance evaluation, this is not recommended. 
Instead, one of several ML frameworks should be used. These frameworks simplify the process of model fitting, evaluation, and optimization while managing the most common pitfalls and providing robust and parallel technical execution. 
In the following paragraphs, we summarize the tuning capabilities of the most important ML frameworks for R.
The feature matrix in Table~\ref{tab:feat_mat_software} gives an encompassing overview of the implemented capabilities.

\begin{table}[ht]
\centering
\footnotesize
\begin{tabular}{@{}l|ccccc@{}}
\toprule
                        & \rpkg{mlr}                  & \rpkg{mlr3} & \rpkg{caret} & \rpkg{tidymodels} & \rpkg{h2o} \\ \midrule
Tuners                  &                             &             &              &                   &            \\
- Random Search         & \cmark                      & \cmark      & \cmark       & \cmark            & \cmark     \\
- Grid Search           & \cmark                      & \cmark      & \cmark       & \cmark            & \cmark     \\
- Evolutionary          & \cmark                      & \cmark      & \xmark       & \xmark            & \xmark     \\
- Bayesian Optimization & \cmark                   & \cmark    & \xmark       & \cmark     & \xmark     \\
- Hyperband             & \xmark                      & \cmark      & \xmark       & \xmark            & \xmark     \\
- Racing                & \cmark                      & $\ast$      & \cmark & \cmark     & \xmark     \\
- Other Tuners          & NSGA2, SA                   & NLOpt, SA   & -            & SA                & -         \\
\midrule
Search Space            &                             &             &              &                   &            \\
- Transformations       & \cmark                      & \cmark      & \xmark       &    \cmark              & \cmark           \\
- Default Search Spaces & \cmark                      & \xmark      & i     &    \cmark              & \cmark     \\
\midrule
Joint Preproc Tuning    & \cmark                      & \cmark      & \xmark       & \cmark            & \cmark     \\
%Sub-model trick         & \cmark                      & \cmark      & \xmark       & \cmark            & \xmark     \\
\bottomrule
\multicolumn{6}{l}{\footnotesize{$\ast$: not yet released on CRAN but prototype available, i: see description in main text.}}\\
\end{tabular}
\caption{Feature matrix of tuning capabilities of R's ML frameworks.
%See the paragraphs on the respective frameworks for more details %JR: Raus da klar.
}
\label{tab:feat_mat_software}
\end{table}

\paragraph[mlr]{\rpkg{mlr}} \citep{bischl_mlr_2016} supports a broad range of optimizers: random search, grid search, CMA-ES via \rpkg{cmaes},  BO via \rpkg{mlrMBO} \citep{bischl2017mlrMBO}, iterated F-racing via \rpkg{irace}, NSGA2 via \rpkg{mco}, and simulated annealing via \rpkg{GenSA} \citep{xiang_2013}.
Random search, grid search, and NSGA2 also support multi-criteria optimization. 
Arbitrary search spaces can be defined using the \rpkg{ParamHelpers} package, which also supports HP transformations.
%Default hyperparameter spaces for many popular ML algorithms are given in the \rpkg{mlrHyperopt} package.
HP spaces of preprocessing steps and ML algorithms can be jointly tuned with the \rpkg{mlrCPO} package.
%Furthermore, it is possible to tune over multiple ML algorithms simultaneously using the \enquote{ModelMultiplexer} where the choice of the currently active learning algorithm becomes a hyperparameter. 

\paragraph[mlr3]{\rpkg{mlr3}} \citep{lang_mlr3_2019} superseded \rpkg{mlr} in 2019. This new package is designed with a more modular structure in mind and offers a much improved system for pipelines in \rpkg{mlr3pipelines} \citep{mlr3pipelines}.
%\citep{lang_mlr3_2019} itself does not ship with tuners, but the accompanying package
\rpkg{mlr3tuning} implements grid search, random search, simulated annealing via \rpkg{GenSA}, CMA-ES via \rpkg{cmaes}, and non-linear optimization via \rpkg{nloptr} \citep{johnson2014nlopt}. 
Additionally, \rpkg{mlr3hyperband} provides Hyperband for multifidelity HPO, and the \href{https://github.com/mlr-org/miesmuschel}{\texttt{miesmuschel}} package implements a flexible toolbox for optimization using ES.
BO is available through \rpkg{mlrMBO} when the \rpkg{mlrintermbo} compatibility bridge package is used. 
A further extension for BO is currently in development\footnote{\url{https://github.com/mlr-org/mlr3mbo}}.
\rpkg{mlr3} uses the \rpkg{paradox} package, a re-implementation of \rpkg{ParamHelpers} with a comparable set of features.
% Multiplexing multiple ML algorithms or pipelines is possible using the \rpkg{mlr3pipelines} package.

\paragraph[caret]{\rpkg{caret}} \citep{kuhn2008} ships with grid search, random search, and adaptive resampling (AR) \citep{kuhn2014futility} -- a racing approach where an iterative optimization algorithm favors regions of empirically good predictive performance.
Default HPs to tune are encoded in the respective call of each ML algorithm. However, if forbidden regions, transformations, or custom search spaces are required, the user must specify a custom design to evaluate.
Embedding preprocessing into the tuning process is possible.
However, tuning over the HPs of preprocessing methods is not supported.
% Hyperparameter transformations are not supported. %JR: Raus, da widerspruch zu Satz 2

\paragraph[tidymodels]{\rpkg{tidymodels}} \citep{kuhn2020} is the successor of \rpkg{caret}.
%In comparison to its predecessor, \rpkg{tidymodels} has caught up to \rpkg{mlr} or \rpkg{mlr3} by an impressive degree when it comes to tuning.
This package ships with grid search and random search, and the recently released \rpkg{tune} package comes with Bayesian optimization.
The AR racing approach and simulated annealing can be found in the \rpkg{finetune} package.
HP defaults and transformations are supported via \rpkg{dials}.
HPs of preprocessing operations using tidymodels' \rpkg{recipes} pipeline system can be tuned jointly with the HPs of the ML algorithm.
%Tuning over multiple models is supported via the \texttt{\href{https://github.com/tidymodels/workflowsets}{workflowsets}} package.

\paragraph[h2o]{\rpkg{h2o}} \citep{ledell2020} connects to the H2O cross-platform ML framework written in Java. 
Unlike the other discussed frameworks, which connect third-party packages from CRAN, \rpkg{h2o} ships with its own implementations of ML models.
The package supports grid search and random grid search, a variant of random search where points to be evaluated are randomly sampled from a discrete grid.
%Additionally, there is a mechanism to set an exploitation (vs.\ exploration) ratio (ER) via a control parameter, which triggers an iterative tuning of the learning rate of supported models.
% Hyperparameter transformations are still possible by just providing a custom grid to evaluate. 
The possibilities for preprocessing are limited to imputation, different encoders for categorical features, and correcting for class imbalances via under- and oversampling.
H2O automatically constructs a search space for a given set of learning algorithms and preprocessing methods, and HPs of both can be tuned jointly.
% For the former, the hyperparameters can be jointly tuned with the learner hyperparameters.
It is worth mentioning that \rpkg{h2o} was developed with a strong focus on AutoML and offers the functionality to perform random search over a pre-defined grid, evaluating configurations of generalized linear models with elastic net penalty, xgboost models, gradient boosting machines, and deep learning models.
The best performing configurations found by the AutoML system are then stacked together \citep{vanderLaan2007} for the final model.

%% file: appendix_D_02_software_python.tex
\subsection{Software in Python}\label{ssec:software_python}
\subsubsection{Machine Learning Package in Python}
%First we will briefly introduce well known general machine learning frameworks to date. We note that HPO can be applied to these, but model-based HPO approaches, such as Bayesian Optimization (BO), also make use of these packages for their surrogate models.

\texttt{Scikit-learn} \citep[sklearn]{scikit_learn} is a general ML framework implemented in Python and the default path for ML projects without the need for deep learning (DL). The framework provides the most important traditional ML models (incl. SVM, RFs, Boosting, etc.). This is further complemented by a variety of preprocessing and post-processing methods (e.g., ensembling). The clean and well-documented API allows users to easily build fairly complex pipelines, which can provide strong performance on tabular data, e.g., see results on auto-sklearn~\citep{feurer_nips2015a}. 
%This also include surrogate models for BO, such as random forest (RF) and Gaussian processes (GP). 

Modern DL frameworks such as \texttt{Tensorflow} \citep{tensorflow}, \texttt{PyTorch} \citep{pytorch_neurips19a} and \texttt{MXNet} \citep{chen_arxiv15a} accelerate large-scale computation task (in particular matrix multiplication) by executing them on GPUs. Furthermore, \texttt{GPflow} \citep{GPflow2017} and \texttt{GPyTorch} \citep{gardner2018gpytorch} are two GP libraries built on top of the two DL frameworks respectively and thus can be again used as surrogate models for BO. 

\subsubsection{Open-Source HPO Packages in Python}
The rapid development of the package landscape in Python has enabled many developers to also provide HPO tools in Python. Here, we will give an selective overview of commonly used and well-known packages.

\texttt{Spearmint} \citep{snoek_nips12a, spearmint} was one of the first successful open-source BO packages for HPO. As proposed by \citet{snoek_nips12a}, \texttt{Spearmint} implements standard BO with Markov chain Monte Carlo (MCMC) integration of the acquisition function for a fully Bayesian treatment of GP's HPs. Additionally, the second version of \texttt{Spearmint} allowed warping the input space with a Beta cumulative distribution function \citep{snoek_icml14a} to deal with non-stationary loss landscapes. \texttt{Spearmint} also supports 
%%ML: not officially implemented in spearmint (Feedback from Matthias Feurer)
%transferring knowledge across different tasks \citep{swersky_nips13a},
constrained BO \citep{gelbart_uai14a} and parallel optimization \citep{snoek_nips12a}.

\texttt{Hyperopt} \citep{hyperopt}, as indicated by its name, is a distributed HPO framework. Unlike other frameworks that estimate the potential performance given a configuration, Hyperopt implements a Tree of Parzen Estimators (TPE) \citep{bergstra_nips11a} that estimates the distribution of configurations given their performance. This approach allows it to easily run different configurations in parallel, as multiple density estimators can be executed at the same time.

\texttt{Scikit-optimize}~(skopt; \citeyear{scikit_optimize}) is a simple yet efficient library to optimize expensive and noisy black-box functions with BO. \texttt{Skopt} trains RF (including decision trees and grading-boosting trees) and GP models based on sklearn~\citep{scikit_learn} as surrogate models. Similar to sklearn, \texttt{skopt} allows for several ways of pre-processing input-features, e.g.\ one-hot encoding, log transformation, normalization, label encoding, etc.

\texttt{SMAC} \citep{hutter_lion11a, smac_2017} was originally developed as a tool for algorithm configurations \citep{eggensperger_jair19a}. However, in recent years, it has also been successfully used as a key component of several AutoML Tools, such as \texttt{auto-sklearn} \citep{feurer_nips2015a} and \texttt{Auto-PyTorch} \citep{zimmer_tpami21a}. \texttt{SMAC} implements both RF and GP as surrogate models to handle various sorts of HP spaces. Additionally, BOHB \citep{falkner_icml18a} is implemented as a multi-fidelity approach in SMAC; however, instead of the TPE model in BOHB, SMAC utilized RF as a surrogate model for BO part. The most recent version of \texttt{SMAC} \citep{SMAC3} also allows for parallel optimization with dask \citep{dask}.

Similarly, \texttt{HyperMapper} \citep{nardi2019practical} also builds an RF as a surrogate model; thus, it also supports mixed and structured HP configuration spaces. However, \texttt{HyperMapper} does not implement a Bayesian Optimization approach in a strict sense. In addition, \texttt{HyperMapper} incorporates previous knowledge by rescaling the samples with a beta distribution and handles unknown constraints by training another RF model as a probabilistic classifier. Furthermore, it supports multi-objective optimization. As \texttt{HyperMapper} cannot be allocated to a distributed computing environment, it might not be applicable to large-scale HPO problems. 

\texttt{OpenBox} \citep{openbox} is a general framework for black-box optimization -- including HPO -- and supports multi-objective optimization, multi-fidelity, early-stopping, transfer learning, and parallel BO via distributed parallelization under both synchronous parallel settings and asynchronous parallel settings. \texttt{OpenBox} further implements an ensemble surrogate model to integrate the information from previously seen similar tasks. 

\texttt{Dragonfly} \citep{Kandasamy_JMLR2020a} is an open source library for scalable BO. Dragonfly extends standard BO in the following ways to scale to higher dimensional and expensive HPO problems: it implements GPs with additive kernels and additive GP-UCB \citep{kandasamy_icml15a} as an acquisition function. In addition, it supports multi-fidelity approach to scale to expensive HPO problems. To increase the robustness of the system w.r.t.\ its HPs, \texttt{Dragonfly} acquires its GP HP by either sampling a set of GP HPs from the posterior, conditioned on the data or optimizing the likelihood to handle different degrees of smoothness of the optimized loss landscape. In addition, \texttt{Dragonfly} maximizes its acquisition function with an evolutionary algorithm, which enables the system to work on different sorts of configuration spaces, e.g.,\ different variable types and constraints. 

The previously described BO-based HPO tools fix their search space during the optimization phases. The package \texttt{Bayesian Optimization} \citep{bayesianoptimization} can concentrate a domain around the current optimal values and adjust this domain dynamically \citep{Stander2002OnTR}. A similar idea is implemented in \texttt{TurBO} \citep{eriksson2019scalable}, which only samples inside a trust region while keeping the configuration space fixed. The trust region shrinks or extends based on the performance of \texttt{TurBO}'s suggested configuration. 

\texttt{GPflowOpt} \citep{gpflowopt} and \texttt{BoTorch} \citep{balandat2020botorch} are the BO-based HPO frameworks that build on top of \texttt{GPflow} and \texttt{GPyTorch}, respectively. The auto-grading system facilitates the users to freely extend their own ideas to existing models. \texttt{Ax} adds an easy-to-use API on top of \texttt{BoTorch}.

\texttt{DEHB} \citep{dehb} uses differential evolution (DE) and combines it with a multi-fidelity HPO framework, inspired by \texttt{BOHB}'s combination of hyperband \citep{li_2018} and BO. \texttt{DEHB} overcomes several drawbacks of \texttt{BOHB}: it does not require a surrogate model, and thus, its runtime overhead does not grow over time; \texttt{DEHB} can have stronger final performance compared to \texttt{BOHB}, especially for discrete-valued and high-dimensional problems, where BO usually fails, as it tends to suggest points on the boundary of the search space \citep{oh_icml18a}; 
additionally, it is simpler to implement an efficient parallel DE algorithm compared to a parallel BO-based approach. For instance, \texttt{DEAP} \citep{DEAP_JMLR2012} and \texttt{Nevergrad} \citep{nevergrad} have many EA implementations that enable parallel computation. 
 
There are several other general HPO tools. For instance, \texttt{Optuna} \citep{akiba_kdd2019} is an automatic HP optimization software framework that allows to dynamically construct the search space (\texttt{Define-by-run} API) and thus makes it much easier for users to construct a highly complex search space. \texttt{Or\'ion} \citep{xavier_bouthillier_2020_4265424} is an asynchronous framework for black-box function optimization. Finally, \texttt{Tune} \citep{tune} is a scalable HP tuning framework that provides APIs for several optimizers mentioned before, e.g.,\ Dragonfly, SKopt, or HyperOpt.

\subsubsection{AutoML tools}
So far, we discussed several HPO tools in Python, which allow for flexible applications of different algorithms, search spaces and data sets. Here, we will briefly discuss several AutoML-packages of which HPO is a part. 

\texttt{Auto-Weka} \citep{thornton_kdd13a} is one of the first AutoML tools that automates the design choice of entire ML packages, namely Weka, with \texttt{SMAC}. Similarly, the same optimizer is applied to \texttt{Auto-sklearn} \citep{feurer_nips2015a}, while a meta-learning and ensemble approach further boost the performance of \texttt{auto-sklearn}. Similarly, \texttt{Auto-keras} \citep{jin2019auto} and \texttt{Auto-PyTorch} \citep{zimmer_tpami21a} use BO to automate the design choices of deep NNs. \texttt{TPOT} \citep{olson_gecco16a} automates the design choice of an ML pipeline with genetic algorithms, while a probabilistic grammatical evolution is implemented in \texttt{AutoGOAL} \citep{estevez2020general} for HPO problems. \texttt{H2O automl} \citep{H2OAutoML20} and \texttt{AutoGluon}  \citep{autogluo} train a set of base models and ensemble them with a single layer (H2O) or multi-layer stacks (AutoGluon). Additionally, \texttt{TransmogrifAI} \citep{transmogrifAI} is an AutoML tool built on the top of \texttt{Apache Spark} and \texttt{MLBox} \citep{mlbox}. 

% IMPORTANT:
%\bibliography{references,strings,lib,proc}